\pdfoutput=1

\documentclass[11pt]{article}

\usepackage[table,xcdraw]{xcolor} 

\usepackage[final]{acl}

\usepackage{times}
\usepackage{latexsym}
\usepackage{booktabs}
\usepackage[T1]{fontenc}

\usepackage[utf8]{inputenc}

\usepackage{microtype}

\usepackage{inconsolata}

\usepackage{graphicx}
\usepackage{bbding}
\usepackage{amssymb}
\usepackage{makecell}
\usepackage{multirow}
\usepackage{tablefootnote}
\usepackage{amsmath}
\usepackage{array} 
\usepackage{mathrsfs}

\title{OpenHuEval: Evaluating Large Language Model on Hungarian Specifics}

\author{
 \textbf{Haote Yang\textsuperscript{1}\thanks{Equal contribution.}},
 \textbf{Xingjian Wei\textsuperscript{1}\footnotemark[1]},
 \textbf{Jiang Wu\textsuperscript{1}\footnotemark[1]\thanks{Project lead.}},
 \textbf{Noémi Ligeti-Nagy\textsuperscript{2}\footnotemark[1]}, \\
 \textbf{Jiaxing Sun\textsuperscript{3}\footnotemark[1]},
 \textbf{Yinfan Wang\textsuperscript{4}\footnotemark[1]},
 \textbf{Zijian Győző Yang\textsuperscript{2}\footnotemark[1]},
 \textbf{Junyuan Gao\textsuperscript{5}\footnotemark[1]},
 \textbf{Jingchao Wang\textsuperscript{6}},  \\
 \textbf{Bowen Jiang\textsuperscript{7}},
 \textbf{Shasha Wang\textsuperscript{1}},
 \textbf{Nanjun Yu\textsuperscript{6}},
 \textbf{Zihao Zhang\textsuperscript{8}},
 \textbf{Shixin Hong\textsuperscript{9}},
 \textbf{Hongwei Liu\textsuperscript{1}}, \\
 \textbf{Wei Li\textsuperscript{1}},
 \textbf{Songyang Zhang\textsuperscript{1}},
 \textbf{Dahua Lin\textsuperscript{1,10}},
 \textbf{Lijun Wu\textsuperscript{1}},
 \textbf{Gábor Prószéky\textsuperscript{2}},
 \textbf{Conghui He\textsuperscript{1}\thanks{Corresponding author. \href{heconghui@pjlab.org.cn}{heconghui@pjlab.org.cn}}}
\\
 \textsuperscript{1}Shanghai Artificial Intelligence Laboratory, \\
 \textsuperscript{2}HUN-REN Hungarian Research Centre for Linguistics,
 \textsuperscript{3}Wuhan University, \\
 \textsuperscript{4}Shanghai University,
 \textsuperscript{5}University of Chinese Academy of Sciences, \\
 \textsuperscript{6}East China Normal University,
 \textsuperscript{7}Peking University,
 \textsuperscript{8}Harbin Institute of Technology, \\
 \textsuperscript{9}Tsinghua University,
 \textsuperscript{10}Chinese University of Hong Kong
\\
}

\begin{document}
\maketitle
\begin{abstract}
We introduce OpenHuEval, the first benchmark for LLMs focusing on the Hungarian language and specifics.
OpenHuEval is constructed from a vast collection of Hungarian-specific materials sourced from multiple origins.
In the construction, we incorporated the latest design principles for evaluating LLMs, such as using real user queries from the internet, emphasizing the assessment of LLMs' generative capabilities, and employing LLM-as-judge to enhance the multidimensionality and accuracy of evaluations.
Ultimately, OpenHuEval encompasses eight Hungarian-specific dimensions, featuring five tasks and 3953 questions.
Consequently, OpenHuEval provides the comprehensive, in-depth, and scientifically accurate assessment of LLM performance in the context of the Hungarian language and its specifics.
We evaluated current mainstream LLMs, including both traditional LLMs and recently developed Large Reasoning Models. The results demonstrate the significant necessity for evaluation and model optimization tailored to the Hungarian language and specifics. 
We also established the framework for analyzing the thinking processes of LRMs with OpenHuEval, revealing intrinsic patterns and mechanisms of these models in non-English languages, with Hungarian serving as a representative example.
We will release OpenHuEval at \url{https://github.com/opendatalab/OpenHuEval} .

\end{abstract}
\section{Introduction}
Recent advancements in Large Language Models (LLMs)~\cite{openai-o1,deepseek-r1}represent significant strides toward artificial general intelligence (AGI). However, notable performance gaps remain between English and other languages in both language-agnostic tasks (e.g., math reasoning, code generation)~\cite{benchmax,PMMEval} and language-specific tasks (e.g., idiom usage, cultural understanding)~\cite{havingbeer,charm,MAPS}, posing challenges to global AI deployment and equitable development.
The disparities in cross-lingual performance arise mainly from two factors: First, the training data, particularly the pre-training data, is heavily skewed toward English. Second, while English evaluation benchmarks are advanced and rapidly evolving, non-English benchmarks are underdeveloped, particularly for language-specific features, limiting the identification of shortcomings in non-English languages and leading to their neglect in research.

This paper focuses on the evaluation of Hungarian language and specifics. Hungarian is spoken by around 14 million people worldwide. Research on the Hungarian language not only improves the user experience for Hungarian speakers but also offers valuable insights for similar studies in other languages and regions.
Existing Hungarian evaluation datasets are largely translations of English ones, missing essential Hungary-specific elements such as language nuances, culture, history, and regional context, which are key for Hungarian users. Among the existing evaluation datasets, HuLU \cite{hulu} is the key benchmark for Hungarian language understanding, but its focus on multiple-choice and true/false questions limits its ability to evaluate broader LLM capabilities, such as language generation, open-domain Q\&A, reasoning and instruction-following.

\begin{table*}[t] 
\scriptsize
\centering
\begin{tabular}{ccccccc}
\Xhline{1.5pt}
\textbf{\begin{tabular}[c]{@{}c@{}}Benchmark\end{tabular}} & \textbf{\begin{tabular}[c]{@{}c@{}}Real user\\ query\end{tabular}} & \textbf{\begin{tabular}[c]{@{}c@{}}Self-awareness\\ evaluation\end{tabular}} & \textbf{\begin{tabular}[c]{@{}c@{}}Proverb \\ Reasoning\end{tabular}} & \textbf{\begin{tabular}[c]{@{}c@{}}Generative task \\ \& llm-as-judge\end{tabular}} & \textbf{\begin{tabular}[c]{@{}c@{}}Hungarian\\ Lang\end{tabular}} & \textbf{\begin{tabular}[c]{@{}c@{}}Comprehensive \\ Hu-specific\end{tabular}} \\
\midrule
\textbf{\begin{tabular}[c]{@{}c@{}}WildBench\cite{WildBench}\end{tabular}} & \Checkmark & \XSolidBrush & \XSolidBrush & \Checkmark & \XSolidBrush & \XSolidBrush \\
\rowcolor[HTML]{F2F2F2}
\textbf{\begin{tabular}[c]{@{}c@{}}SimpleQA\cite{wei2024simpleQA},\\ChineseSimpleQA\cite{he2024chinesesimpleqa}\end{tabular}} & \XSolidBrush & \Checkmark & \XSolidBrush & \Checkmark & \XSolidBrush & \XSolidBrush \\
\textbf{MAPS\cite{MAPS}} & \XSolidBrush & \XSolidBrush & \Checkmark & \XSolidBrush & \XSolidBrush & \XSolidBrush \\
\rowcolor[HTML]{F2F2F2}
\textbf{\begin{tabular}[c]{@{}c@{}}MARC, MMMLU et al in \cite{lai2023okapi}\end{tabular}} & \XSolidBrush & \XSolidBrush & \XSolidBrush & \XSolidBrush & \Checkmark & \XSolidBrush \\
\textbf{\begin{tabular}[c]{@{}c@{}}BenchMAX\cite{benchmax}\end{tabular}} & \XSolidBrush & \XSolidBrush & \XSolidBrush & \Checkmark & \Checkmark & \XSolidBrush \\
\rowcolor[HTML]{F2F2F2}
\textbf{\begin{tabular}[c]{@{}c@{}}MILQA\cite{MILQA}\end{tabular}} & \XSolidBrush & \XSolidBrush & \XSolidBrush & \XSolidBrush & \Checkmark & \XSolidBrush \\
\textbf{HuLU\cite{hulu}} & \XSolidBrush & \XSolidBrush & \XSolidBrush & \XSolidBrush & \Checkmark & \XSolidBrush \\
\rowcolor[HTML]{F2F2F2}
\textbf{OpenHuEval (ours)} & \Checkmark & \Checkmark & \Checkmark & \Checkmark & \Checkmark & \Checkmark \\
\Xhline{1.5pt}
\end{tabular}
\caption{Comparison of related benchmarks.}
\label{tab:Comparison of related benchmarks.}
\end{table*}

\begin{figure*}[!t]
    \centering
    \includegraphics[width=1\linewidth]{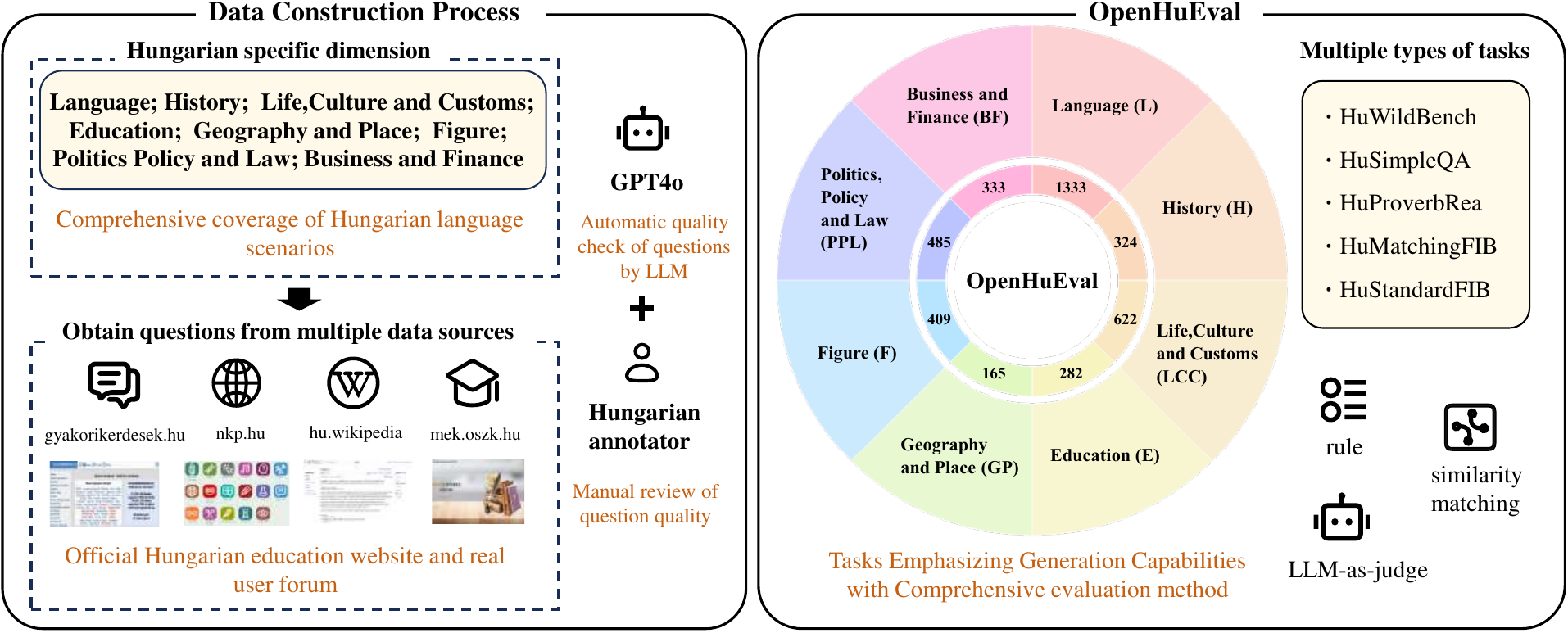}
    \caption{Overview of OpenHuEval.}
    \label{fig:fig1_OpenHuEval}
\end{figure*}

To address this gap, we introduces OpenHuEval, the first evaluation benchmark for LLMs focused on Hungarian language and specifics. The comparison of OpenHuEval with the existing related benchmarks is shown in Table \ref{tab:Comparison of related benchmarks.}. Overall, OpenHuEval has two main distinguishing features:

1) \textbf{Hungarian-Specific} Inspired by \cite{liu2024culturally,charm}, we propose eight distinct Hungarian-specific dimensions (see \S\ref{sec:Hungarian-specific dimensions}), covering a variety of scenarios that users may encounter when querying in Hungarian. Guided by these dimensions, we collected a vast amount of Hungarian specific material from multiple sources and used this to construct the corresponding evaluation tasks.

2) \textbf{Keeping up with the latest advances in LLM evaluation} Significant progress has been made in LLM evaluation, with query sources shifting from manual or rule-based constructions to real-world internet questions~\cite{WildBench}, enhancing practical relevance. Question formats evolved from multiple-choice to open-ended Q\&A~\cite{wei2024simpleQA}, better reflecting actual usage. Evaluation methods transitioned from rule-based approaches to LLM-as-judge and subjective assessments, improving accuracy and objectivity~\cite{llm_as_judge_survey}. However, these advancements primarily apply to English datasets and not Hungarian. Thus, when creating OpenHuEval, we incorporated these principles and methodologies from English evaluations.

Based on OpenHuEval, we evaluated the performance of mainstream LLMs on Hungarian language and specifics. We compared the performance differences of these models on the typical datasets of OpenHuEval with corresponding datasets in other languages. The results indicate a significant necessity for evaluation and model optimization specifically for Hungarian language and specifics. 

Moreover, Large Reasoning Models (LRMs), like o1, mark a new direction in LLM development. Through extensive reasoning, self-reflective negation, and exploring multiple reasoning paths, they greatly improve reasoning abilities, adhering to the test-time scale law—a crucial step toward AGI. Recent studies~\cite{wang2025thoughts} have analyzed these reasoning processes, offering insights for optimization, but have largely focused on English-language contexts, neglecting Hungarian language and specific scenarios. Building on OpenHuEval, we developed the framework for dissecting the reasoning processes of LRMs. Using Hungarian as the example, we uncovered intrinsic patterns of the representative LRMs in non-English languages. These findings provide valuable insights for the research community to further advance the development of LRMs.

In summary, the contributions of this paper include the following three points:

\textbf{-} We developed OpenHuEval, the first benchmark for LLMs focusing on the Hungarian language and specifics. OpenHuEval incorporates the latest design principles for evaluating LLMs, such as using real user queries from the internet, emphasizing the assessment of LLMs' generative capabilities, and employing LLM-as-judge to enhance the multidimensionality and accuracy of evaluations.

\textbf{-} We conducted the comprehensive evaluation of current mainstream LLMs, including traditional LLMs and recently developed LRMs. The results highlight the significant necessity for evaluation and model optimization tailored to Hungarian language and specifics.

\textbf{-} We established the framework for analyzing the thinking processes of the cutting-edge LRMs, revealing the intrinsic patterns and mechanisms of these models in the Hungarian language and specifics, and providing a reference for research related to other non-English languages.

\begin{figure*}[t]
    \centering
    \includegraphics[width=1\linewidth]{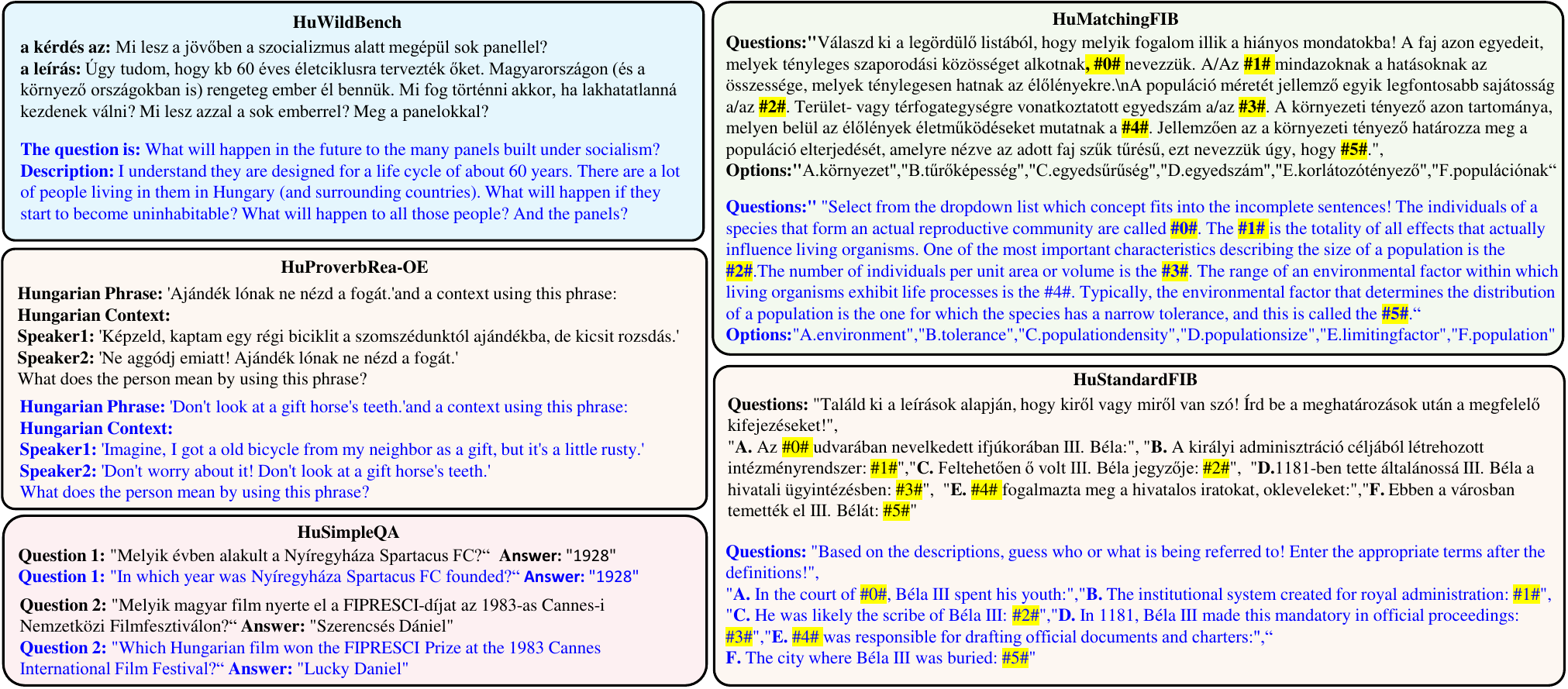}
    \caption{Examples of OpenHuEval. The original text is in black, while the translation into English is in \textcolor{blue}{blue}. In HuMatchingFIB and HuStandardFIB, the blank is \colorbox{yellow}{highlighted}.}
    \label{fig:all_task_example}
\end{figure*}

\begin{table*}[t]
\scriptsize
\centering
\begin{tabular}{l>{\arraybackslash}p{8cm}c}
\Xhline{1.5pt}
\textbf{\begin{tabular}[c]{@{}c@{}}HuSpecificDim\end{tabular}} &
  \textbf{Definition} &
  \textbf{\#Question} \\ \hline
\rowcolor[HTML]{F2F2F2} 
\textbf{Language($\mathcal{L}$)} &
  Basic knowledge of the Hungarian language and Hungarian proverbs and sayings &
  1333 \\
\textbf{History($\mathcal{H}$)} &
  Historical events and historical development of Hungary &
  324 \\
\rowcolor[HTML]{F2F2F2} 
\textbf{\begin{tabular}[c]{@{}c@{}}Life, Culture,  and Custom($\mathcal{LCC}$)\end{tabular}} &
  Religion, rituals, culture, holidays, and the daily life of Hungarians &
  622 \\
\textbf{\begin{tabular}[c]{@{}c@{}}Education and  Profession($\mathcal{EP}$)\end{tabular}} &
  Education system in Hungary and related professions &
  282 \\
\rowcolor[HTML]{F2F2F2} 
\textbf{\begin{tabular}[c]{@{}c@{}}Geography and  Place($\mathcal{GP}$)\end{tabular}} &
  Geographical knowledge of Hungary, cities, and locations &
  165 \\
\textbf{Figure($\mathcal{F}$)} &
  Famous figures of Hungary &
  409 \\
\rowcolor[HTML]{F2F2F2} 
\textbf{\begin{tabular}[c]{@{}c@{}}Politics, Policy and Law($\mathcal{PPL}$)\end{tabular}} &
  Politics, policies, and laws of Hungary &
  485 \\
\textbf{\begin{tabular}[c]{@{}c@{}}Business and Finance($\mathcal{BF}$)\end{tabular}} &
  Business and finance in Hungary &
  333 \\ \Xhline{1.5pt}
\end{tabular}
\caption{The Hungarian-Specific Dimensions (HuSpecificDim).}
\label{tab:Hungarian-specific dimensions}
\end{table*}

\section{OpenHuEval}

OpenHuEval is the benchmark specifically designed to evaluate the performance of LLM in handling Hungarian language and specifics. The overview of OpenHuEval is in Figure \ref{fig:fig1_OpenHuEval}. Examples are shown in Figure \ref{fig:all_task_example}.

\subsection{Hungarian-specific dimensions and OpenHuEval tasks}
\label{sec:Hungarian-specific dimensions}
Inspired by \cite{liu2024culturally,charm}, OpenHuEval encompasses eight Hungarian-Specific Dimensions (HuSpecificDim), as shown in Table \ref{tab:Hungarian-specific dimensions}: \textit{Language} ($\mathcal{L}$), \textit{History} ($\mathcal{H}$), \textit{Life, Culture, and Customs} ($\mathcal{LCC}$), \textit{Education and Profession} ($\mathcal{EP}$), \textit{Geography and Place} ($\mathcal{GP}$), \textit{Figure} ($\mathcal{F}$), \textit{Politics, Policy, and Law} ($\mathcal{PPL}$), and \textit{Business and Finance} ($\mathcal{BF}$). These dimensions comprehensively cover a wide range of scenarios encountered by users when utilizing Hungarian as the query language. As a result, they enable the systematic evaluation of the performance of LLMs in tasks related to the Hungarian.

\begin{table*}[h]
\scriptsize
\centering
\begin{tabular}{c>{\centering\arraybackslash}p{3cm}ccc}
\Xhline{1.5pt}
\textbf{\begin{tabular}[c]{@{}c@{}}Task\end{tabular}} &
\textbf{\begin{tabular}[c]{@{}c@{}}HuSpecificDim\end{tabular}} &
\textbf{\begin{tabular}[c]{@{}c@{}}Judge\end{tabular}} &
\textbf{\begin{tabular}[c]{@{}c@{}}Question type\end{tabular}} &
\textbf{\#Question} \\ \hline

\rowcolor[HTML]{EFEFEF} 
\textbf{HuWildBench}   & $\mathcal{LCC}$, $\mathcal{EP}$, $\mathcal{PPL}$, $\mathcal{BF}$       & llm,checklist & OE                        & 1154 \\
\rowcolor[HTML]{FFFFFF} 
\textbf{HuSimpleQA}    & $\mathcal{L}$,$\mathcal{H}$,$\mathcal{LCC}$,$\mathcal{EP}$,$\mathcal{GP}$,$\mathcal{F}$,$\mathcal{PPL}$,$\mathcal{BF}$ & llm           & OE                        & 1293 \\
\rowcolor[HTML]{EFEFEF} 
\textbf{HuProverbRea}  & $\mathcal{L}$                      & rule,llm      & 2CQ/OE                    & 1135 \\
\rowcolor[HTML]{FFFFFF} 
\textbf{HuMatchingFIB} & $\mathcal{L}$, $\mathcal{H}$                   & rule          & Matching Filling-in-Blank & 278  \\
\rowcolor[HTML]{EFEFEF}
\textbf{HuStandardFIB} &$\mathcal{L}$, $\mathcal{H}$ &
  rule,similarity matching &
  Standard Filling-in-Blank &93 \\ \Xhline{1.5pt}
\end{tabular}
\caption{Tasks of OpenHuEval}
\label{tab:Tasks of OpenHuEval}
\end{table*}

Bearing the above HuSpecificDim in mind, the first step in building OpenHuEval involves gathering corpora rich in Hungarian specifics. Inspired by~\cite{MAPS,li2023cmmlu,he2024chinesesimpleqa,arora2024calmqa}, we collected data from sources such as Hungarian proverbs, exam questions, forums, and Wikipedia. Through processes including filtering, refinement, construction, and quality assurance, we developed a total of five evaluation tasks comprising 3953 questions in total, as detailed in Table \ref{tab:Tasks of OpenHuEval}. The subsequent sections of this chapter will introduce these tasks and their corresponding datasets in detail.

\subsection{Hungarian WildBench}
\textbf{Task Introduction:} The Hungarian WildBench (HuWildBench) task aims to evaluate the performance of LLMs in answering various questions arising from the everyday lives of Hungarians. 
All questions are sourced from Hungary’s well-known forum website\footnote{\url{https://www.gyakorikerdesek.hu/}, which is similar to \url{https://www.quora.com/} for English-speaking world.} (hereinafter referred to as ``g13k'' for brevity).
These questions cover a wide range of topics, including cultural customs, education, tourism, legal regulations, and business and finance, thus reflect \textbf{real-life issues encountered by Hungarians}. 
Examples of HuWildBench questions are shown in Figure~\ref{fig:all_task_example} and Table~\ref{tab:HuWildBench_QuestionExamples}. 
The queries in HuWildBench are user-generated content, therefore their linguistic expressions and question formats tend to be less formal than the structured and polished written language. This poses the realistic challenge for LLMs, as they must adapt to such informal and spontaneous language style.
The construction of HuWildBench is detailed in Appendix~\ref{Appendix:HuWildBench}.

\textbf{Metric and judge:} We use the WB-Score~\cite{WildBench} as evaluation metric for HuWildBench, which is calculated in the following manner:
We have simultaneously developped the customized checklist for each question during the process of creating each question, to assist the LLM judge in evaluating the responses. Examples of these checklists can be seen in Table~\ref{tab:HuWildBench_QuestionExamples}.
Following \cite{WildBench}, GPT-4o is used as the judge model, which evaluates the quality of each response based on the checklist and provides detailed strengths and weaknesses before assigning a score from 1 to 10. The definition of scores is shown in Table~\ref{tab:HuWildBench_socre_definition} and the judge prompt is detailed in Appendix~\ref{Appendix:HuWildBenchLLMAsJudge}. 
Different from \cite{WildBench}, our final scores are calculated as the average of all test sample scores, with each score multiplied by 10.

\subsection{Hungarian SimpleQA}
\textbf{Task Introduction:} Hungarian SimpleQA (HuSimpleQA) is designed to assess the ability of LLMs to answer short, fact-seeking questions related to Hungary. Inspired by \cite{wei2024simpleQA} and \cite{he2024chinesesimpleqa}, we constructed HuSimpleQA based on Hungarian Wikipedia\footnote{\url{https://hu.wikipedia.org/}}, with the following key characteristics. \textbf{Hungarian:} The questions in HuSimpleQA are in Hungarian, and they focus on facts specifically related to Hungary. \textbf{Diverse:} The questions in HuSimpleQA cover the eight Hungary-specific dimensions proposed in \S\ref{sec:Hungarian-specific dimensions}. \textbf{High-quality:} The construction process of HuSimpleQA (in Appendix~\ref{App:husimpleqa}) includes comprehensive and strict quality control procedures, ensuring the quality and accuracy of the questions. \textbf{Static:} Similar to SimpleQA, the answers to the questions in HuSimpleQA do not change over time, ensuring that the dataset remains evergreen. \textbf{Easy-to-evaluate:} The questions and answers in HuSimpleQA are short and concise, making them ideal for evaluation through LLMs.
The examples of HuSimpleQA are shown in Figure~\ref{fig:all_task_example} and Table~\ref{tab:husimleqa_examples}.
The construction of HuSimpleQA is detailed in Appendix~\ref{App:husimpleqa}.

\textbf{Metric and Judge:} Following~\cite{wei2024simpleQA}, we use GPT-4o as a judge to categorize the responses of the LLM to HuSimpleQA into three classes: \texttt{CORRECT}, \texttt{INCORRECT} or  \texttt{NOT\_ATTEMPTED}. The judge prompt can be found in Appendix~\ref{App:husimpleqa_judge}, Figure \ref{fig:Appendix_HuSimpleQA_prompt_judge}.

\subsection{Hungarian Proverb Reasoning}

\textbf{Task Introduction}: Hungarian Proverb Reasoning (HuProverbRea), which consists of the collection of Hungarian proverbs, idioms, abbreviations, is a task that requires the LLM to \textbf{understand and reason the meaning of Hungarian proverbs in a specific context}. As shown by the examples in Figure~\ref{fig:all_task_example}, LLM is provided with a context in which a Hungarian proverb is used, accompanied by a question: \textit{``What does the speaker mean by the saying?''}. Then, the LLM is tasked with discerning the speaker's true intention, either by selecting the correct interpretation from two provided options (2CQ setting), or by directly articulating the speaker's intended meaning (OE setting).
The construction of HuProverbRea is detailed in Appendix~\ref{Appendix:HuProverbRea}.

\textbf{Metric and judge:} For the 2CQ setting, we simply measure the correct ratio of candidate LLMs. For the OE setting, we adopt GPT-4o as judge to decide if the answer is acceptable. We provide the original proverb, its context and the English explanation of the proverb as references when judging OE responses. Detailed prompt templates are listed in Appendix \ref{Appendix:HuProverbRea}.

\subsection{Hungarian Matching and Standard Filling-in-Blank}
\textbf{Task Introduction}: Hungarian Matching Fill-in-the-Blank (HuMatchingFIB) is a task where key terms in a text are removed, and a pool of candidate words or phrases is provided. This pool includes both correct answers and distractors. The task requires the LLM to choose the most suitable words from the pool to fill in the blanks, thus restoring the full meaning of the text. The example is shown in Figure ~\ref{fig:all_task_example} and Figure ~\ref{fig:Appendix_HuMatchingFIB_example}. 
HuMatchingFIB effectively tests the LLM's abilities in understanding information, reasoning within context, and differentiating correct answers from distractors.

In contrast, Hungarian Standard Fill-in-the-Blank (HuStandardFIB) also involves filling in blanks but does not offer a candidate pool. Instead, the model must rely on its internal knowledge and the provided context to complete the text.
The examples are shown in Figure \ref{fig:Appendix_HuStandardFIB_example}. Consequently, HuStandardFIB evaluates the LLM's overall ability to recall knowledge and reason within context.

The constructions of HuMatchingFIB and HuStandardFIB are detailed in Appendix~\ref{Appendix:HuMatchingFIB and HuStandardFIB}.

\textbf{Metric and Judge:} 
In our inference prompts, we explicitly instruct the LLM to generate responses in a specified format. Additionally, we have established a set of well-defined rules to evaluate the correctness of the LLM's answers for each blank. The detailed format requirements and judgment criteria can be found in the Appendix~\ref{Appendix:HuMatchingFIB and HuStandardFIB}.

For both HuMatchingFIB and HuStandardFIB, we evaluate performance at two levels: blank-level and question-level accuracy. Specifically, \textbf{Acc\_b} (Blank-level Accuracy) measures the proportion of blanks that the model answers correctly across all questions. On the other hand, \textbf{Acc\_q} (Question-level Accuracy) evaluates the proportion of questions that the model answers entirely correctly. A question is only considered correct if \textbf{all} its associated blanks are answered accurately.

\section{Experiments and Analysis}
\subsection{Experimental setup}
We evaluated the currently mainstream LLMs, including GPT-4o \cite{gpt4o}, GPT-4o mini\footnote{We used the gpt-4o-2024-11-20 version for GPT-4o and the gpt-4o-mini-2024-07-18 version for GPT-4o-mini.}, Deepseek-V3 \cite{deepseekv3}, Qwen2.5-Instruct \cite{qwen2.5}, and Llama-3.1-Instruct \cite{llama3}, as well as the latest Large Reasoning Models (LRMs) such as OpenAI o1-mini \cite{openai-o1}, QwQ-32B-Preview \cite{qwq} (abbreviated as QwQ in following text), and Deepseek-R1 \cite{deepseek-r1}. Detailed specifications of these models are provided in Table \ref{tab:models}.

We used OpenCompass\footnote{\url{https://github.com/open-compass/opencompass}} in all our experiments. For traditional instruction-based LLMs, we adopted OpenCompass's default settings for the maximum output length. For Large Reasoning Models, we set the output length to 8192 to ensure sufficient space for reasoning process and to produce a complete final answer, avoiding premature output truncation. For OpenAI models (GPT series and o1-mini), we used their official API with settings following OpenCompass's default configuration. For Deepseek-V3 and Deepseek-R1, due to the high usage volume of Deepseek's official API causing instability, we used equivalent API services provided by Alibaba Cloud\footnote{\url{https://cn.aliyun.com/}} and Silicon Valley Flow\footnote{\url{https://siliconflow.cn/}}. The settings followed OpenCompass's configurations, with the temperature set to 0.7. For other models in Table \ref{tab:models}, we performed inference locally with NVIDIA A100 GPUs, using LMDeploy\footnote{\url{https://github.com/InternLM/lmdeploy}} as the inference backend. The settings followed OpenCompass's default configuration (Temperature = 1e-6, top\_k = 1).

\begin{table}[t]
\centering
\tiny  
\begin{tabular}{ccccc}
\Xhline{1.5pt}
\textbf{Model} &
\textbf{Size} &
\begin{tabular}[c]{@{}c@{}}\textbf{Reasoning}\\ \textbf{Model}\end{tabular} &
\begin{tabular}[c]{@{}c@{}}\textbf{Open-}\\ \textbf{source}\end{tabular} &
\textbf{Inference Method} \\ \hline
\rowcolor[HTML]{EFEFEF} 
\cellcolor[HTML]{EFEFEF} &
  \cellcolor[HTML]{EFEFEF} &
  \cellcolor[HTML]{EFEFEF} &
  \cellcolor[HTML]{EFEFEF} &
  \cellcolor[HTML]{EFEFEF} \\
\rowcolor[HTML]{EFEFEF} 
\multirow{-2}{*}{\cellcolor[HTML]{EFEFEF}GPT-4o} &
  \multirow{-2}{*}{\cellcolor[HTML]{EFEFEF}-} &
  \multirow{-2}{*}{\cellcolor[HTML]{EFEFEF}N} &
  \multirow{-2}{*}{\cellcolor[HTML]{EFEFEF}N} &
  \multirow{-2}{*}{\cellcolor[HTML]{EFEFEF}Official API} \\
GPT-4o-mini &-
   &
  N &
  N &
  Official API \\
\rowcolor[HTML]{EFEFEF} 
\cellcolor[HTML]{EFEFEF} &
  \cellcolor[HTML]{EFEFEF} &
  \cellcolor[HTML]{EFEFEF} &
  \cellcolor[HTML]{EFEFEF} &
  \cellcolor[HTML]{EFEFEF} \\
\rowcolor[HTML]{EFEFEF} 
\multirow{-2}{*}{\cellcolor[HTML]{EFEFEF}Deepseek-V3} &
  \multirow{-2}{*}{\cellcolor[HTML]{EFEFEF}-} &
  \multirow{-2}{*}{\cellcolor[HTML]{EFEFEF}N} &
  \multirow{-2}{*}{\cellcolor[HTML]{EFEFEF}Y} &
  \multirow{-2}{*}{\cellcolor[HTML]{EFEFEF}\begin{tabular}[c]{@{}c@{}}Alibaba Cloud and\\  SiliconFlow API\end{tabular}} \\
Qwen2.5-Instruct &
  7B,72B &
  N &
  Y &
  Local GPU \\
\rowcolor[HTML]{EFEFEF} 
Llama-3.1-Instruct &
  8B,70B &
  N &
  Y &
  Local GPU \\
o1-mini &
  - &
  Y &
  N &
  Official API \\
\rowcolor[HTML]{EFEFEF} 
QwQ &
  32B &
  Y &
  Y &
  Local GPU \\
Deepseek-R1 &
  - &
  Y &
  Y &
  \begin{tabular}[c]{@{}c@{}}Alibaba Cloud and\\  SiliconFlow API\end{tabular} \\ \Xhline{1.5pt}
\end{tabular}
\caption{LLMs evaluated in our experiments. }
\label{tab:models}
\end{table}

\subsection{Overall performance}

\begin{table*}[h]
\scriptsize
\centering
\begin{tabular}{
>{\columncolor[HTML]{FFFFFF}}c 
>{\columncolor[HTML]{EFEFEF}}c 
>{\columncolor[HTML]{FFFFFF}}c 
>{\columncolor[HTML]{EFEFEF}}c 
>{\columncolor[HTML]{EFEFEF}}c 
>{\columncolor[HTML]{FFFFFF}}c 
>{\columncolor[HTML]{FFFFFF}}c 
>{\columncolor[HTML]{EFEFEF}}c 
>{\columncolor[HTML]{EFEFEF}}c }
\Xhline{1.5pt}
\cellcolor[HTML]{FFFFFF} &
  \textbf{HuWildBench} &
  \textbf{HuSimpleQA} &
  \multicolumn{2}{c}{\cellcolor[HTML]{EFEFEF}\textbf{HuProverbRea}} &
  \multicolumn{2}{c}{\cellcolor[HTML]{FFFFFF}\textbf{HuMatchingFIB}} &
  \multicolumn{2}{c}{\cellcolor[HTML]{EFEFEF}\textbf{HuStandardFIB}} \\ \cline{2-9} 
\multirow{-2}{*}{\cellcolor[HTML]{FFFFFF}\textbf{Model}} &
  \textbf{WBScore} &
  \textbf{Acc} &
  \textbf{Acc. (OE)} &
  \textbf{Acc. (2CQ)} &
  \textbf{B acc.} &
  \textbf{Q acc.} &
  \textbf{B acc.} &
  \textbf{Q acc.} \\ \Xhline{1.0pt}
\textbf{GPT-4o} &
  {\color[HTML]{32CB00} 81.09} &
  {\color[HTML]{CB0000} 50.3} &
  {\color[HTML]{CB0000} 89.16} &
  {\color[HTML]{CB0000} 95.51} &
  {\color[HTML]{32CB00} 77.78} &
  {\color[HTML]{32CB00} 43.88} &
  {\color[HTML]{32CB00} 57.36} &
  {\color[HTML]{32CB00} 15.05} \\
\textbf{GPT-4o-mini} &
  74.19 &
  25.56 &
  {\color[HTML]{32CB00} 84.67} &
  92.16 &
  55.68 &
  19.78 &
  35.08 &
  7.53 \\
\textbf{QwQ} &
  58.02 &
  9.09 &
  67.49 &
  84.23 &
  38.65 &
  12.23 &
  6.05 &
  0 \\
\textbf{Deepseek-R1} &
  {\color[HTML]{CB0000} 82.96} &
  {\color[HTML]{3166FF} 34.58} &
  82.29 &
  91.72 &
  {\color[HTML]{CB0000} 80.87} &
  {\color[HTML]{CB0000} 47.12} &
  {\color[HTML]{CB0000} 61.76} &
  {\color[HTML]{CB0000} 17.2} \\
\textbf{Deepseek-V3} &
  {\color[HTML]{3166FF} 78.42} &
  32.71 &
  {\color[HTML]{3166FF} 83.26} &
  {\color[HTML]{3531FF} 92.51} &
  {\color[HTML]{3166FF} 68.87} &
  {\color[HTML]{3166FF} 39.93} &
  {\color[HTML]{3166FF} 51.44} &
  {\color[HTML]{3166FF} 9.68} \\
\textbf{Llama-3.1-Instruct-70B} &
  61.78 &
  {\color[HTML]{32CB00} 35.99} &
  80.18 &
  {\color[HTML]{009901} 93.83} &
  59.56 &
  24.46 &
  40.99 &
  6.45 \\
\textbf{Llama-3.1-Instruct-8B} &
  53.62 &
  15.2 &
  63.35 &
  73.48 &
  5.74 &
  0.72 &
  16.64 &
  1.08 \\
\textbf{o1-mini} &
  76.43 &
  15.8 &
  77.44 &
  87.67 &
  60.83 &
  17.63 &
  45.25 &
  13.98 \\
\textbf{Qwen2.5-Instruct-72B} &
  74.05 &
  14.9 &
  77.8 &
  90.22 &
  63.8 &
  24.1 &
  32.32 &
  8.6 \\
\textbf{Qwen2.5-Instruct-7B} &
  42.01 &
  5.22 &
  50.48 &
  67.05 &
  31.88 &
  1.08 &
  7.43 &
  0 \\ \Xhline{1.5pt}
\end{tabular}
\caption{Overall performance of 10 LLMs on OpenHuEval. The first, second, and third place in each metric are marked with {\color[HTML]{CB0000}red}, {\color[HTML]{009901}green}, and {\color[HTML]{3166FF}blue} text, respectively. In the FIB task evaluation metric, \textbf{B} represents the blank level, and \textbf{Q} represents the question level.}
\label{tab:main_result}
\end{table*}
The overall performance of all LLMs on OpenHuEval is presented in Table \ref{tab:main_result}.
It can be observed that across a total of five tasks, Deepseek-R1 ranks first in three tasks and achieves top-tier performance in the other two tasks. GPT-4o ranks first in two tasks and second in the remaining three tasks. These results demonstrate the exceptional performance of the two models in Hungary-specific tasks.

\textbf{Open-source models vs Closed-source models:} Among open-source models, Deepseek-R1 stands out, while Deepseek-V3 also demonstrates strong overall performance, ranking highly across all tasks. Llama-3.1-Instruct-70B achieved impressive scores of 93.83\% in the HuProverbRea-2CQ task and 36\% in the HuSimpleQA task, ranking second only to the closed-source model GPT-4o. This highlights the growing potential of open-source models, led by Deepseek-R1, which are increasingly showing capabilities comparable to closed-source models in Hungarian language tasks.

\textbf{Traditional LLMs vs. Large Reasoning Models:}
We compared Traditional LLMs and LRMs within the same series. Across five tasks, Deepseek-R1 consistently outperforms Deepseek-V3 in four of them. Specifically, in the HuMatchingFIB task, Deepseek-R1 achieves relative improvements of 12\% at the blank level and 7.19\% at the question level compared to Deepseek-V3. Similarly, for the HuStandardFIB task, it achieves gains of 10.32\% (blank level) and 7.52\% (question level). Although Deepseek-R1 performs slightly worse than Deepseek-V3 on the HuProverbRea task, the performance gap is less than 1\%. Considering that both Deepseek-R1 and Deepseek-V3 are based on the same pretrained model, the significantly stronger performance of Deepseek-R1 on the OpenHuEval benchmark demonstrates the effectiveness of LRMs architectures in Hungarian language tasks and domain-specific scenarios. This result underscores the potential of LRMs as a key avenue of exploration in advancing Artificial General Intelligence (AGI).

\textbf{Model size: }From the results, models with larger parameter sizes perform better on OpenHuEval. For example, GPT-4o, Llama-3.1-Instruct-70B, and Qwen2.5-Instruct-72B outperform their smaller counterparts in the same series (GPT-4o-mini, Llama-3.1-Instruct-7B, and Qwen2.5-Instruct-7B) across all tasks.


\begin{figure}[!t]
    \centering
    \includegraphics[width=1\linewidth]{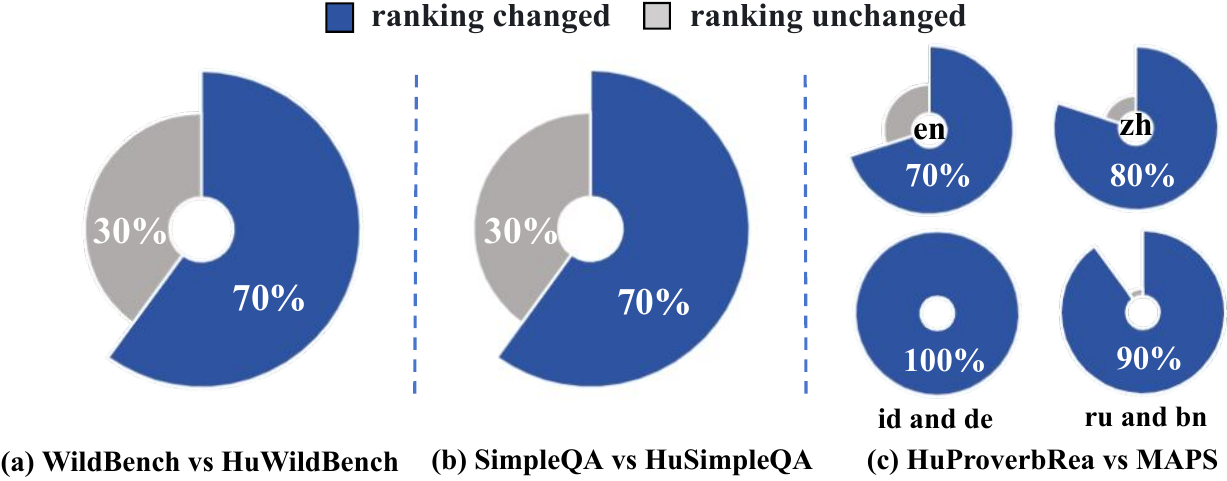}
    \caption{Comparison of model performance on OpenHuEval and similar datasets, highlighting that most LLMs experience rank changes.}
    \label{fig:pies_of_ranking_changed}
\end{figure}

\subsection{Comparison with Existing Benchmarks}

We compare the LLM's performance rankings on the datasets in OpenHuEval and the existing similar datasets:

\textbf{HuWildBench vs WildBench~\cite{WildBench}:} As shown in Figure~\ref{fig:pies_of_ranking_changed}(a), 70\% LLMs experienced ranking changes. Detailed results can be found in Table ~\ref{tab:WildBench_vs_HuWildBench}.

\textbf{HuSimpleQA vs SimpleQA~\cite{wei2024simpleQA}:} As shown in Figure~\ref{fig:pies_of_ranking_changed}(b), 70\% LLMs experienced ranking changes. Detailed results can be found in Table~\ref{tab:SimpleQA_vs_HuSimpleQA}.

\textbf{HuProverbRea vs MAPS:} 
HuProverbRea was constructed with reference to MAPS~\cite{MAPS}, which is the proverb reasoning dataset comprising six subsets, each corresponding to a different language: English (en), German (de), Russian (ru), Bengali (bn), Mandarin Chinese (zh), and Indonesian (id).
We compared the model performance rankings on HuProverbRea and each subset of MAPS.
As shown in Figure~\ref{fig:pies_of_ranking_changed}(c), the percentage of LLMs with ranking changes were: 70\% for en, 80\% for zh, 90\% for ru \& bn, and 100\% for id \& de. Detailed results can be found in Table~\ref{tab:HuProverbRea_vs_MAPS_multilingul_1} and Table~\ref{tab:HuProverbRea_vs_MAPS_multilingul_2}.

These results underscore the importance of evaluating LLMs on Hungarian proverbs and Hungarian-specific questions, highlighting the need for targeted optimization of models to better handle language-specific proverbs and cultural nuances across diverse languages.

\section{Framework for Analyzing the Thinking Process of LRM}

When responding to the user's query, the LRM's response typically consists of two parts: the \textbf{thinking process} and the \textbf{answer}.\footnote{In DeepSeek-V1, the thinking process and answer are enclosed within <think> </think> and <answer> </answer> tags, respectively.}
We developped the framework for the in-depth analysis of the LRM's thinking process on OpenHuEval. For LRM, we chose Deepseek-R1 and QwQ, as these are the only two models with accessible reasoning processes.

\subsection{Task Selection}

Among the OpenHuEval tasks, we selected HuSimpleQA and HuMatchingFIB as our subjects of study.
Unlike recent work \cite{wang2025thoughts}, which focuses solely on math reasoning datasets, the two tasks we selected each have distinctive characteristics:
HuSimpleQA assesses the LLM’s ability to recall and retrieve Hungarian-specific knowledge, as well as its awareness of its own knowledge boundaries.
HuMatchingFIB involves questions where multiple competitive blanks exist within the same problem, requiring the model to carefully choose which answers to fill in.
Therefore, analyzing these tasks allows us to explore the reasoning mechanisms of LRMs in both broader and deeper contexts, providing the research community with more valuable conclusions and insights.

\subsection{Analysis on HuSimpleQA}

\begin{figure}[!t]
    \centering
    \includegraphics[width=1\linewidth]{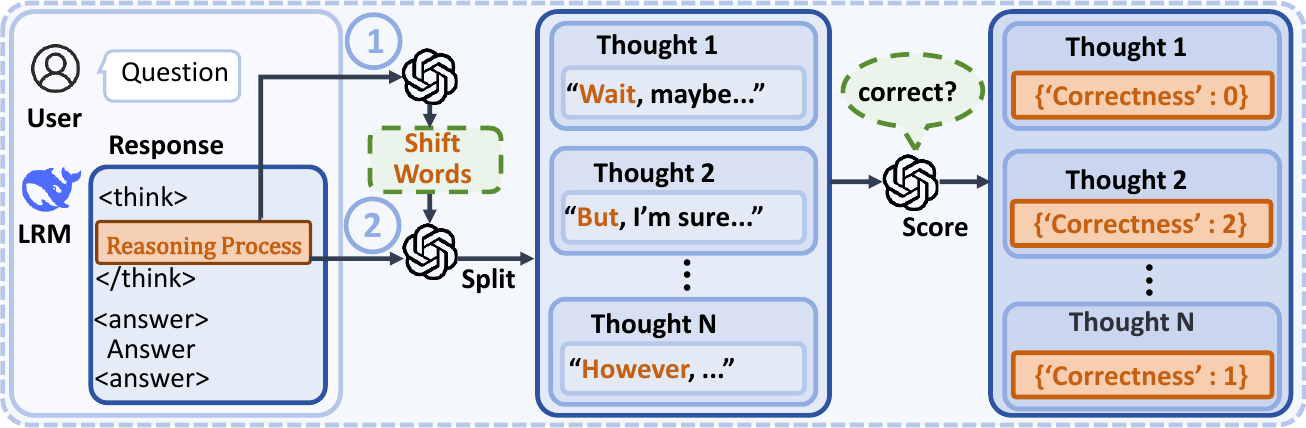}
    \caption{Method for analyzing the LRM's thinking process on HuSimpleQA.}
    \label{fig:simpleQA}
\end{figure}

\textbf{Method:} Following \cite{wang2025thoughts}, the LRM's thinking process can be broken down into ``thoughts''. 
A ``thought'' refers to the intermediate cognitive step output by a LRM during its thinking process. Throughout the thinking process, the LRM transitions between multiple thoughts, which are typically separated by reflective phrases such as ``Alternative'', ``Várni''(wait). The examples of the thoughts and the transitions can be found in Figure \ref{fig:example of husimpleqa thought segmentation in ds} and Figure \ref{fig:example of husimpleqa thought segmentation in qwq}.
Thoughts can be further classified as ``correct'' or ``incorrect'': reasoning along correct thoughts leads to \texttt{CORRECT} responses, while incorrect thoughts result in \texttt{INCORRECT}.

The LRM's responses to HuSimpleQA have been judged as \texttt{CORRECT}, \texttt{INCORRECT}, or \texttt{NOT\_ATTEMPTED}. Then we used GPT-4o to split the thinking process into thoughts (see the prompt in Figure \ref{fig:husimpleqa_shiftword_step1} and Figure \ref{fig:husimpleqa_shiftword_step2}). We evaluated the correctness of each thought (see the prompt in Figure \ref{fig:husimpleqa_thought_evaluation}), with examples provided Figure \ref{fig:example of husimpleqa thought segmentation in ds} and Figure \ref{fig:example of husimpleqa thought segmentation in qwq} in Appendix \ref{Appendix:LRM_reasoning_process}.

\begin{figure}[!t]
    \centering
    \includegraphics[width=1\linewidth]{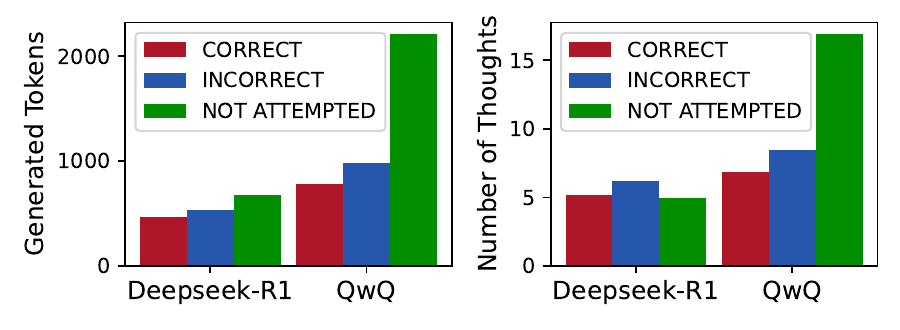}
    \caption{Average number of tokens and thoughts generated per response on HuSimpleQA.}
    \label{fig:analysis_on_husimpleqa_token_thoughts}
\end{figure}

\textbf{Efficiency of thinking process:} We measure the length of the process (in terms of token count) and the number of thoughts under the three evaluation outcomes of the responses, as shown in Figure \ref{fig:analysis_on_husimpleqa_token_thoughts}. The results indicate that both the reasoning length (in tokens) and the thought count were generally shorter for Deepseek-R1 compared to QwQ. Considering that Deepseek-R1 performs better than QwQ on the HuSimpleQA task, it suggests that Deepseek-R1 achieves its superior performance with relatively lower reasoning overhead.

\textbf{Confidence in thinking process:} For Deepseek-R1, the reasoning length and thought count showed no significant differences across the three types of evaluation outcomes. In contrast, for QwQ, the length and the number of thoughts were significantly higher in the \texttt{NOT\_ATTEMPTED} cases compared to the other two types. This observation suggests that, compared to Deepseek-R1, QwQ is less ``confident'', which tends to repeatedly attempt generating answers when faced with uncertainty and is more inclined to abstain from answering altogether.

\begin{figure}[!t]
    \centering
    \includegraphics[width=1\linewidth]{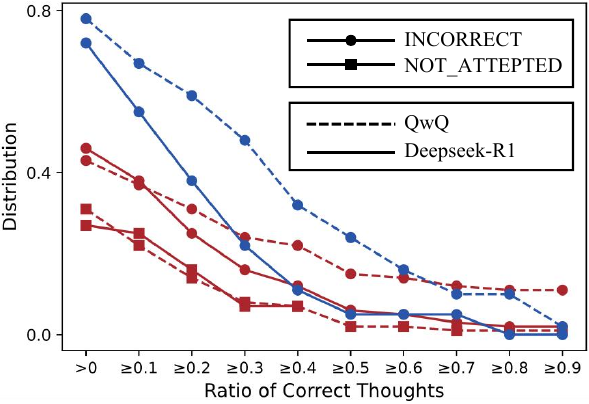}
    \caption{The distribution of thought correctness ratio in \texttt{INCORRECT} and \texttt{NOT\_ATTEPTED} responses. \textcolor{red}{Red} lines for HuSimpleQA and \textcolor{blue}{blue} lines for MATH500-Hard from \cite{wang2025thoughts}.}
    \label{fig:analysis_on_husimpleqa_ratio}
\end{figure}

\textbf{Correct thoughts in \texttt{INCORRECT} responses:} 
In HuSimpleQA, we analyzed the ratio of \texttt{INCORRECT} and \texttt{NOT\_ATTEMPTED} responses from  Deepseek-R1 and QwQ that contained correct thoughts. We compared these results with those from MATH500-Hard as reported in \cite{wang2025thoughts}, as shown in Figure~\ref{fig:analysis_on_husimpleqa_ratio}. In the math reasoning task (MATH500-Hard), the significant portion of LRM's ultimately incorrect responses still included correct thoughts: 72\% for Deepseek-R1 and 78\% for QwQ contained at least one correct thought. However, in HuSimpleQA, the ratio of \texttt{INCORRECT} responses containing correct thoughts dropped to 46\% for Deepseek-R1 and 42\% for QwQ.
These findings suggest that the reasoning processes of LRMs differ significantly between memory-intensive tasks (HuSimpleQA) and reasoning-intensive tasks (MATH500-Hard), highlighting the need for targeted analysis and optimization.

\begin{figure}[!t]
    \centering
    \includegraphics[width=1\linewidth]{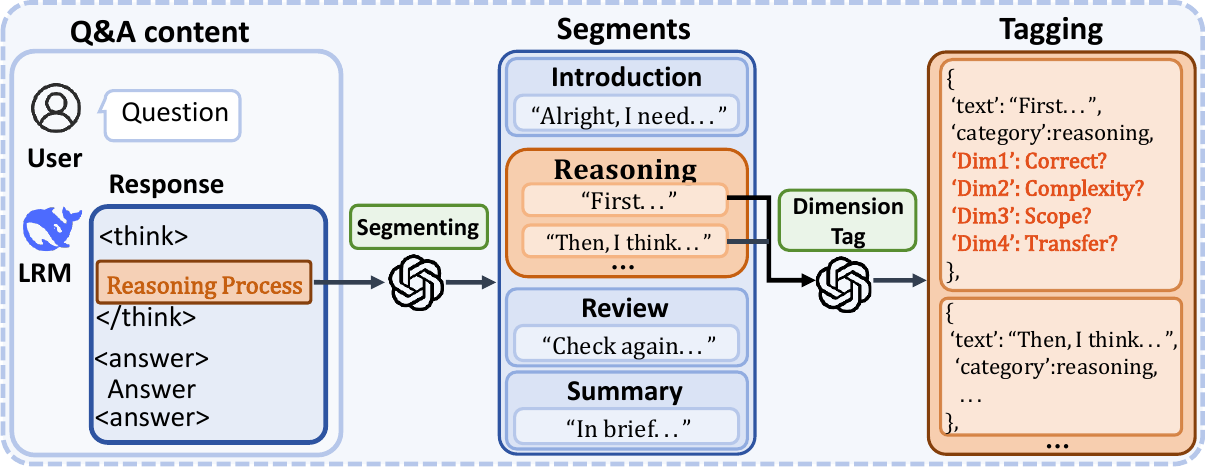}
    \caption{Method for analyzing the LRM's thinking process on HuMatchingFIB.}
    \label{fig:analysis_HuMatchingFIB_overview}
\end{figure}

\begin{figure}[!ht]
    \centering
    \includegraphics[width=1\linewidth]{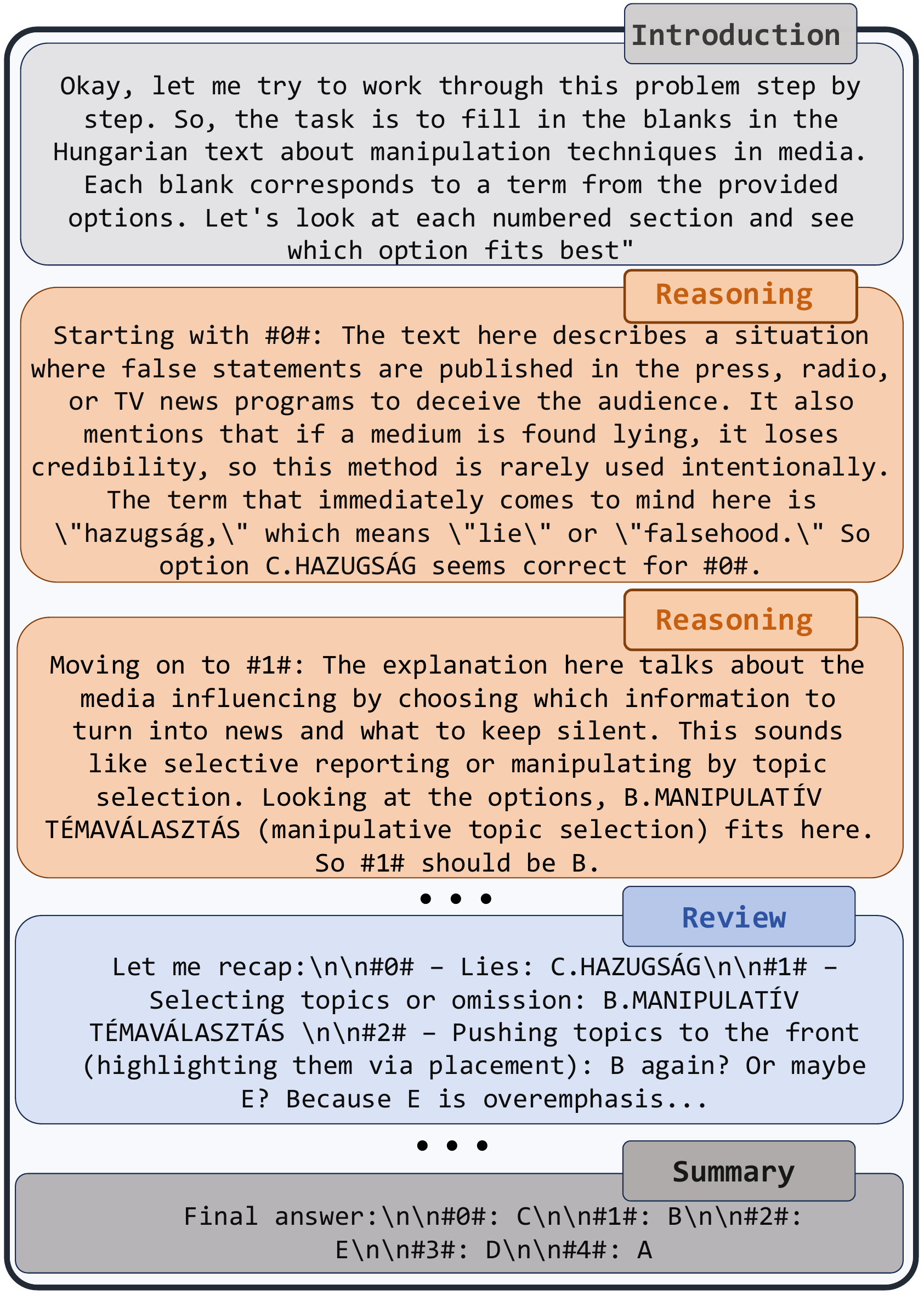}
    \caption{Example for splitting DeepSeek-R1's thinking process into segments and categorizing these segments on HuMatchingFIB.}
    \label{fig:Chapter4_HuMatchingFIB_deepseekR1_segment_sample}
\end{figure}

\subsection{Analysis on HuMatchingFIB}

\textbf{Method:}
Unlike HuSimpleQA where each query contains only one question, HuMatchingFIB involves \textbf{multiple competitive blanks within the same question} that need to be filled (see the example in Figure~\ref{fig:all_task_example} and Figure~\ref{fig:Appendix_HuMatchingFIB_example}). 
Based on extensive case studies, we have developed the analytical method to facilitate further in-depth analysis, as illustrated in Figure~\ref{fig:analysis_HuMatchingFIB_overview}.

Initially, we split the thinking process of LRMs into several segments, categorizing each as \texttt{introduction}, \texttt{reasoning}, \texttt{review}, or \texttt{summary}. Typically, each thinking process begins with an \texttt{introduction} segment, includes several \texttt{reasoning} segments and some \texttt{review} segments in the middle, and concludes with a \texttt{summary} segment. In each \texttt{reasoning} segment, the LRM addresses one or more blanks through reasoning. During the \texttt{review} segment, there is usually a reflection on the completed blanks, with potential revisions to the answers for earlier blanks.
Examples of these segments can be found in Figure~\ref{fig:Chapter4_HuMatchingFIB_deepseekR1_segment_sample}, and detailed definitions are provided in Table \ref{tab:thinking_process_segments_cates_def}.

We further tagged the \texttt{reasoning} segments according to the following four dimensions: \textbf{(Dim1) correctness}: Are the answers in this \texttt{reasoning} segment correct? \textbf{(Dim2) complexity}: In this \texttt{reasoning} segment, does the LRM simply assert the answer, or does it involve more complex reasoning? \textbf{(Dim3) scope}: Does this \texttt{reasoning} segment focuses on a single blank, modifies previous blanks, or addresses multiple blanks? \textbf{(Dim4) language transfer}: Does the LRM switch languages within this \texttt{reasoning} segment? 
The details of the tagging can be found in Appendix~\ref{Appendix:LRM_reasoning_process_HuMatchingFIB_tag_dim}.
Examples of the tagging results can be found in Figure~\ref{fig:humatchingfib_segment_example_deepseek} and Figure~\ref{fig:humatchingfib_segment_example_qwq}.

\begin{figure}[!t]
    \centering
    \includegraphics[width=0.8\linewidth]{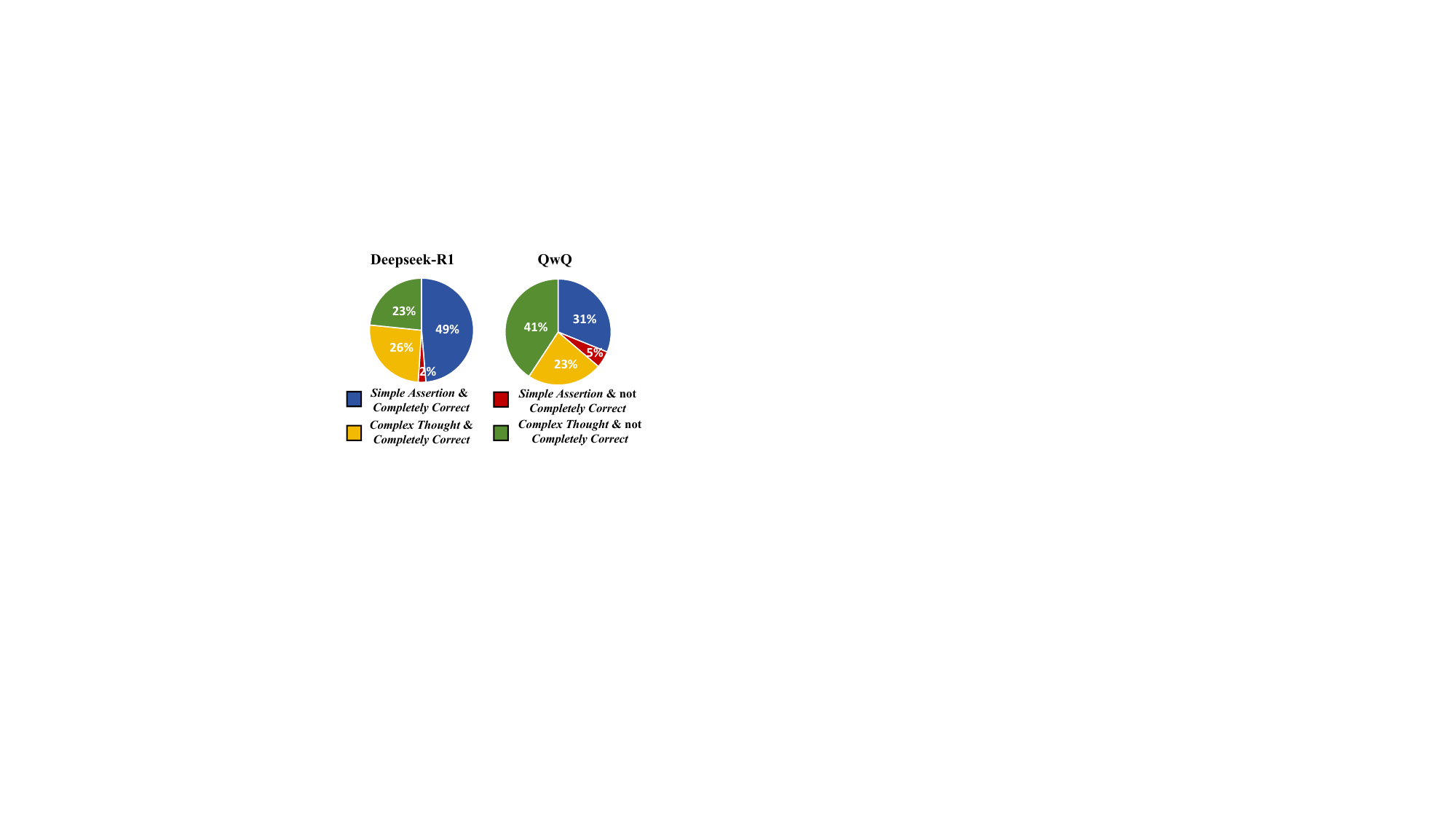}
    \caption{Tagging results of LRM's \texttt{reasoning} segments in the correctness and complexity dimensions.}
    \label{fig:Chapter4_Assertion_Thought}
\end{figure}

\textbf{\textit{Simple Assertion} or \textit{Complex Thought}?}
As shown in Table~\ref{tab:HuMatchingFIB_tagging_dimensions}, the \texttt{reasoning} segments can be categorized into two classes. The first is referred to as \textit{Simple Assertion}, where LRM directly provides the answer to the blank. The second type is termed \textit{Complex Thought}, where the segment involves repeated thinking, logical reasoning, hypothesis validation, or other complex processes. Examples can be found in Figure~\ref{fig:humatchingfib_segment_example_deepseek} and~\ref{fig:humatchingfib_segment_example_qwq}.

We analyzed the correctness of the \texttt{reasoning} segments for both \textit{Simple Assertion} and \textit{Complex Thought} (tagging results of dim1), as shown in Figure~\ref{fig:Chapter4_Assertion_Thought}. By comparing the statistical distributions of the Deepseek-R1 and QwQ models, we noted the following:
Firstly, the proportion of \textit{Simple Assertion} segments that are not \textit{Completely Correct} is quite low (3\% for Deepseek-R1 and 6\% for QwQ), indicating that both models achieve high accuracy when resolving blanks through \textit{Simple Assertion}. This suggests that the models' calibration is relatively reliable, implying that the models ``know what they know''.
Secondly, the proportion of \textit{Simple Assertion} segments that are \textit{Completely Correct} is significantly higher for Deepseek-R1 at 49\%, compared to only 31\% for QwQ. This difference reflects the performance disparity between the two models in the thinking processes.
Thirdly, the correctness for \textit{Complex Thought} is notably lower than for \textit{Simple Assertion}, and both models show a higher proportion of cases where no conclusion is reached for the \textit{Complex Thought} tag. This indicates that when faced with unfamiliar problems, the LRM could actively extend the reasoning and analysis process.

\textbf{Explicit Translation Insertion (ETI):} We observed that in some \texttt{reasoning} segments, when faced with the problem in Hungarian, the LRM first translates the key phrase of the original question into English and then proceeds with analysis and reasoning based on this translation. For example, \textit{``... Erőteljes \#3\# és a költői \#4\# gazdag használata jellemzi. \textbf{This translates to} "It is characterized by strong \#3\# and rich use of poetic \#4\#." ...''}. We refer to this phenomenon as Explicit Translation Insertion (ETI). Statistical analysis shows that ETI occurs in 42.5\% of DeepSeek-R1's \texttt{reasoning} segments, while for QwQ, the proportion is 31.9\%.
This demonstrates that the LRM can adaptively translate input from the language in which it is relatively less proficient (such as Hungarian) into the more proficient language (such as English) during its thinking process. By scaling up the length of reasoning during the test stage, the LLM can partially overcome the cross-language performance discrepancies typical of traditional instruction-based LLM.

\section{Conclusion}

In this paper, we constructed the first benchmark for LLMs focusing on the Hungarian language and its specifics. The results highlight the significant need for evaluation and model optimization tailored to Hungarian language and specifics.
We also developed the framework for analyzing the thinking processes of the cutting-edge LRMs. Our work not only advances LLM technology in Hungarian but also provides valuable insights for studying languages of other countries and regions.

\section{Limitation}
Given the rapid advancements in English evaluation datasets, this research represents only an initial step towards bridging the gap between Hungarian and English evaluation resources. Currently, evaluation datasets for less-resourced languages like Hungarian still lag behind their English counterparts in terms of depth and breadth. Moving forward, we plan to closely monitor developments in English evaluation methodologies, continually refining and enhancing evaluation techniques and datasets for low-resource languages to reduce this disparity.

Moreover, as the field of LLMs evolves rapidly, many promising models, especially those tailored for low-resource languages, remain under-evaluated. Our future goal is to establish a vibrant OpenHuEval community that will regularly update evaluation results for the latest models. This will ensure comprehensive and up-to-date assessments, fostering the optimization and development of models in the low-resource language domain.

\section{Ethical Consideration}
This work involved human annotation. For all annotators, we explicitly informed them about the use of the data and required them to ensure that the questions included in OpenHuEval do not involve any social bias, ethical issues or privacy concerns during the annotation process.

\section*{Acknowledgments}
This research was supported by National Key R\&D Program of China (2022ZD0160201), and Shanghai Artificial Intelligence Laboratory. 

\bibliography{custom}

\begin{thebibliography}{43}
\providecommand{\natexlab}[1]{#1}

\bibitem[{Adelani et~al.(2023)Adelani, Liu, Shen, Vassilyev, Alabi, Mao, Gao, and Lee}]{adelani2023sib}
David~Ifeoluwa Adelani, Hannah Liu, Xiaoyu Shen, Nikita Vassilyev, Jesujoba~O Alabi, Yanke Mao, Haonan Gao, and Annie En-Shiun Lee. 2023.
\newblock Sib-200: A simple, inclusive, and big evaluation dataset for topic classification in 200+ languages and dialects.
\newblock \emph{arXiv preprint arXiv:2309.07445}.

\bibitem[{Ahuja et~al.(2023)Ahuja, Diddee, Hada, Ochieng, Ramesh, Jain, Nambi, Ganu, Segal, Axmed et~al.}]{ahuja2023mega}
Kabir Ahuja, Harshita Diddee, Rishav Hada, Millicent Ochieng, Krithika Ramesh, Prachi Jain, Akshay Nambi, Tanuja Ganu, Sameer Segal, Maxamed Axmed, et~al. 2023.
\newblock Mega: Multilingual evaluation of generative ai.
\newblock \emph{arXiv preprint arXiv:2303.12528}.

\bibitem[{Arora et~al.(2024)Arora, Karpinska, Chen, Bhattacharjee, Iyyer, and Choi}]{arora2024calmqa}
Shane Arora, Marzena Karpinska, Hung-Ting Chen, Ipsita Bhattacharjee, Mohit Iyyer, and Eunsol Choi. 2024.
\newblock Calmqa: Exploring culturally specific long-form question answering across 23 languages.
\newblock \emph{arXiv preprint arXiv:2406.17761}.

\bibitem[{Bandarkar et~al.(2023)Bandarkar, Liang, Muller, Artetxe, Shukla, Husa, Goyal, Krishnan, Zettlemoyer, and Khabsa}]{bandarkar2023belebele}
Lucas Bandarkar, Davis Liang, Benjamin Muller, Mikel Artetxe, Satya~Narayan Shukla, Donald Husa, Naman Goyal, Abhinandan Krishnan, Luke Zettlemoyer, and Madian Khabsa. 2023.
\newblock The belebele benchmark: a parallel reading comprehension dataset in 122 language variants.
\newblock \emph{arXiv preprint arXiv:2308.16884}.

\bibitem[{Clark et~al.(2018)Clark, Cowhey, Etzioni, Khot, Sabharwal, Schoenick, and Tafjord}]{clark2018arc}
Peter Clark, Isaac Cowhey, Oren Etzioni, Tushar Khot, Ashish Sabharwal, Carissa Schoenick, and Oyvind Tafjord. 2018.
\newblock Think you have solved question answering? try arc, the ai2 reasoning challenge.
\newblock \emph{arXiv preprint arXiv:1803.05457}.

\bibitem[{Dubey et~al.(2024)Dubey, Jauhri, Pandey, Kadian, Al-Dahle, Letman, Mathur, Schelten, Yang, Fan et~al.}]{llama3}
Abhimanyu Dubey, Abhinav Jauhri, Abhinav Pandey, Abhishek Kadian, Ahmad Al-Dahle, Aiesha Letman, Akhil Mathur, Alan Schelten, Amy Yang, Angela Fan, et~al. 2024.
\newblock The llama 3 herd of models.
\newblock \emph{arXiv preprint arXiv:2407.21783}.

\bibitem[{Fung et~al.(2024)Fung, Zhao, Doo, Sun, and Ji}]{CultureAtlas}
Yi~Fung, Ruining Zhao, Jae Doo, Chenkai Sun, and Heng Ji. 2024.
\newblock Massively multi-cultural knowledge acquisition \& lm benchmarking.
\newblock \emph{arXiv preprint arXiv:2402.09369}.

\bibitem[{He et~al.(2024)He, Li, Liu, Tan, Wang, Huang, Bu, Guo, Hu, Zheng et~al.}]{he2024chinesesimpleqa}
Yancheng He, Shilong Li, Jiaheng Liu, Yingshui Tan, Weixun Wang, Hui Huang, Xingyuan Bu, Hangyu Guo, Chengwei Hu, Boren Zheng, et~al. 2024.
\newblock Chinese simpleqa: A chinese factuality evaluation for large language models.
\newblock \emph{arXiv preprint arXiv:2411.07140}.

\bibitem[{Hendrycks et~al.(2020)Hendrycks, Burns, Basart, Zou, Mazeika, Song, and Steinhardt}]{mmlu}
Dan Hendrycks, Collin Burns, Steven Basart, Andy Zou, Mantas Mazeika, Dawn Song, and Jacob Steinhardt. 2020.
\newblock Measuring massive multitask language understanding.
\newblock \emph{arXiv preprint arXiv:2009.03300}.

\bibitem[{Hu et~al.(2020)Hu, Ruder, Siddhant, Neubig, Firat, and Johnson}]{hu2020xtreme}
Junjie Hu, Sebastian Ruder, Aditya Siddhant, Graham Neubig, Orhan Firat, and Melvin Johnson. 2020.
\newblock Xtreme: A massively multilingual multi-task benchmark for evaluating cross-lingual generalisation.
\newblock In \emph{International Conference on Machine Learning}, pages 4411--4421. PMLR.

\bibitem[{Huang et~al.(2024)Huang, Mo, Li, Li, Zhang, Yi, Mao, Liu, Xu, Xu et~al.}]{benchmarksurvey}
Kaiyu Huang, Fengran Mo, Hongliang Li, You Li, Yuanchi Zhang, Weijian Yi, Yulong Mao, Jinchen Liu, Yuzhuang Xu, Jinan Xu, et~al. 2024.
\newblock A survey on large language models with multilingualism: Recent advances and new frontiers.
\newblock \emph{arXiv preprint arXiv:2405.10936}.

\bibitem[{Huang et~al.(2025)Huang, Zhu, Hu, He, Li, Huang, and Yuan}]{benchmax}
Xu~Huang, Wenhao Zhu, Hanxu Hu, Conghui He, Lei Li, Shujian Huang, and Fei Yuan. 2025.
\newblock \href {https://arxiv.org/abs/2502.07346} {Benchmax: A comprehensive multilingual evaluation suite for large language models}.
\newblock \emph{Preprint}, arXiv:2502.07346.

\bibitem[{Hurst et~al.(2024)Hurst, Lerer, Goucher, Perelman, Ramesh, Clark, Ostrow, Welihinda, Hayes, Radford et~al.}]{gpt4o}
Aaron Hurst, Adam Lerer, Adam~P Goucher, Adam Perelman, Aditya Ramesh, Aidan Clark, AJ~Ostrow, Akila Welihinda, Alan Hayes, Alec Radford, et~al. 2024.
\newblock Gpt-4o system card.
\newblock \emph{arXiv preprint arXiv:2410.21276}.

\bibitem[{Jaech et~al.(2024)Jaech, Kalai, Lerer, Richardson, El-Kishky, Low, Helyar, Madry, Beutel, Carney et~al.}]{openai-o1}
Aaron Jaech, Adam Kalai, Adam Lerer, Adam Richardson, Ahmed El-Kishky, Aiden Low, Alec Helyar, Aleksander Madry, Alex Beutel, Alex Carney, et~al. 2024.
\newblock Openai o1 system card.
\newblock \emph{arXiv preprint arXiv:2412.16720}.

\bibitem[{Koto et~al.(2023)Koto, Aisyah, Li, and Baldwin}]{indommlu}
Fajri Koto, Nurul Aisyah, Haonan Li, and Timothy Baldwin. 2023.
\newblock \href {https://doi.org/10.18653/v1/2023.emnlp-main.760} {Large language models only pass primary school exams in {I}ndonesia: A comprehensive test on {I}ndo{MMLU}}.
\newblock In \emph{Proceedings of the 2023 Conference on Empirical Methods in Natural Language Processing}, pages 12359--12374, Singapore. Association for Computational Linguistics.

\bibitem[{Koto et~al.(2024)Koto, Li, Shatnawi, Doughman, Sadallah, Alraeesi, Almubarak, Alyafeai, Sengupta, Shehata et~al.}]{koto2024arabicmmlu}
Fajri Koto, Haonan Li, Sara Shatnawi, Jad Doughman, Abdelrahman~Boda Sadallah, Aisha Alraeesi, Khalid Almubarak, Zaid Alyafeai, Neha Sengupta, Shady Shehata, et~al. 2024.
\newblock Arabicmmlu: Assessing massive multitask language understanding in arabic.
\newblock \emph{arXiv preprint arXiv:2402.12840}.

\bibitem[{Lai et~al.(2023)Lai, Van~Nguyen, Ngo, Nguyen, Dernoncourt, Rossi, and Nguyen}]{lai2023okapi}
Viet~Dac Lai, Chien Van~Nguyen, Nghia~Trung Ngo, Thuat Nguyen, Franck Dernoncourt, Ryan~A Rossi, and Thien~Huu Nguyen. 2023.
\newblock Okapi: Instruction-tuned large language models in multiple languages with reinforcement learning from human feedback.
\newblock \emph{arXiv preprint arXiv:2307.16039}.

\bibitem[{Li et~al.(2024{\natexlab{a}})Li, Dong, Chen, Su, Zhou, Ai, Ye, and Liu}]{llm_as_judge_survey}
Haitao Li, Qian Dong, Junjie Chen, Huixue Su, Yujia Zhou, Qingyao Ai, Ziyi Ye, and Yiqun Liu. 2024{\natexlab{a}}.
\newblock Llms-as-judges: a comprehensive survey on llm-based evaluation methods.
\newblock \emph{arXiv preprint arXiv:2412.05579}.

\bibitem[{Li et~al.(2023)Li, Zhang, Koto, Yang, Zhao, Gong, Duan, and Baldwin}]{li2023cmmlu}
Haonan Li, Yixuan Zhang, Fajri Koto, Yifei Yang, Hai Zhao, Yeyun Gong, Nan Duan, and Timothy Baldwin. 2023.
\newblock Cmmlu: Measuring massive multitask language understanding in chinese.
\newblock \emph{arXiv preprint arXiv:2306.09212}.

\bibitem[{Li et~al.(2024{\natexlab{b}})Li, Wang, Hu, and Jiang}]{CUNIT}
Jialin Li, Junli Wang, Junjie Hu, and Ming Jiang. 2024{\natexlab{b}}.
\newblock How well do llms identify cultural unity in diversity?
\newblock \emph{arXiv preprint arXiv:2408.05102}.

\bibitem[{Ligeti-Nagy et~al.(2024)Ligeti-Nagy, Ferenczi, H{\'e}ja, Laki, Vad{\'a}sz, Yang, and V{\'a}radi}]{hulu}
No{\'e}mi Ligeti-Nagy, Gerg{\H{o}} Ferenczi, Enik{\H{o}} H{\'e}ja, L{\'a}szl{\'o}~J{\'a}nos Laki, No{\'e}mi Vad{\'a}sz, Zijian~Gy{\H{o}}z{\H{o}} Yang, and Tam{\'a}s V{\'a}radi. 2024.
\newblock Hulu: Hungarian language understanding benchmark kit.
\newblock In \emph{Proceedings of the 2024 Joint International Conference on Computational Linguistics, Language Resources and Evaluation (LREC-COLING 2024)}, pages 8360--8371.

\bibitem[{Lin et~al.(2024)Lin, Deng, Chandu, Brahman, Ravichander, Pyatkin, Dziri, Bras, and Choi}]{WildBench}
Bill~Yuchen Lin, Yuntian Deng, Khyathi Chandu, Faeze Brahman, Abhilasha Ravichander, Valentina Pyatkin, Nouha Dziri, Ronan~Le Bras, and Yejin Choi. 2024.
\newblock Wildbench: Benchmarking llms with challenging tasks from real users in the wild.
\newblock \emph{arXiv preprint arXiv:2406.04770}.

\bibitem[{Lin et~al.(2021)Lin, Hilton, and Evans}]{lin2021truthfulqa}
Stephanie Lin, Jacob Hilton, and Owain Evans. 2021.
\newblock Truthfulqa: Measuring how models mimic human falsehoods.
\newblock \emph{arXiv preprint arXiv:2109.07958}.

\bibitem[{Liu et~al.(2024{\natexlab{a}})Liu, Feng, Xue, Wang, Wu, Lu, Zhao, Deng, Zhang, Ruan et~al.}]{deepseekv3}
Aixin Liu, Bei Feng, Bing Xue, Bingxuan Wang, Bochao Wu, Chengda Lu, Chenggang Zhao, Chengqi Deng, Chenyu Zhang, Chong Ruan, et~al. 2024{\natexlab{a}}.
\newblock Deepseek-v3 technical report.
\newblock \emph{arXiv preprint arXiv:2412.19437}.

\bibitem[{Liu et~al.(2024{\natexlab{b}})Liu, Gurevych, and Korhonen}]{liu2024culturally}
Chen~Cecilia Liu, Iryna Gurevych, and Anna Korhonen. 2024{\natexlab{b}}.
\newblock Culturally aware and adapted nlp: A taxonomy and a survey of the state of the art.
\newblock \emph{arXiv preprint arXiv:2406.03930}.

\bibitem[{Liu et~al.(2024{\natexlab{c}})Liu, Koto, Baldwin, and Gurevych}]{MAPS}
Chen~Cecilia Liu, Fajri Koto, Timothy Baldwin, and Iryna Gurevych. 2024{\natexlab{c}}.
\newblock \href {https://arxiv.org/abs/2309.08591} {Are multilingual llms culturally-diverse reasoners? an investigation into multicultural proverbs and sayings}.
\newblock \emph{Preprint}, arXiv:2309.08591.

\bibitem[{Myung et~al.(2024)Myung, Lee, Zhou, Jin, Putri, Antypas, Borkakoty, Kim, Perez-Almendros, Ayele et~al.}]{myung2024blend}
Junho Myung, Nayeon Lee, Yi~Zhou, Jiho Jin, Rifki Putri, Dimosthenis Antypas, Hsuvas Borkakoty, Eunsu Kim, Carla Perez-Almendros, Abinew~Ali Ayele, et~al. 2024.
\newblock Blend: A benchmark for llms on everyday knowledge in diverse cultures and languages.
\newblock \emph{Advances in Neural Information Processing Systems}, 37:78104--78146.

\bibitem[{Naous et~al.(2023)Naous, Ryan, Ritter, and Xu}]{havingbeer}
Tarek Naous, Michael~J Ryan, Alan Ritter, and Wei Xu. 2023.
\newblock Having beer after prayer? measuring cultural bias in large language models.
\newblock \emph{arXiv preprint arXiv:2305.14456}.

\bibitem[{Nov{\'a}k et~al.(2023)Nov{\'a}k, Nov{\'a}k, Zombori, Szab{\'o}, Sz{\'a}nt{\'o}, and Farkas}]{MILQA}
Attila Nov{\'a}k, Borb{\'a}la Nov{\'a}k, Tam{\'a}s Zombori, Gerg{\H{o}} Szab{\'o}, Zsolt Sz{\'a}nt{\'o}, and Rich{\'a}rd Farkas. 2023.
\newblock A question answering benchmark database for hungarian.
\newblock In \emph{Proceedings of the 17th Linguistic Annotation Workshop (LAW-XVII)}, pages 188--198.

\bibitem[{Osváth et~al.(2023)Osváth, Yang, and Kósa}]{bertopic}
Mátyás Osváth, Zijian~Győző Yang, and Karolina Kósa. 2023.
\newblock Analyzing narratives of patient experiences: A bert topic modeling approach.
\newblock \emph{Acta Polytechnica Hungarica}, 20(7):153--171.

\bibitem[{Rajpurkar et~al.(2016)Rajpurkar, Zhang, Lopyrev, and Liang}]{squad}
Pranav Rajpurkar, Jian Zhang, Konstantin Lopyrev, and Percy Liang. 2016.
\newblock Squad: 100,000+ questions for machine comprehension of text.
\newblock \emph{arXiv preprint arXiv:1606.05250}.

\bibitem[{Sakai et~al.(2024)Sakai, Kamigaito, and Watanabe}]{sakai2024mcsqa}
Yusuke Sakai, Hidetaka Kamigaito, and Taro Watanabe. 2024.
\newblock mcsqa: Multilingual commonsense reasoning dataset with unified creation strategy by language models and humans.
\newblock \emph{arXiv preprint arXiv:2406.04215}.

\bibitem[{Son et~al.(2024)Son, Lee, Kim, Kim, Muennighoff, Choi, Park, Yoo, and Biderman}]{son2024kmmlu}
Guijin Son, Hanwool Lee, Sungdong Kim, Seungone Kim, Niklas Muennighoff, Taekyoon Choi, Cheonbok Park, Kang~Min Yoo, and Stella Biderman. 2024.
\newblock Kmmlu: Measuring massive multitask language understanding in korean.
\newblock \emph{arXiv preprint arXiv:2402.11548}.

\bibitem[{Sun et~al.(2024)Sun, Huang, Wu, Gu, Li, Zhang, Yan, and He}]{charm}
Jiaxing Sun, Weiquan Huang, Jiang Wu, Chenya Gu, Wei Li, Songyang Zhang, Hang Yan, and Conghui He. 2024.
\newblock Benchmarking chinese commonsense reasoning of llms: From chinese-specifics to reasoning-memorization correlations.
\newblock \emph{arXiv preprint arXiv:2403.14112}.

\bibitem[{Team(2024{\natexlab{a}})}]{deepseek-r1}
DeepSeek Team. 2024{\natexlab{a}}.
\newblock Deepseek-r1-lite-preview is now live: unleashing supercharged reasoning power.

\bibitem[{Team(2024{\natexlab{b}})}]{qwq}
Qwen Team. 2024{\natexlab{b}}.
\newblock Qwq: Reflect deeply on the boundaries of the unknown.
\newblock \emph{Hugging Face}.

\bibitem[{Wang et~al.(2025)Wang, Liu, Xu, Liang, Chen, He, Song, Yu, Li, Zhang et~al.}]{wang2025thoughts}
Yue Wang, Qiuzhi Liu, Jiahao Xu, Tian Liang, Xingyu Chen, Zhiwei He, Linfeng Song, Dian Yu, Juntao Li, Zhuosheng Zhang, et~al. 2025.
\newblock Thoughts are all over the place: On the underthinking of o1-like llms.
\newblock \emph{arXiv preprint arXiv:2501.18585}.

\bibitem[{Wei et~al.(2024)Wei, Karina, Chung, Jiao, Papay, Glaese, Schulman, and Fedus}]{wei2024simpleQA}
Jason Wei, Nguyen Karina, Hyung~Won Chung, Yunxin~Joy Jiao, Spencer Papay, Amelia Glaese, John Schulman, and William Fedus. 2024.
\newblock Measuring short-form factuality in large language models.
\newblock \emph{arXiv preprint arXiv:2411.04368}.

\bibitem[{Yang et~al.(2024)Yang, Yang, Zhang, Hui, Zheng, Yu, Li, Liu, Huang, Wei et~al.}]{qwen2.5}
An~Yang, Baosong Yang, Beichen Zhang, Binyuan Hui, Bo~Zheng, Bowen Yu, Chengyuan Li, Dayiheng Liu, Fei Huang, Haoran Wei, et~al. 2024.
\newblock Qwen2. 5 technical report.
\newblock \emph{arXiv preprint arXiv:2412.15115}.

\bibitem[{Yin et~al.(2022)Yin, Bansal, Monajatipoor, Li, and Chang}]{yin2022geomlama}
Da~Yin, Hritik Bansal, Masoud Monajatipoor, Liunian~Harold Li, and Kai-Wei Chang. 2022.
\newblock Geomlama: Geo-diverse commonsense probing on multilingual pre-trained language models.
\newblock \emph{arXiv preprint arXiv:2205.12247}.

\bibitem[{Y{\"u}ksel et~al.(2024)Y{\"u}ksel, K{\"o}ksal, {\c{S}}enel, Korhonen, and Sch{\"u}tze}]{yuksel2024turkishmmlu}
Arda Y{\"u}ksel, Abdullatif K{\"o}ksal, L{\"u}tfi~Kerem {\c{S}}enel, Anna Korhonen, and Hinrich Sch{\"u}tze. 2024.
\newblock Turkishmmlu: Measuring massive multitask language understanding in turkish.
\newblock \emph{arXiv preprint arXiv:2407.12402}.

\bibitem[{Zellers et~al.(2019)Zellers, Holtzman, Bisk, Farhadi, and Choi}]{hellaswag}
Rowan Zellers, Ari Holtzman, Yonatan Bisk, Ali Farhadi, and Yejin Choi. 2019.
\newblock \href {https://doi.org/10.18653/v1/P19-1472} {{H}ella{S}wag: Can a machine really finish your sentence?}
\newblock In \emph{Proceedings of the 57th Annual Meeting of the Association for Computational Linguistics}, pages 4791--4800, Florence, Italy. Association for Computational Linguistics.

\bibitem[{Zhang et~al.(2024)Zhang, Deng, Wan, Yang, Wei, Huang, Yu, Lin, and Zhou}]{PMMEval}
Yidan Zhang, Boyi Deng, Yu~Wan, Baosong Yang, Haoran Wei, Fei Huang, Bowen Yu, Junyang Lin, and Jingren Zhou. 2024.
\newblock P-mmeval: A parallel multilingual multitask benchmark for consistent evaluation of llms.
\newblock \emph{arXiv preprint arXiv:2411.09116}.

\end{thebibliography}

\clearpage
\appendix

\section{Related Works}
\subsection{Hungarian benchmarks for LLMs from Translation}
Directly translating an existing English evaluation dataset into non-English is an effective and straightforward method for constructing non-English evaluation datasets. Many existing multilingual evaluation datasets are constructed using this approach, such as those described in~\cite{ahuja2023mega,benchmarksurvey}. Among these, quite a few evaluation datasets include Hungarian as one of the languages, such as \cite{hu2020xtreme} and \cite{adelani2023sib}.
\cite{lai2023okapi} utilized GPT-3.5 to translate datasets such as ARC \cite{clark2018arc}, MMLU \cite{mmlu}, TruthfulQA \cite{lin2021truthfulqa}, and HellaSwag \cite{hellaswag} into Hungarian.
Belebele \cite{bandarkar2023belebele} includes Hungarian as part of its multilingual parallel corpus reading comprehension evaluation dataset.
BenchMaX \cite{benchmax} is a multilingual parallel corpus evaluation dataset focused on comprehensively assessing LLMs' capabilities across various generative tasks.

These translation-based parallel corpus evaluation datasets provide a comprehensive assessment of LLMs’ language-agnostic capabilities in Hungarian, including areas like world knowledge, mathematical reasoning, logical reasoning, and code generation. However, they overlook the unique characteristics and linguistic features of Hungarian, such as language nuances, culture, history, and regional context, which are critical for Hungarian users.

\subsection{Language specific benchmarks for LLMs}
There has been extensive research focusing on evaluating the unique features and capabilities of languages \cite{liu2024culturally}.
Some studies have constructed benchmarks similar to MMLU \cite{mmlu} by collecting exam questions specific to various countries, such as CMMLU \cite{li2023cmmlu}, IndoMMLU \cite{indommlu}, ArabicMMLU \cite{koto2024arabicmmlu}, KMMLU \cite{son2024kmmlu}, and TurkishMMLU \cite{yuksel2024turkishmmlu}.
Other studies have built evaluation datasets by crawling material and user queries from internet forums, filtering for queries related to linguistic and cultural features, such as the benchmark in \cite{havingbeer} and CaLMQA \cite{arora2024calmqa}.
Some works manually construct evaluation datasets that emphasize linguistic and cultural characteristics, such as BLEnD \cite{myung2024blend}, CHARM \cite{charm}, and HuLU \cite{hulu}.
Others adopt a ``LLM generation combined with human expert modification and review'' method to construct culturally characteristic evaluation datasets, such as MAPS \cite{MAPS}, mCSQA \cite{sakai2024mcsqa}, and ChineseSimpleQA \cite{he2024chinesesimpleqa}.
Although Hungarian is rarely covered in these works, they provide crucial inspiration and approaches for our work.

Among these works, MILQA~\cite{MILQA} is the Hungarian question answering benchmark created mainly following the SQuAD~\cite{squad}. HuLU \cite{hulu} is a comprehensive Hungarian evaluation benchmark kit~\footnote{\url{https://hulu.nytud.hu/}} that includes a total of seven tasks and corresponding datasets. Of the seven tasks, four are constructed by translating existing English evaluation datasets, and three are manually created based on native Hungarian corpora. However, HuLU only supports multiple-choice and true/false questions, which limits its ability to assess broader LLM capabilities such as language generation, open-domain Q\&A, reasoning, and instruction-following.

\section{Hungarian-Specific Dimensions}
OpenHuEval encompasses eight Hungarian-Specific Dimensions (HuSpecificDim), as shown in Table \ref{tab:Hungarian-specific dimensions}: \textit{Language} ($\mathcal{L}$), \textit{History} ($\mathcal{H}$), \textit{Life, Culture, and Customs} ($\mathcal{LCC}$), \textit{Education and Profession} ($\mathcal{EP}$), \textit{Geography and Place} ($\mathcal{GP}$), \textit{Figure} ($\mathcal{F}$), \textit{Politics, Policy, and Law} ($\mathcal{PPL}$), and \textit{Business and Finance} ($\mathcal{BF}$). 

To select these Hungarian-Specific Dimensions, e first conducted extensive literature research to summarize the knowledge dimensions in existing cultural benchmarks, as shown in Table \ref{tab:dimension_chosen}.
\begin{table*}[h]
\centering
\small
\begin{tabular}{lp{10cm}}
\Xhline{1.5pt}
\textbf{Benchmark} & \textbf{Dimension} \\
\hline
\rowcolor[HTML]{EFEFEF} 
CHARM \cite{charm} & History, Traditional Culture and Arts, Daily Life and Customs, Entertainment, Public Figures, Geography, Chinese Language \\
\rowcolor[HTML]{FFFFFF} 
CUNIT \cite{CUNIT} & Clothing, Food \\
\rowcolor[HTML]{EFEFEF} 
BLEnD \cite{myung2024blend} & Food, Sports, Family, Education, Holidays/Celebrations/Leisure, Work-Life \\
\rowcolor[HTML]{FFFFFF} 
CultureAtlas \cite{CultureAtlas} & Culture, Holidays, Dining Etiquette, Education, Honorifics, etc. \\
\rowcolor[HTML]{EFEFEF}
GEOMLAMA \cite{yin2022geomlama} & Habits and Personal Choices, Cultures and Customs, Policies and Regulations, Geography \\
\Xhline{1.5pt}
\end{tabular}
\caption{The knowledge dimensions present in existing cultural benchmarks.}
\label{tab:dimension_chosen}
\end{table*}

Based on above and by considering the knowledge needs of typical Hungarian users and the content on authoritative Hungarian websites, we established the Hu-specific dimensions for OpenHuEval. We believe these dimensions cover most user needs related to Hungarian specifics.

\section{Inference Detail}

\subsection{Default Maximum Output Length for LLM in Opencompass}

For traditional instruction-based LLMs, we adopted OpenCompass's default settings for the maximum output length, as shown in Table \ref{tab:llm_length}.

\begin{table}[]
\begin{tabular}{lc}
\Xhline{1.5pt}
Model                  & Max out length \\ \hline
Qwen2.5-Instruct-7B    & 4096           \\
Qwen2.5-Instruct-72B   & 4096           \\
GPT4-o                 & 2048           \\
GPT-4o-mini            & 2048           \\
Llama-3.1-Instruct-8B  & 1024           \\
Llama-3.1-Instruct-70B & 4096           \\
Deepseek-V3            & 2048           \\ \Xhline{1.5pt}
\end{tabular}
\caption{Default maximum output length for
LLM in OpenCompass.}
\label{tab:llm_length}
\end{table}

\section{HuWildBench}
\label{Appendix:HuWildBench}

\subsection{Overall Construction Pipeline of HuWildBench}

\begin{figure}[!t]
    \centering
    \includegraphics[width=1\linewidth]{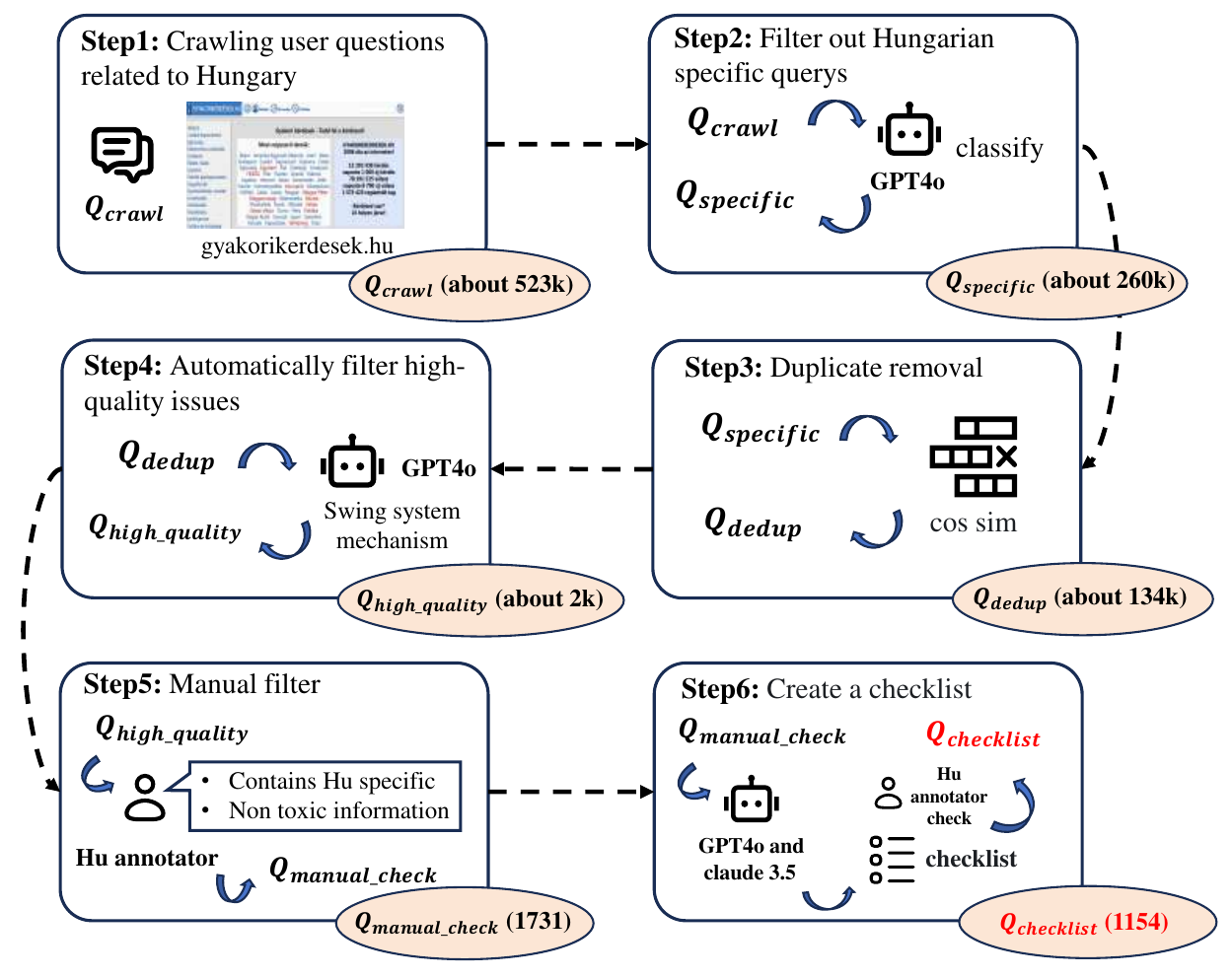}
    \caption{Constuction of HuWildBench.}
    \label{fig:Construction_HuWildBench}
\end{figure}

The construction of HuWildBench contains the following steps(see Figure \ref{fig:Construction_HuWildBench}):

\textbf{(1) Crawling:} All user queries on the g13k website\footnote{\url{https://www.gyakorikerdesek.hu/}} are systematically categorized into a multi-level tag system, which consists of 27 primary tags and 231 secondary tags. We manually reviewed all the secondary tags and selected 37 of them that contain a higher number of questions related to Hungary, such as népszokások (folk customs), egészségügyi-ellátások (healthcare services), and rezsi (overheads). We then crawled user queries under these 37 secondary tags, with a query date range from January 1, 2019, to August 31, 2024, resulting in the dataset $Q_{crawl}$(approximately 523K queries).

\textbf{(2) Filtering for Hungary-specific content:} Although these 37 secondary tags are closely related to Hungary, many of the questions still do not focus on Hungary-specific topics. Therefore, we used GPT-4o to classify the questions in $Q_{crawl}$ (detailed prompt in Figure~\ref{fig:Appendix_HuWildBenchPromptSelectHuSpecialQuestion}), resulting in a subset of approximately 260K questions, $Q_{specific}$

\textbf{(3) Deduplication:} To ensure the diversity of questions, we performed deduplication on the Hungary-related questions within each secondary tag. The detailed process is outlined in Appendix~\ref{Appendix:HuWildBenchDeduplication}. After deduplication, the number of user questions in $Q_{dedup}$ was approximately 134K.

\textbf{(4) Automatic high-quality question filtering:} To ensure that only high-quality questions are extracted from the question pool, we designed a comparative-based high-quality question filtering strategy, as detailed in Appendix~\ref{Appendix:HuWildBenchPromptSelectHQQuestion}. After filtering, the resulting set $Q_{high\_quality}$ contained around 2K questions.

\textbf{(5) Manual filtering:} We hired a group of Hungarian native speakers to further manually review the questions in $Q_{high\_quality}$. Only questions that met the following two criteria were retained: First, the question should be Hungary-specific and closely related to Hungary. Second, the question must be harmless, meaning it does not contain inappropriate content such as pornography, violence, politics, or taboo topics specific to Hungary. The final set $Q_{manual\_check}$ consists of 1731 questions.

\textbf{(6) Checklist construction:} Based on WildBench~\cite{WildBench}, we constructed a checklist for each question. The purpose of the checklist is to assist the LLM judger in evaluating the answers. Each item in the checklist queries a specific aspect of the answer to a question. An example of the checklist can be found in Table~\ref{tab:HuWildBench_QuestionExamples}, and the detailed construction method is provided in Appendix~\ref{Appendix:HuWildBenchGenerateChecklist}. To ensure the relevance of the checklist items to the questions, we hired a Hungarian native speaker to review the checklist for quality, filtering out non-compliant items and performing deduplication. The filtering criterion was whether the item was suitable as an evaluation dimension for the model's response. To ensure the reliability of the LLM-as-judge, we filtered out user questions with fewer than 8 checklist items. The final set $Q_{checklist}$ contains 1154 questions.
In the end, we obtained 1154 user questions along with their corresponding checklists.

\subsection{Deduplication of similar questions}
\label{Appendix:HuWildBenchDeduplication}

Since there are similar questions in the results obtained in the previous step, we design a method to remove similar ones. 
Specifically, we first use the SentenceTransformer~\cite{bertopic}\footnote{\url{https://github.com/UKPLab/sentence-transformers}} model to extract the embedding vector of each question. 
Then, we calculate the cosine similarity between the embedding of each two questions, and choose a threshold between [0.15-0.25] according to the number of questions under each secondary tag.The larger the number of problems, the larger the threshold. 
Finally, one of the questions whose similarity is less than the threshold is removed, ensuring that the similarities between all questions are greater than the threshold.

\subsection{Automatic high-quality question filtering}
\label{Appendix:HuWildBenchPromptSelectHQQuestion}

In order to automate the filtering of high-quality sample pots, we constructed the prompt that allows the GPT-4o to select the two best Hungarian questions out of the five based on the criteria of linguistic complexity, Hungarian relevance, common-sense accuracy, context-dependence, answer diversity, ambiguity, reasoning requirements, socio-ethical considerations, format diversity, and breadth of knowledge and outputs their indexes in JSON format to output their indexes, as shown in Figure~\ref{fig:Appendix_HuWildBenchPromptSelectHQQuestion}.
Specifically, we first set the criteria for high-quality questions in Prompt. Then we ask GPT-4o to compare the input questions based on the criteria. In order to mitigate the occurrence of some high-quality questions being eliminated prematurely (or vice versa) when all the questions in the same batch are of high quality, we follow the following 3 rules when filtering the high-quality questions:
(1) filter 2 high-quality questions from 5 questions at a time, instead of filtering 1 high-quality question directly from 2 questions.
(2) use the Swiss system mechanism instead of the knockout mechanism. In each screening round, each question can win in the current round as long as it ensures that it wins in two comparisons, and it will not be eliminated directly because of a failure in one comparison.
(3) Our question screening strategy eliminates 65\% of the questions in each round, in order to ensure that each secondary label has a sufficient number of high-quality questions. We conducted different elimination rounds for questions under different labels, and finally got about 2K questions.

To ensure the reasonableness and robustness of the standard, we reviewed the results obtained using it. We selected 500 sets of questions filtered by GPT-4o according to this standard (where the model had to select two high-quality questions from each set of five). Human experts annotated these sets, and the overlap rate between the selections of the human experts and the model was 81\% (a set was considered overlapping if the two questions chosen by the model matched those chosen by the experts). This demonstrates the effectiveness of the standard for selecting high-quality questions. The model's recall rate for the questions selected by the human experts was 90\%.


The final constructed HuWildBench is shown in Table~\ref{tab:HuWildBench_QuestionExamples}.

\subsection{Checklist construction}
\label{Appendix:HuWildBenchGenerateChecklist}

In the process of building the Checklist, we mainly use large language models to generate it. In order to ensure the diversity of the Checklist and make the judge model can better evaluate the quality of the answers, here we use two non-open source LLM GPT-4o and Claude-3.5, each model generates a list of length 3-5. then we merge the two Checklists into one final Checklist. Checklists are then merged into a final Checklist. ultimately, each problem has a length of 6-10 and a Checklist. The details of our designed Prompt are shown in Figure~\ref{fig:Appendix_HuWildBenchPromptGenerateChecklist} and 
the final constructed partial Checklist is shown in Table~\ref{tab:HuWildBench_QuestionExamples}.

\subsection{LLM-as-judge}
\label{Appendix:HuWildBenchLLMAsJudge}
\begin{table}[h]
\begin{tabular}{lp{5cm}}
\Xhline{1.5pt}
\rowcolor[HTML]{EFEFEF} 
\textbf{Score} & \textbf{Definition} \\ \midrule
\textbf{Score 1-2}  & The response is very poor and does not make sense at all.              \\ 
\rowcolor[HTML]{EFEFEF} 
\textbf{Score 3-4}  & The response is poor and does not help the user solve the problem meaningfully. \\
\rowcolor[HTML]{FFFFFF} 
\textbf{Score 5-6} & The response is fair but has issues (e.g., factual errors, hallucinations, missing key information). \\
\rowcolor[HTML]{EFEFEF} 
\textbf{Score 7-8}  & The response is good but could be improved.                                     \\
\rowcolor[HTML]{FFFFFF} 
\textbf{Score 9-10} & The response is perfect and provides helpful information to solve the problem. \\\Xhline{1.5pt}

\end{tabular}
\caption{Definition of WB-Scores.}
\label{tab:HuWildBench_socre_definition}
\end{table}

Following WB-Score~\cite{WildBench}, we use GPT-4o as the judge model to evaluate the score of responses generated by LLM. The prompt used in judge is shown in Figure~\ref{fig:Appendix_HuWildBenchPromptScore}.

\subsection{Comparing the Performance Ranking on WildBench and HuWildBench}

We compare the performance rankings of LLM on HuWildBench and WildBench~\cite{WildBench}, as shown in Table ~\ref{tab:WildBench_vs_HuWildBench}.

\begin{table}[t]
\scriptsize
\centering
\begin{tabular}{ccc}
 \Xhline{1.5pt}
\textbf{Rank} & \textbf{WildBench} & \textbf{HuWildBench} \\ \Xhline{1.0pt}
\textbf{1}    & Deepseek-V3            & Deepseek-R1(\textcolor{red}{↑3})      \\
\textbf{2}    & GPT-4o                 & GPT-4o(-)                             \\
\textbf{3}    & o1-mini                & Deepseek-V3(\textcolor{green}{↓2})    \\
\textbf{4}    & Deepseek-R1            & o1-mini(\textcolor{green}{↓1})        \\
\textbf{5}    & Qwen2.5-Instruct-72B   & GPT-4o-mini(\textcolor{red}{↑1})      \\
\textbf{6}    & GPT-4o-mini            & Qwen2.5-Instruct-72B(\textcolor{green}{↓1})   \\
\textbf{7}    & Llama-3.1-Instruct-70B & Llama-3.1-Instruct-70B(-)             \\
\textbf{8}    & QwQ                    & QwQ(-)                                \\
\textbf{9}    & Qwen2.5-Instruct-7B    & Llama-3.1-Instruct-8B(\textcolor{red}{↑1})    \\
\textbf{10}   & Llama-3.1-Instruct-8B  & Qwen2.5-Instruct-7B(\textcolor{green}{↓1})    \\ \Xhline{1.5pt}
\end{tabular}
\caption{Performance rankings on WildBench and HuWildBench.}
\label{tab:WildBench_vs_HuWildBench}
\end{table}

\section{HuSimpleQA}
\label{App:husimpleqa}

\begin{figure}[!t]
    \centering
    \includegraphics[width=1.0\linewidth]{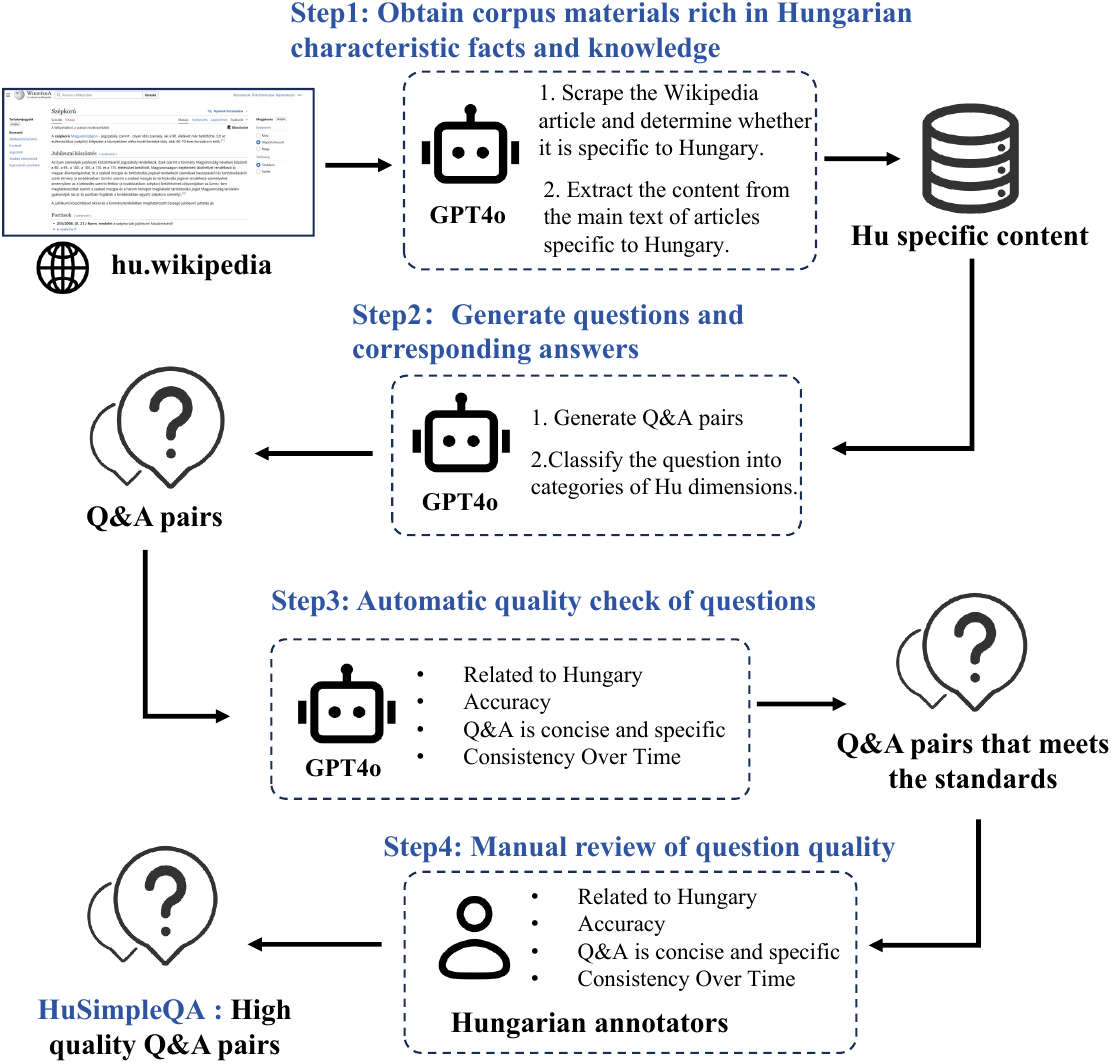}
    \caption{Constuction of HuSimpleQA.}
    \label{fig:Construction_HuSimpleQA}
\end{figure}

\subsection{Construction pipeline of HuSimpleQA}

The construction of HuSimpleQA consists of the following steps, as shown in Figure \ref{fig:Construction_HuSimpleQA}:

\textbf{(1) Obtaining corpora rich in Hungary-specific facts and knowledge:} First, we chose the Hungarian Wikipedia website \footnote{https://hu.wikipedia.org/} as the source of corpus material for question construction. We crawled all the entry pages and extracted their content. Next, we used GPT-4o to classify whether the entries were Hungary-specific, with the prompts detailed in Appendix \ref{App:husimpleqa_obtaindata}, Figure \ref{fig:Appendix_HuSimpleQA_prompt_entryfilter}. \footnote{We did not classify all the pages but instead randomly selected pages until we reached 8K Hungary-specific entries, at which point we stopped.}
We then used GPT to extract key information from the content of these entries suitable for question-answering. The extraction prompts are detailed in Appendix \ref{App:husimpleqa_obtaindata}, Figure \ref{fig:Appendix_HuSimpleQA_prompt_extractkeyinfo}. An example of the extracted key information is shown in Figure \ref{fig:Appendix_HuSimpleQA_example_keyinfo}.
As a result, we obtained Hungary-specific key information covering the eight distinct dimensions, totaling 4428 pieces of information.

\textbf{(2) Generating questions and corresponding answers:} We used the GPT-4o model to generate open-ended questions and corresponding answers based on the Hungary-specific key information obtained in the previous step. The prompt used is detailed in Appendix \ref{App:husimpleqa_oe}, Figure \ref{fig:Appendix_HuSimpleQA_prompt_oe}.
In this step, we generated a total of 9424 questions based on 4K entries. We then classified the generated questions according to the eight Hungary-specific dimensions outlined in Section \ref{sec:Hungarian-specific dimensions}, using GPT-4o with the prompt detailed in Appendix \ref{App:husimpleqa_oe}, Figure \ref{fig:Appendix_HuSimpleQA_prompt_questionclassify}.

\textbf{(3) Automatic quality checking of questions:} To ensure the quality of the questions, we used GPT-4o to check and filter the generated questions. We set the following four criteria, retaining only those questions that met all four standards (the corresponding prompt is detailed in Appendix \ref{App:husimpleqa_check}):

\textbf{- Criterion 1:} Hungary-specific: The content of the question-answer pair must align with the eight Hungary-specific dimensions proposed in this paper.

\textbf{- Criterion 2: }Accuracy: The information in the question-answer pair must align with the entry description and facts, and the answer should not be directly inferable from the question itself.

\textbf{- Criterion 3:} Concise and specific: The question and answer should be clear and concise, with no redundant information. The question should not contain nested sub-questions. The phrasing should be specific and direct, matching the scope of the answer (e.g., for time and location questions, the exact year/month/day/district/city must be specified).

\textbf{- Criterion 4:} Consistency Over Time: The answer should remain consistent over time and not be influenced by future events.

After the automatic checking process, we retained 5503 questions corresponding to 2666 entries.

\textbf{(4) Manual review of question quality:} To further ensure the quality of the questions, we hired Hungarian native speakers to manually review the questions. Annotators checked whether the questions met the four criteria mentioned in Step 3. During the annotation process, each question was assigned to two annotators, who received the questions but not the answers. A question was considered valid and retained only if both annotators agreed that it met all four criteria and that the provided answer matched the original reference answer. Detailed procedures are provided in Appendix \ref{App:husimpleqa_check}.
After these four steps, we obtained a total of 1293 questions, with their distribution across the eight Hungary-specific dimensions shown in Table \ref{tab:husimleqa_examples}.

\subsection{Obtaining corpora rich in Hungary-specific facts and knowledge}
\label{App:husimpleqa_obtaindata}
In the process of filtering Wikipedia entries with Hungarian characteristics, we randomly selected entries and provided both the entries and the first two paragraphs of the main content to GPT-4o (prompt shown in Figure~\ref{fig:Appendix_HuSimpleQA_prompt_entryfilter}) to determine if they were related to Hungary. If the entry was deemed relevant, it was categorized based on the eight characteristic dimensions proposed in this paper. At this stage, an “Others” category was added to ensure the focus on the eight thematic categories and to exclude interference from entries that belonged to other themes. The screening process stopped once the total number of Hungarian characteristic entries reached 8,000.

Due to the uneven distribution of entry themes on Wikipedia,  with more data in the categories of figures, geography and place, and history, we filtered the data based on the proportion of themes, ensuring that no single category exceeded 1,000 entries. This resulted in 4428 characteristic entries covering the eight dimensions.

Given the varying lengths of content describing entries on Wikipedia, we aimed to streamline the complexity of constructing subsequent question-answer pairs. To achieve this, we first employed GPT-4o to extract key information from the main text of each entry. This step aims to avoid any deviation from the theme caused by redundant content during the construction of the question-answer pairs (prompt shown in Figure \ref{fig:Appendix_HuSimpleQA_prompt_extractkeyinfo}). The results of the key information extraction are presented in Figure \ref{fig:Appendix_HuSimpleQA_example_keyinfo}.

\subsection{Generating questions and corresponding answers}
\label{App:husimpleqa_oe}
Based on the key information extracted and the provided entries, we utilized GPT-4o to generate 1-3 Hungarian open-ended question-answer pairs for each entry (prompt details in Figure \ref{fig:Appendix_HuSimpleQA_prompt_oe}). In total, 9,424 question-answer pairs were generated based on 4,000 entries.

Given that the focus and orientation of the generated question-answer pairs may differ from the original entry categories, this paper employed GPT-4o to reclassify the obtained question-answer pairs, with the corresponding prompt detailed in Figure \ref{fig:Appendix_HuSimpleQA_prompt_questionclassify}.

\subsection{Automatic quality checking of questions}
\label{App:husimpleqa_check}
We focused on evaluating the quality of the generated questions from two perspectives: the information contained in the question-answer pairs and the formulation of the questions. The quality assessment was divided into two stages, with each stage generating two evaluation metrics.
The first stage focuses on the relevance and correctness of the question information. We provided GPT-4o with the entry, its corresponding key information, and the generated question-answer pairs to verify whether the questions contain Hungarian-specific content and whether the information in the question-answer pairs aligns with the provided background material (prompt shown in Figure \ref{fig:Appendix_HuSimpleQA_prompt_quality1}).

Second, from the perspective of the precision of the question formulation, we only provided GPT-4o with the generated question-answer pairs to simulate real user response scenarios. This step emphasized evaluating whether the questions were based on objective facts, and whether the descriptions were precise and specific enough to allow independent answering without ambiguity. Additionally, we required that the answers remain unaffected by future events, ensuring consistency across any time period and guaranteeing the long-term validity of the dataset (prompt details in Figure \ref{fig:Appendix_HuSimpleQA_prompt_quality2}).

Based on the results of the above automated quality assessment, we retained only those question-answer pairs that passed all four evaluation criteria, resulting in a final set of 5,503 questions.

\subsection{Manual review of question quality}
\label{App:husimpleqa_anno}
To further ensure the quality of the constructed question-answer pairs, we engaged native Hungarian speakers to review these questions. Each question was independently reviewed by two annotators who could only see the questions and not the reference answers. The annotation process consisted of three main steps.

First, the annotators were required to determine whether the given questions aligned with the eight Hungarian-specific dimensions proposed in this paper. Next, they evaluated whether the questions met the four assessment criteria outlined in Step 3, ensuring that the questions were objectively framed, precisely described, had unique answers, contained correct information, and maintained consistent answers over time. Finally, if a question satisfied all the above criteria, the annotators provided the correct answer. During this process, annotators were permitted to search for relevant information online and provided reference sources for their answers.

To address potential issues such as overly obscure questions or non-fixed answers, we used GPT-4o to verify whether the annotated results matched the generated reference answers. If the annotated answer matched the reference answer, it was labeled as "CORRECT"; otherwise, it was labeled as "INCORRECT" (prompt details in Figure \ref{fig:Appendix_HuSimpleQA_prompt_check}). We selected question-answer pairs that both annotators deemed valid, Hungarian-specific, and consistent with the original reference answers as candidates for the HuSimpleQA dataset, resulting in a total of 2134 questions.

Considering that the HuSimpleQA dataset should exhibit diversity and broad coverage, we removed question-answer pairs belonging to the same entry, retaining only one question-answer pair per entry that best met the construction and evaluation criteria. The details of selecting optimal question-answer pairs prompt can be seen in Figure \ref{fig:Appendix_HuSimpleQA_prompt_select}.

Through this process, we obtained a total of 1293 pieces of Hungarian open-ended question-answer pairs, with the category distribution shown in Table \ref{tab:husimleqa_examples}.

\subsection{Inference prompt}
\label{App:husimpleqa_inference}
We constructed prompts in two languages for model inference, as shown in Figure \ref{fig:Appendix_HuSimpleQA_prompt_inference}, while also instructing the model to provide a confidence score (ranging from 1 to 100)to measure the model’s confidence in its generated answers.

\subsection{LLM-as-judge}
\label{App:husimpleqa_judge}

Following the approach of SimpleQA~\cite{wei2024simpleQA}, we employed GPT-4o as the judge to evaluate the LLM's responses. The evaluation criteria for this step were similar to those used in the manual review process of Step 4. In addition to the classification labels \texttt{CORRECT} and \texttt{INCORRECT}, we introduced an additional category, \texttt{NOT\_ATTEMPTED} to further assess the model's ability to respond to questions and the breadth of its knowledge coverage (prompt details in Figure \ref{fig:Appendix_HuSimpleQA_prompt_judge}).

Based on the results from the judge, we evaluate the performance of the LLM on HuSimpleQA using the following five metrics:

\textbf{- Correct (CO):} The predicted answer completely encompasses the reference answer without any conflicting or contradictory information.

\textbf{- Not Attempted (NA):} The predicted answer does not fully include the reference answer, but there are no contradictions between the two.

\textbf{- Incorrect (IN):} The predicted answer contradicts the reference answer, regardless of whether the contradiction is resolved.

\textbf{- Correct Given Attempted (CGA):} This metric measures the percentage of accurately answered questions out of all attempted questions.

\textbf{- F-score:} This metric calculates the harmonic mean between the proportion of correct answers and the proportion of correct answers among attempted questions.

The formulas for CGA and F-score are as follows:

\begin{equation}
\text{CGA} = \frac{c}{c + i}
\label{eq:cga}
\end{equation}
\begin{equation}
\text{F-Score} = \frac{2}{\frac{c+i}{c} + \frac{c+i+n}{c}} = \frac{2c}{2c + 2i + n}
\label{eq:f-score}
\end{equation}

Here, \(c\) represents the number of \texttt{CORRECT}ly answered questions, \( i \) represents the number of \texttt{INCORRECT}ly answered questions, and \( n \) represents the number of \texttt{NOT\_ATTEMPTED} questions.
\begin{table}[t]
\small
\centering
\begin{tabular}{lc}
\Xhline{1.5pt}
\textbf{Model} & \textbf{Judge Correct Ratio} \\
\Xhline{1.0pt}
Deepseek-R1              & 0.98 \\
Deepseek-V3              & 1.00 \\
Qwen2.5-Instruct-7B      & 0.98 \\
Qwen2.5-Instruct-72B     & 0.98 \\
Llama-3.1-Instruct-8B    & 1.00 \\
Llama-3.1-Instruct-70B   & 1.00 \\
GPT-4o                   & 0.99 \\
GPT-4o-mini              & 1.00 \\
o1-mini                  & 1.00 \\
QwQ                      & 0.98 \\
\Xhline{1.5pt}
\end{tabular}
\caption{Accuracy of LLM-as-Judge for HuSimpleQA.}
\label{tab:llm_judge_correct_ratio}
\end{table}

To verify the effectiveness of LLM-as-judge, we randomly selected 100 samples from each of the 10 evaluated models for inspection, with accuracy exceeding 98\% for all models, as shown in Table \ref{tab:llm_judge_correct_ratio}.

\begin{table*}[h]
\footnotesize
\centering
\begin{tabular}{llllp{2cm}l}
\Xhline{1.5pt}
\textbf{model} & \textbf{correct} & \textbf{incorrect} & \textbf{not\_attempted} & \textbf{correct given attempted} & \textbf{F-Score} \\ \hline
\textbf{GPT-4o}                 & 50.3   & 40.61   & 9.09    & 55.33   & 52.69   \\
\textbf{GPT-4o-mini}                       & 24.52  & 74.07   & 1.42    & 24.87   & 24.69   \\
\textbf{Deepseek-V3}         & 32.71  & 64.08   & 3.2     & 33.8    & 33.24   \\
\textbf{QwQ}                   & 9.17   & 52.68   & 38.15   & 14.82   & 11.33   \\
\textbf{Deepseek-R1}         & 34.58  & 62.15   & 3.28    & 35.75   & 35.15   \\
\textbf{Qwen2.5-7B-Instruct}     & 5.29   & 84.13   & 10.58   & 5.92    & 5.59    \\
\textbf{Qwen2.5-72B-Instruct}    & 15.05  & 78.61   & 6.33    & 16.07   & 15.54   \\
\textbf{Llama-3.1-8B-Instruct}  & 14.9   & 80.25   & 4.84    & 15.66   & 15.27   \\
\textbf{Llama-3.1-70B-Instruct} & 36.36  & 61.03   & 2.61    & 37.34   & 36.84   \\
\textbf{o1-mini}                & 16.24  & 44.19   & 39.57   & 26.88   & 20.25   \\ \Xhline{1.5pt}
\end{tabular}
\caption{Complete results of HuSimpleQA}
\label{tab:The complete result of HuSimpleQA}
\end{table*}

\begin{table}[t]
\scriptsize
\centering
\begin{tabular}{ccc}
 \Xhline{1.5pt}
\textbf{Rank} & \textbf{SimpleQA} & \textbf{HuSimpleQA} \\ \Xhline{1.0pt}
\textbf{1}    & GPT-4o            & GPT-4o(-)           \\
\textbf{2}    & Deepseek-R1      & Llama-3.1-Instruct-70B(\textcolor{red}{↑1})  \\
\textbf{3}    & Llama-3.1-Instruct-70B    & Deepseek-R1(\textcolor{green}{↓1})    \\
\textbf{4}    & Deepseek-V3      & Deepseek-V3(-)     \\
\textbf{5}    & QwQ   & GPT-4o-mini(\textcolor{red}{↑3})     \\
\textbf{6}    & Llama-3.1-Instruct-8B     & o1-mini(\textcolor{red}{↑3})         \\
\textbf{7}    & Qwen2.5-Instruct-72B       & Llama-3.1-Instruct-8B(\textcolor{green}{↓1})      \\
\textbf{8}    & GPT-4o-mini       & Qwen2.5-Instruct-72B(\textcolor{green}{↓1})   \\
\textbf{9}    & o1-mini           & QwQ(\textcolor{green}{↓4}) \\
\textbf{10}   & Qwen2.5-Instruct-7B        & Qwen2.5-Instruct-7B(-)       \\ \Xhline{1.5pt}
\end{tabular}
\caption{Performance rankings on SimpleQA~\cite{wei2024simpleQA} and HuSimpleQA}
\label{tab:SimpleQA_vs_HuSimpleQA}
\end{table}

\subsection{Comparing the Performance Ranking on HuSimpleQA and SimpleQA}

We compare the performance rankings of LLM on HuSimpleQA and SimpleQA~\cite{wei2024simpleQA}, as shown in Table ~\ref{tab:SimpleQA_vs_HuSimpleQA}.

\subsection{Results of Translating HuSimpleQA to English}
\begin{table}[t]
\scriptsize
\centering
\begin{tabular}{ccc}
 \Xhline{1.5pt}
\textbf{Model} & \textbf{Original HuSimpleQA} & \textbf{Translated to EN} \\ \Xhline{1.0pt}
GPT-4o           & 50.3   & 35.96 (\textcolor{green}{↓14.34}) \\
GPT-4o-mini      & 24.52  & 19.72 (\textcolor{green}{↓4.80})  \\
QwQ              & 9.17   & 12.84 (\textcolor{red}{↑3.67})    \\
Deepseek-R1      & 34.58  & 32.02 (\textcolor{green}{↓2.56})  \\
Deepseek-V3      & 32.71  & 29.08 (\textcolor{green}{↓3.63})  \\
Llama-3.1-70B    & 36.36  & 23.82 (\textcolor{green}{↓12.54}) \\
Llama-3.1-8B     & 14.9   & 14.08 (\textcolor{green}{↓0.82})  \\
o1-mini          & 16.24  & 14.31 (\textcolor{green}{↓1.93})  \\
Qwen2.5-72B      & 15.05  & 8.89  (\textcolor{green}{↓6.16})  \\
Qwen2.5-7B       & 5.29   & 15.47 (\textcolor{red}{↑10.18})   \\ \Xhline{1.5pt}
\end{tabular}
\caption{Results of translating HuSimpleQA to
English.}
\label{tab:HuSimpleQA_transEN}
\end{table}

As shown in Table~\ref{tab:main_result}, the current leading LLMs perform poorly on HuSimpleQA. To further analyze and confirm whether these issues stem from the models' lack of knowledge of the Hungarian language or Hungarian-specific cultural aspects, we translated HuSimpleQA into English and reevaluated the models. The results, presented in Table \ref{tab:HuSimpleQA_transEN}, show that after translation, 2 out of 10 models improved, while 8 experienced a decline in performance, indicating that translation generally worsens performance. This suggests that LLMs have more difficulty understanding Hungarian cultural nuances than the language itself. These findings are consistent with the conclusions in 

\subsection{Human Performance on SimpleQA and HuSimpleQA}

\begin{table*}[h]
    \scriptsize
    \centering
    \begin{tabular}{lcccc}
        \Xhline{1.0pt}
        \textbf{Model}          &  \textbf{HuSimpleQA-abs}  &  \textbf{HuSimpleQA-rel}  &  \textbf{SimpleQA-abs}    &  \textbf{SimpleQA-rel}    \\
        \midrule
        GPT-4o                  & 55.33                     & 0.85                      &  38.0                     &  1.12 \\
        GPT-4o-mini             & 24.87                     & 0.38                      &  8.7                      &  0.26 \\
        Deepseek-V3             & 33.8                      & 0.52                      &  24.9                     &  0.73 \\
        QwQ                     & 14.82                     & 0.23                      &  21.0                     &  0.62 \\
        Deepseek\_R1            & 35.75                     & 0.55                      &  30.1                     &  0.89 \\
        Qwen2.5-7B-Instruct     & 5.92                      & 0.09                      &  5.3                      &  0.16 \\
        Qwen2.5-72B-Instruct    & 16.07                     & 0.25                      &  12.2                     &  0.36 \\
        Llama-3.1-8B-Instruct   & 15.66                     & 0.24                      &  12.6                     &  0.37 \\
        Llama-3.1-70B-Instruct  & 37.34                     & 0.58                      &  29.3                     &  0.86 \\
        o1-mini                 & 26.88                     & 0.41                      &  11.3                     &  0.33 \\
        \midrule
        AVG                     & 26.65                     & 0.40                      &  19.34                    &  0.57 \\
        \Xhline{1.5pt}
    \end{tabular}
    \caption{The results of HuSimpleQA and SimpleQA, abs is the absolute performance of CGA metric, rel is the performance compared to human.}
    \label{tab:husimpleqa&simpleqaVShuman}
\end{table*}

We hire human annotators to complete the tasks in the SimpleQA and HuSimpleQA datasets in order to assess human performance on these datasets.
HuSimpleQA was annotated by Hungarian native speakers, while SimpleQA was handled by English native speakers. We used two annotators per dataset, averaging their performance metrics to determine human performance. Annotators were instructed NOT to use external tools like search engines or GPT. For HuSimpleQA and SimpleQA, they were told to answer "I don't know" if uncertain about a question.

The results for HuSimpleQA and SimpleQA are shown in Table \ref{tab:husimpleqa&simpleqaVShuman}.
The average absolute performance (CGA) of the ten models is 26.65 vs 19.34, highlighting significant room for improvement in LLM performance on both datasets.
Regarding average relative performance to humans, the scores are 0.40 vs 0.57, indicating that LLMs perform much worse on HuSimpleQA compared to SimpleQA. 
Despite better absolute performance on HuSimpleQA, LLMs are significantly weaker in simple factual ability in Hungarian than in English.

\begin{table*}[t]
\tiny
\centering
\begin{tabular}{ccccc}
\Xhline{1.5pt}
\textbf{Rank} & \textbf{HuProverbRea} & \textbf{MAPS (en)} & \textbf{MAPS (bn)} & \textbf{MAPS (id)} \\ \Xhline {1.0pt}
\textbf{1} & {GPT-4o} & {GPT-4o (-)} & {GPT-4o (-)} & {GPT-4o-mini (\textcolor{red}{↑3})} \\
\textbf{2} & {Llama-3.1-Instruct-70B} & {Llama-3.1-Instruct-70B (-)} & {Deepseek-V3 (\textcolor{red}{↑1})} & {Qwen2.5-Instruct-72B (\textcolor{red}{↑4})} \\
\textbf{3} & {Deepseek-V3} & {Qwen2.5-Instruct-72B (\textcolor{red}{↑3})} & {Deepseek-R1 (\textcolor{red}{↑2})} & {Llama-3.1-Instruct-70B (\textcolor{green}{↓1})} \\
\textbf{4} & {GPT-4o-mini} & {GPT-4o-mini (-)} & {Qwen2.5-Instruct-72B (\textcolor{red}{↑2})} & {GPT-4o (\textcolor{green}{↓3})} \\ 
\textbf{5} & {Deepseek-R1} & {Deepseek-V3 (\textcolor{green}{↓2})} & {o1-mini (\textcolor{red}{↑2})} & {Deepseek-V3 (\textcolor{green}{↓2})} \\
\textbf{6} & {Qwen2.5-Instruct-72B} & {Deepseek-R1 (\textcolor{green}{↓1})} & {GPT-4o-mini (\textcolor{green}{↓2})} & {o1-mini (\textcolor{red}{↑1})} \\
\textbf{7} & {o1-mini} & {Qwen2.5-Instruct-7B (\textcolor{red}{↑3})} & {Llama-3.1-Instruct-70B (\textcolor{green}{↓5})} & {Deepseek-R1 (\textcolor{green}{↓2})} \\
\textbf{8} & {QwQ} & {Llama-3.1-Instruct-8B (\textcolor{red}{↑1})} & {Llama-3.1-Instruct-8B (\textcolor{red}{↑1})} & {Llama-3.1-Instruct-8B (\textcolor{red}{↑1})} \\
\textbf{9} & {Llama-3.1-Instruct-8B} & {o1-mini (\textcolor{green}{↓2})} & {Qwen2.5-Instruct-7B (\textcolor{red}{↑1})} & {Qwen2.5-Instruct-7B (\textcolor{red}{↑1})} \\
\textbf{10} & {Qwen2.5-Instruct-7B} & {QwQ (\textcolor{green}{↓2})} & {QwQ (\textcolor{green}{↓2})} & {QwQ (\textcolor{green}{↓2})} 
\\
\Xhline{1.5pt}
\end{tabular}
\caption{Performance rankings on HuProverbRea and MAPS~\cite{MAPS} (part1/2)}
\label{tab:HuProverbRea_vs_MAPS_multilingul_1}
\end{table*}
\begin{table*}[t]
\tiny
\centering
\begin{tabular}{ccccc}
\Xhline{1.5pt}
\textbf{Rank} & \textbf{HuProverbRea} & \textbf{MAPS (de)} & \textbf{MAPS (ru)} & \textbf{MAPS (zh)} \\ \Xhline{1.0pt}
\textbf{1} & {GPT-4o} & {Llama-3.1-Instruct-70B (\textcolor{red}{↑1})} & {Deepseek-V3 (\textcolor{red}{↑2})} & {GPT-4o (-)} \\      
\textbf{2} & {Llama-3.1-Instruct-70B} & {GPT-4o-mini (\textcolor{red}{↑2})} & {o1-mini (\textcolor{red}{↑5})} & {Qwen2.5-Instruct-72B (\textcolor{red}{↑4})} \\
\textbf{3} & {Deepseek-V3} & {Qwen2.5-Instruct-72B (\textcolor{red}{↑3})} & {GPT-4o (\textcolor{green}{↓2})} & {Deepseek-V3 (-)} \\ 
\textbf{4} & {GPT-4o-mini} & {GPT-4o (\textcolor{green}{↓3})} & {Qwen2.5-Instruct-72B (\textcolor{red}{↑2})} & {o1-mini (\textcolor{red}{↑3})} \\
\textbf{5} & {Deepseek-R1} & {Llama-3.1-Instruct-8B (\textcolor{red}{↑4})} & {GPT-4o-mini (\textcolor{green}{↓1})} & {Llama-3.1-Instruct-70B (\textcolor{green}{↓3})} \\
\textbf{6} & {Qwen2.5-Instruct-72B} & {o1-mini (\textcolor{red}{↑1})} & {Deepseek-R1 (\textcolor{green}{↓1})} & {GPT-4o-mini (\textcolor{green}{↓2})} \\
\textbf{7} & {o1-mini} & {Deepseek-V3 (\textcolor{green}{↓4})} & {Qwen2.5-Instruct-7B (\textcolor{red}{↑3})} & {Qwen2.5-Instruct-7B 
(\textcolor{red}{↑3})} \\
\textbf{8} & {QwQ} & {Qwen2.5-Instruct-7B (\textcolor{red}{↑2})} & {Llama-3.1-Instruct-70B (\textcolor{green}{↓6})} & {Llama-3.1-Instruct-8B (\textcolor{red}{↑1})} \\
\textbf{9} & {Llama-3.1-Instruct-8B} & {Deepseek-R1 (\textcolor{green}{↓4})} & {Llama-3.1-Instruct-8B (-)} & {Deepseek-R1 (\textcolor{green}{↓4})} \\
\textbf{10} & {Qwen2.5-Instruct-7B} & {QwQ (\textcolor{green}{↓2})} & {QwQ (\textcolor{green}{↓2})} & {QwQ (\textcolor{green}{↓2})} 
\\
\Xhline{1.5pt}
\end{tabular}
\caption{Performance rankings on HuProverbRea and MAPS~\cite{MAPS} (part2/2)}
\label{tab:HuProverbRea_vs_MAPS_multilingul_2}
\end{table*}

\section{HuProverbRea}
\label{Appendix:HuProverbRea}

\subsection{Construction pipeline of HuProverb}

\begin{figure}[!t]
    \centering
    \includegraphics[width=1\linewidth]{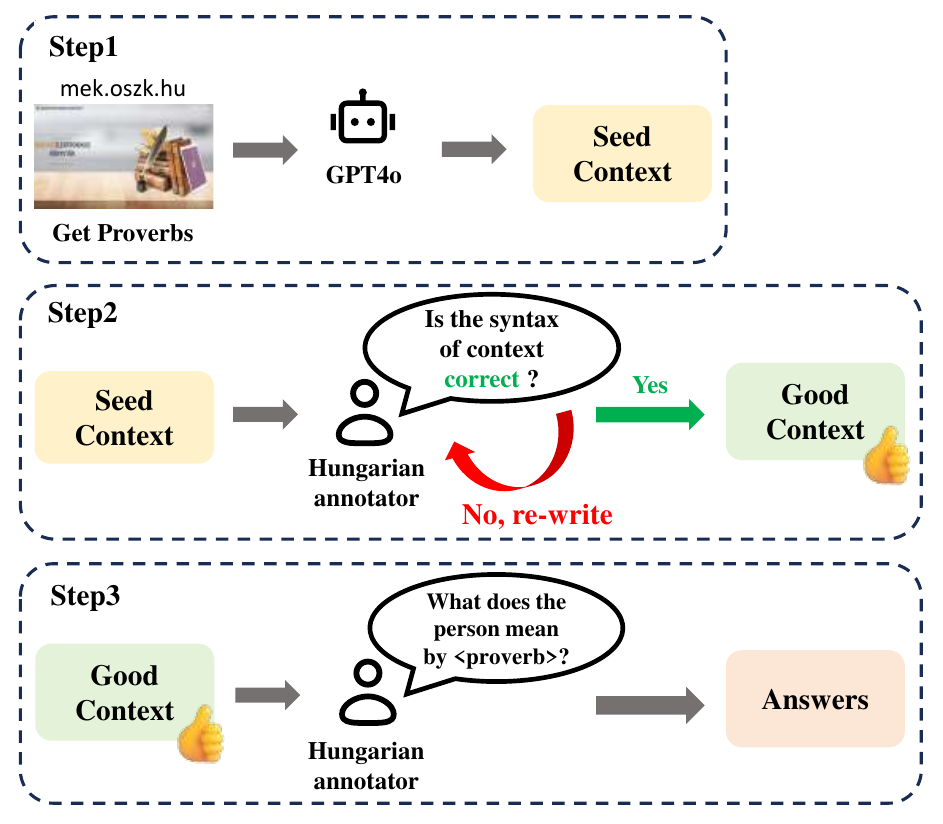}
    \caption{Construction of HuProverbRea.}
    \label{fig:Construction_HuProverbRea}
\end{figure}

The proverbs in HuProverbRea are from 2 separate sources. The first part, 733 traditional Hungarian proverbs, are collected from the website\footnote{\url{https://mek.oszk.hu/}}, where each proverb is assigned an English or Hungarian explanation. The other 402 proverbs, focusing on abbreviations and Internet slang, are manually collected and explained by native speakers of Hungarian.
Inspired by MAPS \cite{MAPS}, we adopt a human-in-loop pipeline to generate and refine the context for each Hungarian specific usage, as shown in Figure \ref{fig:Construction_HuProverbRea}. For each proverb, we first let GPT4 generate a seed context where the proverb is used. Then, we assign it to a Hungarian native speaker to check whether this context is grammatically correct and the use of slang is appropriate. If not, the annotator is required to manually write down a new context for the saying, which will be sent back to another annotator for inspection again. We continue the above procedures until all contexts pass the quality check. It's worth noting that each option of the 2CQ setting is manually constructed by a human annotator, and only when it passes the double check of two other annotators could it be considered usable. We choose not to involve LLM in this part because designing correct/incorrect options requires deep understanding of sayings, LLM may generate ambiguous options if it does not understand the proverb used in the context, and such pre-provided ambiguous options may negatively influence the creativity of the annotators. Finally, we obtain 1,135 Hungarian proverbs, each equipped with a context, an English explanation, and two candidate options for question \textit{``What does the speaker mean by the saying?''}.

\subsection{More examples of HuProverbRea}

The example of HuProverbRea is shown in Figure \ref{fig:Appendix_HuProverbRea_example_2CQ} and Figure \ref{fig:Appendix_HuProverbRea_example_OE}. The prompt for judging HuProverbRea is shown in Figure \ref{fig:Appendix_HuProverbRea_prompt_judge}.The prompt for  model inference on HuProverbRea is shown in Figure \ref{fig:Appendix_HuProverbRea_inf_prompt_2cq} and \ref{fig:Appendix_HuProverbRea_inf_prompt_oe}.

\begin{table*}[h]
    \scriptsize
    \centering
    \begin{tabular}{lcccc}
        \Xhline{1.0pt}
        \textbf{Model} &  \textbf{HuProverbRea-abs} &  \textbf{HuProverbRea-rel} &  \textbf{MAPs(en)-abs} &  \textbf{MAPs(en)-rel}\\
        \midrule
        Deepseek-V3 & 92.51 & 1.15 & 96.95 & 1.02 \\
        Deepseek\_R1 & 91.72 & 1.14 & 98.48 & 1.03 \\
        GPT-4o & 95.51 & 1.19 & 97.97 & 1.03 \\
        GPT-40-mini & 92.16 & 1.15 & 97.97 & 1.03 \\
        Llama-3.1-70B-Instruct & 93.83 & 1.17 & 98.48 & 1.03 \\
        Llama-3.1-8B-Instruct & 73.48 & 0.91 & 92.39 & 0.97 \\
        o1-mini & 87.67 & 1.09 & 92.89 & 0.97 \\
        Qwen2.5-72B-Instruct & 90.22 & 1.12 & 98.22 & 1.03 \\
        Qwen2.5-7B-Instruct & 67.05 & 0.83 & 94.16 & 0.99 \\
        QwQ & 84.23 & 1.05 & 67.51 & 0.71 \\
        \midrule
        AVG & 86.84 & 1.08 & 93.50 & 0.98 \\
        \Xhline{1.5pt}
    \end{tabular}
    \caption{The results of HuProverbRea and MAPs(en), abs is absolute performance, rel is the performance compared to human.}
    \label{tab:Huproverb&mapVShuman}
\end{table*}

\subsection{Comparing the Performance Ranking on HuProverbRea and MAPS}

We compare the performance rankings of LLM on HuProverbRea and MAPS~\cite{MAPS}, as shown in Table ~\ref{tab:HuProverbRea_vs_MAPS_multilingul_1} and ~\ref{tab:HuProverbRea_vs_MAPS_multilingul_2}.

\subsection{Human Performance on HuProverbRea and MAP(en)}

We hire human annotators to complete the tasks in the  HuProverbRea and MAP(en) datasets in order to assess human performance on these datasets.
HuProverbRea was annotated by Hungarian native speakers, while MAP(en) was handled by English native speakers. We used two annotators per dataset, averaging their performance metrics to determine human performance. Annotators were instructed NOT to use external tools like search engines or GPT.
The accuracy for HuProverbrea is 80.44\%. The accuracy for MAP(en) is 95.31\%.

Furthermore, we calculated the LLMs' relative performance to human, as shown in Table ~\ref{tab:Huproverb&mapVShuman}.

The average absolute performance of the ten models is 0.8684 vs 0.9350, indicating that LLMs perform worse on HuProverbRea compared to English proverb reasoning, suggesting room for improvement in Hungarian proverb tasks.

For average relative performance to humans, the scores are 1.08 vs 0.98. This indicates that while LLMs slightly underperform compared to humans on the MAP(en) dataset, they outperform humans on the HuProverbRea dataset, highlighting their strengths in Hungarian language-related knowledge and reasoning.

\section{HuMatchingFIB and HuStandardFIB}
\label{Appendix:HuMatchingFIB and HuStandardFIB}

\subsection{Construction of HuMatchingFIB and HuStandardFIB}

The questions for both HuMatchingFIB and HuStandardFIB are sourced from the Hungarian National Public Education Portal(NKP)\footnote{\url{https://www.nkp.hu/}}, a comprehensive platform for cultural funding and support in Hungary. This portal connects artists, cultural organizations, and the public with resources and opportunities to promote Hungarian culture both domestically and internationally. Notably, this website is a government initiative, reflecting the collaborative efforts between the Hungarian government and the European Union, particularly through projects or programs supported by the European Social Fund. 
After extracting the original questions from the NKP website, we engaged native Hungarian speakers to annotate the data. The annotation process involved manually extracting questions and their corresponding answers\footnote{The questions for HuMatchingFIB and HuStandardFIB on the NKP website are not in plain text but are instead presented in interactive modules, and the answers can only be obtained through additional interactive operations. As a result, the commonly used data cleaning and extraction methods for LLM pre-training datasets are unable to accurately extract these questions and their corresponding answers. Consequently, it can be concluded that the likelihood of these questions being incorporated into the LLM pre-training data in their proper format is minimal, thereby significantly reducing the potential risk of data contamination. This ensures the reasonableness and effectiveness of the test sets for HuMatchingFIB and HuStandardFIB.} , classifying the questions into appropriate categories, and filtering out questions that required additional modalities such as images, tables, audio, or video. This ensured that only purely language-based questions were retained. Through this process, we obtained 278 questions for the HuMatchingFIB task and 93 questions for the HuStandardFIB task, as shown in Table x.

\subsection{More examples of HuMatchingFIB and HuStandardFIB}

Examples of questions from HuMatchingFIB and HuStandardFIB are provided in Figure \ref{fig:Appendix_HuMatchingFIB_example} and Figure \ref{fig:Appendix_HuStandardFIB_example}. The prompt for  model inference on HuMatchingFIB and HuStandardFIB is shown in Figure \ref{fig:Appendix_HuMatchingFIB_inf_prompt} and Figure \ref{fig:Appendix_HuStandardFIB_inf_prompt}.

\subsection{Metric and Judge of HuMatchingFIB and HuStandardFIB }

HuMatchingFIB employs a rule-based evaluation approach, where the assessment is conducted at two levels: the blank level and the question level (as a single question may contain multiple blanks). The evaluation process is analogous to that of multiple-choice questions, and accuracy (acc) is used as the metric to determine performance.
The corresponding formula for blank level accuracy is as follows, where c represents the number of correctly predicted blanks in one question, t represents the number of blanks in one question.

\begin{equation}
\text{Acc}_{\text{blank level}} = \frac{\sum \text{blank}_{\text{c}}}{\sum \text{blank}_{\text{t}}}
\end{equation}

HuStandardFIB questions are designed with open-ended reference answers to accommodate variations in part of speech and semantics. We employ a many-to-one fuzzy matching mechanism. Fuzzy matching is a technique that calculates the similarity between strings, allowing for flexibility in matching by considering variations such as typos, synonyms, or different word orders. In this context, the model's answer is compared against a set of possible reference answers (where multiple correct answers may exist for a single question or blank). If the similarity score between the model's answer and any of the reference answers exceeds a predefined threshold, the answer is considered correct. This approach is particularly suitable for evaluating open-ended questions where exact matches are often infeasible due to the variability in acceptable responses. The annotator information involved in all tasks of OpenHuEval can be found in Appendix \ref{Appendix:Information_of_the_Annotators}.

\section{Analyzing LRM's thinking process on OpenHuEval}
\label{Appendix:LRM_reasoning_process}

\subsection{Analyzing LRM's thinking process on HuSimpleQA}

We use GPT-4o to break down the answers into thoughts. This is done in two steps: the first step is to identify expressions that may be a shift in thought (see the prompt in Figure \ref{fig:husimpleqa_shiftword_step1}), and the second step is to confirm whether it is indeed a shift in thought (set the prompt in Figure \ref{fig:husimpleqa_shiftword_step2}). 
Then, We utilized the LLM to evaluate whether each idea would lead to the correct answer, the prompt is shown in Figure \ref{fig:husimpleqa_thought_evaluation}. We consider a confident score of 2 as the correct thought.

Examples of the thoughts and corresponding correctness can be found in Figure \ref{fig:example of husimpleqa thought segmentation in ds} and Figure \ref{fig:example of husimpleqa thought segmentation in qwq}.

\subsection{Analyzing LRM's thinking process on HuMatchingFIB}

\subsubsection{Splitting LRM's thinking process into segments and classifying these segments}
\label{Appendix:LRM_reasoning_process_HuMatchingFIB_segment}

We first divided the thinking process of LRMs into multiple segments, with each segment categorized as \texttt{introduction}, \texttt{reasoning}, \texttt{review}, or \texttt{summary}. The definitions of these four segment types are provided in Table \ref{tab:thinking_process_segments_cates_def}. Typically, each thinking process begins with an \texttt{introduction} segment, includes several \texttt{reasoning} segments and some \texttt{review} segments in the middle, and concludes with a \texttt{summary} segment. The segmentation and classification were performed using GPT-4o, with the prompt template detailed in Figures \ref{fig:Appendix_HuMatchingFIB_prompt_segment_p1},  \ref{fig:Appendix_HuMatchingFIB_prompt_segment_p2} and \ref{fig:Appendix_HuMatchingFIB_prompt_segment_p3}.
Examples of the segmentation and classification can be found in Figure~\ref{fig:humatchingfib_segment_example_deepseek} and Figure~\ref{fig:humatchingfib_segment_example_qwq}.

\begin{table*}[t]
\small
\centering
\begin{tabular}{lp{8cm}}
\Xhline{1.5pt}
\textbf{Segment types} & \textbf{Definition} \\ \midrule
\texttt{introduction} &
typically located at the beginning of the reasoning process; usually consists of the LRM's brief restatement of the question and the descriptive account of the work it is about to undertake; does not include the actual start of the analysis of the question.
   \\\midrule
\texttt{reasoning} & typically constitutes the main body of the reasoning process; includes the detailed thinking and reasoning steps undertaken by the LRM to solve the fill-in-the-blank questions.  \\\midrule
\texttt{review} & usually occurs after the reasoning process is essentially
complete but before the final output. This section typically includes a review of the entire reasoning process and may contain keywords or phrases such as ``Overall, ...'' or ``double check...'' .   \\\midrule
\texttt{summary}   & Summarizes the overall content or provides final conclusions, often using phrases like 'in conclusion' or 'overall'.  \\ \Xhline{1.5pt}
\end{tabular}
\caption{Definitions of the four categories of segments in the thinking process of LRM on HuMatchingFIB.}
\label{tab:thinking_process_segments_cates_def}
\end{table*}

\subsubsection{Tagging the \texttt{reasoning} segments along the dimensions}

\label{Appendix:LRM_reasoning_process_HuMatchingFIB_tag_dim}

We further tagged the \texttt{reasoning} segments according to the following four dimensions: \textbf{(Dim1) correctness}: Are the answers in this \texttt{reasoning} segment correct? \textbf{(Dim2) complexity}: In this \texttt{reasoning} segment, does the LRM simply assert the answer, or does it involve more complex reasoning? \textbf{(Dim3) scope}: Does this \texttt{reasoning} segment focuses on a single blank, modifies previous blanks, or addresses multiple blanks? \textbf{(Dim4) language transfer}: Does the LRM switch languages within this \texttt{reasoning} segment? 
The details of the tagging can be found in Table~\ref{tab:HuMatchingFIB_tagging_dimensions}.
Examples of the tagging results can be found in Figure~\ref{fig:humatchingfib_segment_example_deepseek} and Figure~\ref{fig:humatchingfib_segment_example_qwq}.

\begin{table*}[]
\small
\centering
\setlength{\tabcolsep}{10pt}
\renewcommand{\arraystretch}{1.5}

\begin{tabular}{lp{4cm}l}
 \Xhline{1.5pt}
\textbf{Dimension}                  & \textbf{Description}                                                                                                 & \textbf{Tags} \\ \hline
Dim 1: Correctness &
  Are the answers in this \texttt{reasoning} segment correct? Class 4 is used when no conclusion is reached. &
  \begin{tabular}[t]{@{}l@{}}
  Class 1: Completely Incorrect\\
  Class 2: Partially Correct\\
  Class 3: Completely Correct\\
  Class 4: Non Conclusion
  \end{tabular} \\ \midrule
Dim 2: Complexity & 
  In this \texttt{reasoning} segment, does the LRM simply assert the answer, or does it involve more complex reasoning? &
  \begin{tabular}[t]{@{}l@{}}
  Class 1: Simple Assertion\\
  Class 2: Complex Thought
  \end{tabular} \\ \midrule
Dim 3: Scope &
  Does this \texttt{reasoning} segment focuses on a single blank, modifies previous blanks, or addresses multiple blanks? &
  \begin{tabular}[t]{@{}l@{}}
  Class 1: Only Current Blank\\
  Class 2: Modify Previous Blanks\\
  Class 3: Current Blank and Consecutive Blank
  \end{tabular} \\ \midrule
Dim 4: Language Transfer    & 
  Does the LRM switch languages within this \texttt{reasoning} segment (e.g., Hungarian to English)?  &
  \begin{tabular}[t]{@{}l@{}}
  Class 1: Contains Language Transfer\\
  Class 2: No Language Transfer
  \end{tabular} \\  \Xhline{1.5pt}
\end{tabular}
\caption{Tagging dimensions of the \texttt{reasoning} segments in LRM's thinking process on HuMatchingFIB}
\label{tab:HuMatchingFIB_tagging_dimensions}
\end{table*}

\section{Information of the Annotators}
\label{Appendix:Information_of_the_Annotators}
\begin{table}[h]
\footnotesize
\centering
\begin{tabular}{ccc}
\Xhline{1.5pt}
Task                                                                       & \# Anotater & Total working hours \\ \hline
HuSimpleQA                                                                 & 14          & 161.9               \\
HuWildBench                                                                & 5           & 55.2                \\
HuProverbRea                                                               & 15          & 118.2               \\
\begin{tabular}[c]{@{}c@{}}HuMatchingFIB and\\  HuStandardFIB\end{tabular} & 8           & 84.5                \\ \Xhline{1.5pt}
\end{tabular}
\caption{Information of the Annotators}
\label{tab:Information of the Annotators}

\end{table}
We submitted the annotation task online to a professional data annotation company, which organized annotators to complete the annotation work. In the construction phase of OpenHuEval, the annotations were carried out by professional annotators who are native Hungarian speakers. Table \ref{tab:Information of the Annotators} shows the number of annotators and the total time spent on each task.
All annotators involved in this project hold a bachelor's or master's degree, with academic backgrounds in fields such as Social Sciences, Translating and Interpreting, English Studies, and IT Engineering. They all possess the ability to distinguish subtle aspects of the Hungarian language and handle Hungarian-specific knowledge effectively.


\begin{figure*}[!t]
    \centering
    \includegraphics[width=0.8\linewidth]{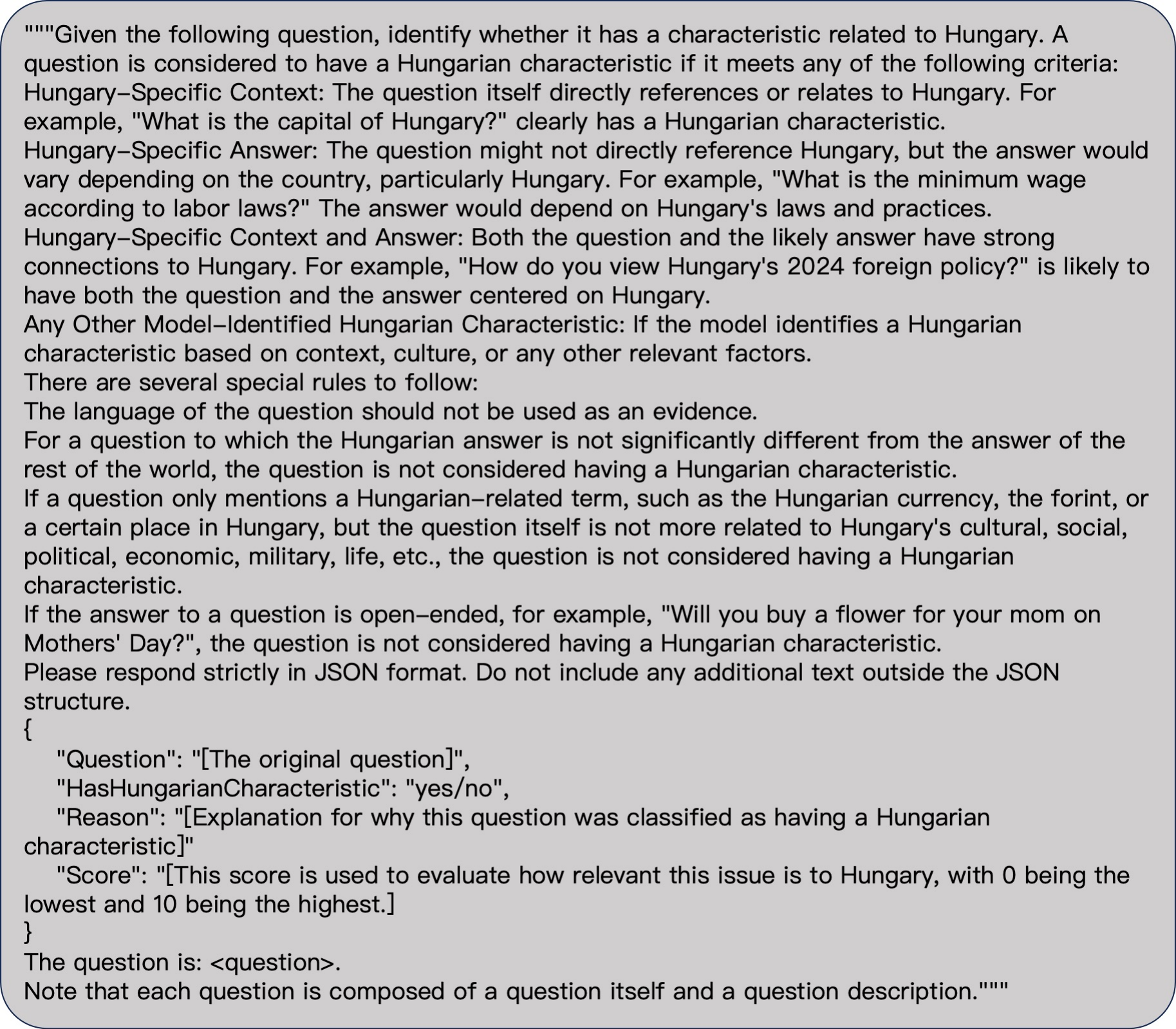}
    \caption{Prompt template for automatic filtering of user questions related to Hungarian specifics in the construction of HuWildBench.}
    \label{fig:Appendix_HuWildBenchPromptSelectHuSpecialQuestion}
\end{figure*}

\begin{figure*}[!t]
    \centering
    \includegraphics[width=0.8\linewidth]{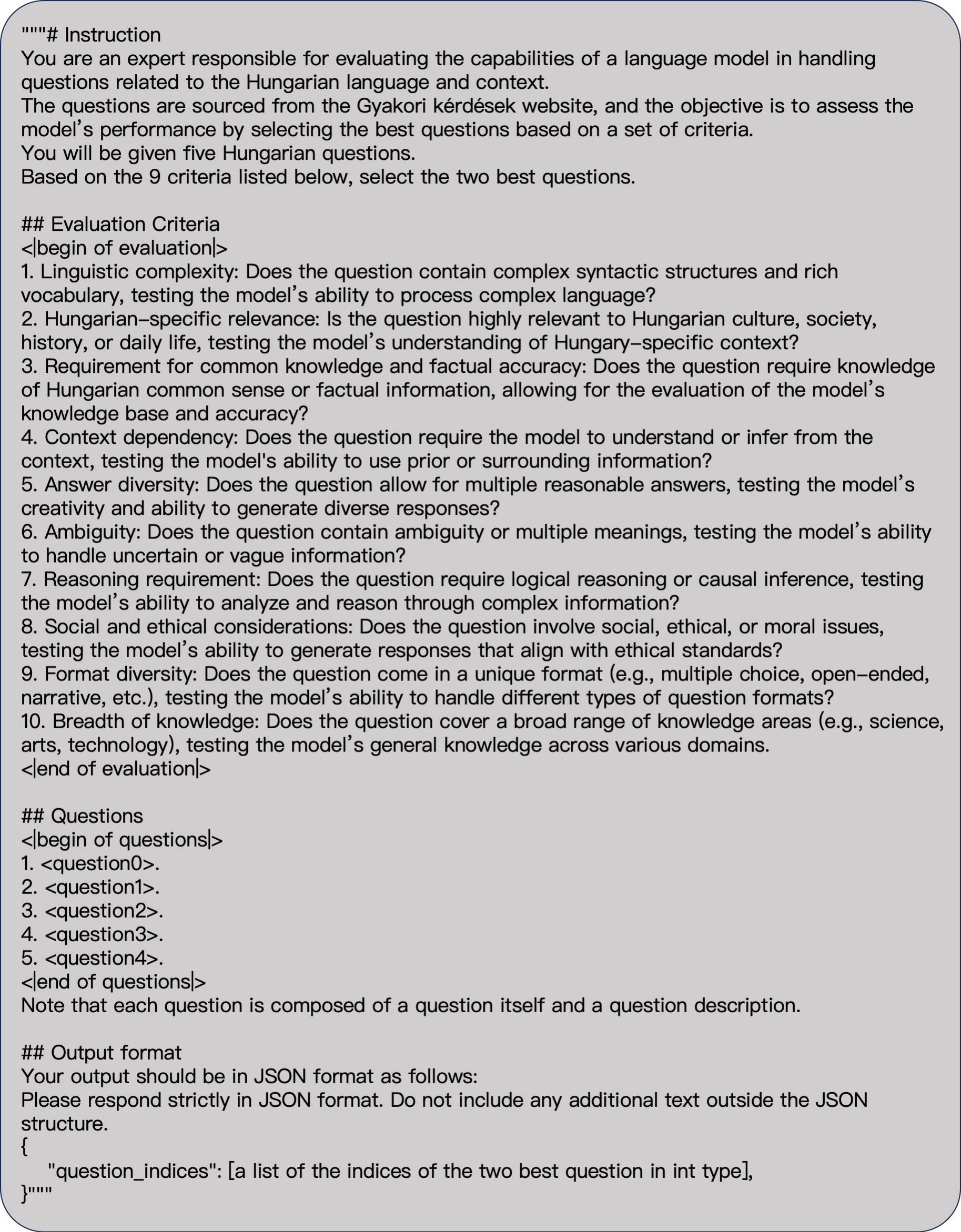}
    \caption{Prompt template for automatic filtering of high-quality question in the construction of HuWildBench.}
    \label{fig:Appendix_HuWildBenchPromptSelectHQQuestion}
\end{figure*}

\begin{table*}[]
\scriptsize
\centering
\begin{tabular}{ll}

\Xhline{1.5pt}

\textbf{Hungarian-specific dimensions:} $\mathcal{LCC}$ &  \textbf{Count:} 365 \\
\Xhline{1pt}

\begin{tabular}[c]{@{}p{6cm}@{}@{}}
\textbf{Question Example:} \\ 
a kérdés az: Mi lesz a jövőben a szocializmus alatt megépül sok panellel? \\  
a leírás: Úgy tudom, hogy kb 60 éves életciklusra tervezték őket. Magyarországon (és a környező országokban is) rengeteg ember él bennük. Mi fog történni akkor, ha lakhatatlanná kezdenek válni? Mi lesz azzal a sok emberrel? Meg a panelokkal? \\
 \\
\textcolor{blue}{\textbf{Translation to EN:}} \\ 
\textcolor{blue}{The question is: What will happen in the future to the many panels built under socialism?} \\  
\textcolor{blue}{The description is: I understand they are designed for a life cycle of about 60 years. There are a lot of people living in them in Hungary (and surrounding countries). What will happen if they start to become uninhabitable? What will happen to all those people? And the panels?}
\end{tabular} &

\begin{tabular}[c]{@{}p{9cm}@{}}
\textbf{Checklist:} \\  
"Does the response provide an analysis of the current condition and expected lifespan of the panel buildings in Hungary and neighboring countries?", \\ 
"Does the response address the expected lifespan of panel buildings and their current age?", \\ 
"Are there any historical or international examples included to illustrate possible outcomes or strategies?", \\  
"Does the response consider the economic implications of renovating or replacing panel buildings?", \\ 
"Does the response include potential government or private sector plans or policies addressing the future of these buildings and their residents?", \\ 
"Does the answer discuss potential scenarios for when these buildings become uninhabitable?", \\ "
Are environmental and urban planning aspects of dealing with aging panel buildings mentioned?", \\ 
"Is there an explanation of possible solutions or government plans for relocating residents?"\end{tabular} \\

\Xhline{1.5pt}

\textbf{Hungarian-specific dimensions:} $\mathcal{EP}$  &  \textbf{Count:} 201   \\
\Xhline{0.5pt}

\begin{tabular}[c]{@{}p{6cm}@{}@{}}
\textbf{Question Example:} \\ 
a kérdés az: A kárpátaljai magyarok Ukrajnában oroszul vagy ukránul tanultak meg a 2000-es évek közepén? \\  
a leírás: Mit tanítottak az iskolákban? Mennyire reális az, hogy valakire szinte semmi se ragad a környezetéből? Vannak olyan tömb területek ahol mondjuk egy magyar gyereknek egyáltalán nem kell helyi ukránokkal beszélnie? Egyáltalán a helyi ukránok ukránul beszéltek a 2000-es években? \\
 \\
\textcolor{blue}{\textbf{Translation to EN:}} \\ 
\textcolor{blue}{The question is: Did Hungarians in Transcarpathia learn Russian or Ukrainian in Ukraine in the mid-2000s?} \\  
\textcolor{blue}{The description is: What was taught in schools? How realistic is it that almost nothing sticks to someone from their environment? Are there block areas where, say, a Hungarian child doesn't have to speak to local Ukrainians at all? Did local Ukrainians even speak Ukrainian in the 2000s?}\end{tabular} &

\begin{tabular}[c]{@{}p{9cm}@{}}
\textbf{Checklist:}\\ 
"Does the answer provide information on the language predominantly spoken by local Ukrainians in Transcarpathia in the 2000s?", \\ 
"Does the response discuss the social and linguistic dynamics in areas with significant Hungarian populations, including interactions with local Ukrainians?", \\ 
"Does the response clearly explain the educational policies and language of instruction in schools for Hungarians in Transcarpathia during the mid-2000s?", \\ 
"Does the response accurately describe the language of instruction in Transcarpathian Hungarian schools in the mid-2000s?", \\ 
"Does the response consider the historical and political context of language policies in Ukraine during this period?", \\ 
"Does the response provide insight into whether local Ukrainians predominantly spoke Ukrainian during the 2000s?", \\ 
"Does the response offer a balanced view of cultural and linguistic integration in Transcarpathia during the specified period?", \\ 
"Does the answer address the likelihood of a Hungarian child not acquiring any local language skills from their environment?", \\ 
"Does the response discuss the existence of predominantly Hungarian areas where interaction with local Ukrainians might be limited?"\end{tabular} \\

\Xhline{1.5pt}

\textbf{Hungarian-specific dimensions:} $\mathcal{PPL}$  &  \textbf{Count:} 299   \\
\Xhline{0.5pt}

\begin{tabular}[c]{@{}p{6cm}@{}@{}}\textbf{Question Example:} \\ a kérdés az: Mi történt azzal, aki az 50-es években a felhívás ellenére sem jegyzett "önként" békekölcsönt? Érhette ezért retorzió az embert? \\  a leírás: Persze nyilván volt, amilyen "bolondos" idők jártak nálunk akkortájt. Biztos kikiáltották reakciósnak vagy fasisztának, meg a "népi demokrácia" ellenségének. \\
 \\
\textcolor{blue}{\textbf{Translation to EN:}} \\ 
\textcolor{blue}{The question is: What happened to the man who did not "voluntarily" subscribe to a peace charter in the 1950s, despite the call? Could he have been retaliated against for this?} \\  
\textcolor{blue}{The description is: He must have been branded a reactionary or a fascist or an enemy of 'people's democracy'.}\end{tabular} &

\begin{tabular}[c]{@{}p{9cm}@{}}\textbf{Checklist:}\\ 
"Does the answer address the political labels mentioned in the description (e.g., 'reactionary', 'fascist', 'enemy of people's democracy')?", \\ 
"Does the response differentiate between official consequences and social/societal repercussions for not subscribing to the peace loan?", \\ 
"Does the response address potential consequences for individuals who did not subscribe to the peace loan, with references to historical examples or documentation?", \\ 
"Does the response provide a balanced view, considering both potential punitive measures and any instances of leniency or exceptions, if applicable?", \\ 
"Is there a clear explanation of what 'békekölcsön' (peace loan) was and its significance during that time period?", \\ 
"Does the response accurately describe the historical context of the 1950s in Hungary?", \\ 
"Is there an analysis of the societal and governmental attitudes toward dissenters in Hungary during the 1950s, including any possible labels or accusations they might have faced?", \\ 
"Does the response provide specific examples of potential retaliations against those who didn't subscribe to the peace loan?"\end{tabular} \\

\Xhline{1.5pt}

\textbf{Hungarian-specific dimensions:} $\mathcal{BF}$  &  \textbf{Count:} 289   \\
\Xhline{0.5pt}

\begin{tabular}[c]{@{}p{6cm}@{}@{}}\textbf{Question Example:} \\ a kérdés az: Meddig tartható fent Magyarország negatív külkereskedelmi mérlege? \\  
a leírás: Nem a háború óta, hanem már 2021 nyarától folyamatosan negatív az ország külkereskedelmi mérlege. Júliusban és augusztusban összesen több, mint 1000 milliárd forintnyi mínusz keletkezett. Persze a többi hónap nem volt ennyire szörnyű, de ez csak erre az évre már több, mint 2000 milliárd forintnyi mínusz. Változatlan devizaimport mellett a mérséklődött energiaárakkal is több, mint 1000 milliárdos negatív mérleg hozható össze 2023-ban. Meddig lehet ezt tovább folytatni? Meddig elég a devizatartalék a hiány pótlására? \\
 \\
\textcolor{blue}{\textbf{Translation to EN:}} \\ 
\textcolor{blue}{The question is: How long can Hungary maintain a negative trade balance?} \\  
\textcolor{blue}{The description is: In July and August there was a total deficit of more than HUF 1000 billion. Of course, the other months were not so bad, but for this year alone it is already more than HUF 2000 billion in deficit. Even with unchanged foreign exchange imports and moderating energy prices, a negative balance of more than 1,000 billion in 2023 could be created. How long can this go on? How long will foreign exchange reserves be enough to cover the deficit?}\end{tabular} &

\begin{tabular}[c]{@{}p{9cm}@{}}
\textbf{Checklist:}\\ 
"Does the response analyze Hungary's current foreign exchange reserves and their sufficiency in covering the trade deficit?", \\ 
"Is there an exploration of historical trends and comparisons to similar situations in other countries to provide context?", \\ 
"Is the impact of energy prices on the trade balance accurately assessed in the response?", \\ 
"Does the response offer a clear and supported prediction or timeframe for how long Hungary can sustain its negative trade balance?", \\ 
"Is there an analysis of the factors affecting Hungary's foreign exchange reserves and their ability to cover the deficit?", \\ 
"Does the answer provide a clear timeline or projection for how long the negative balance can be sustained?", \\ 
"Are there comparisons made to similar situations in other countries or historical precedents in Hungary?", \\ 
"Does the response accurately explain the current state of Hungary's foreign trade balance?"\end{tabular} \\

\Xhline{1.5pt}
  
\end{tabular}
\caption{Examples of HuWildBench. The blue font is the English translation of the original OpenHuEval examples, used for visualization.}
\label{tab:HuWildBench_QuestionExamples}
\end{table*}

\begin{figure*}[!t]
    \centering
    \includegraphics[width=1\linewidth]{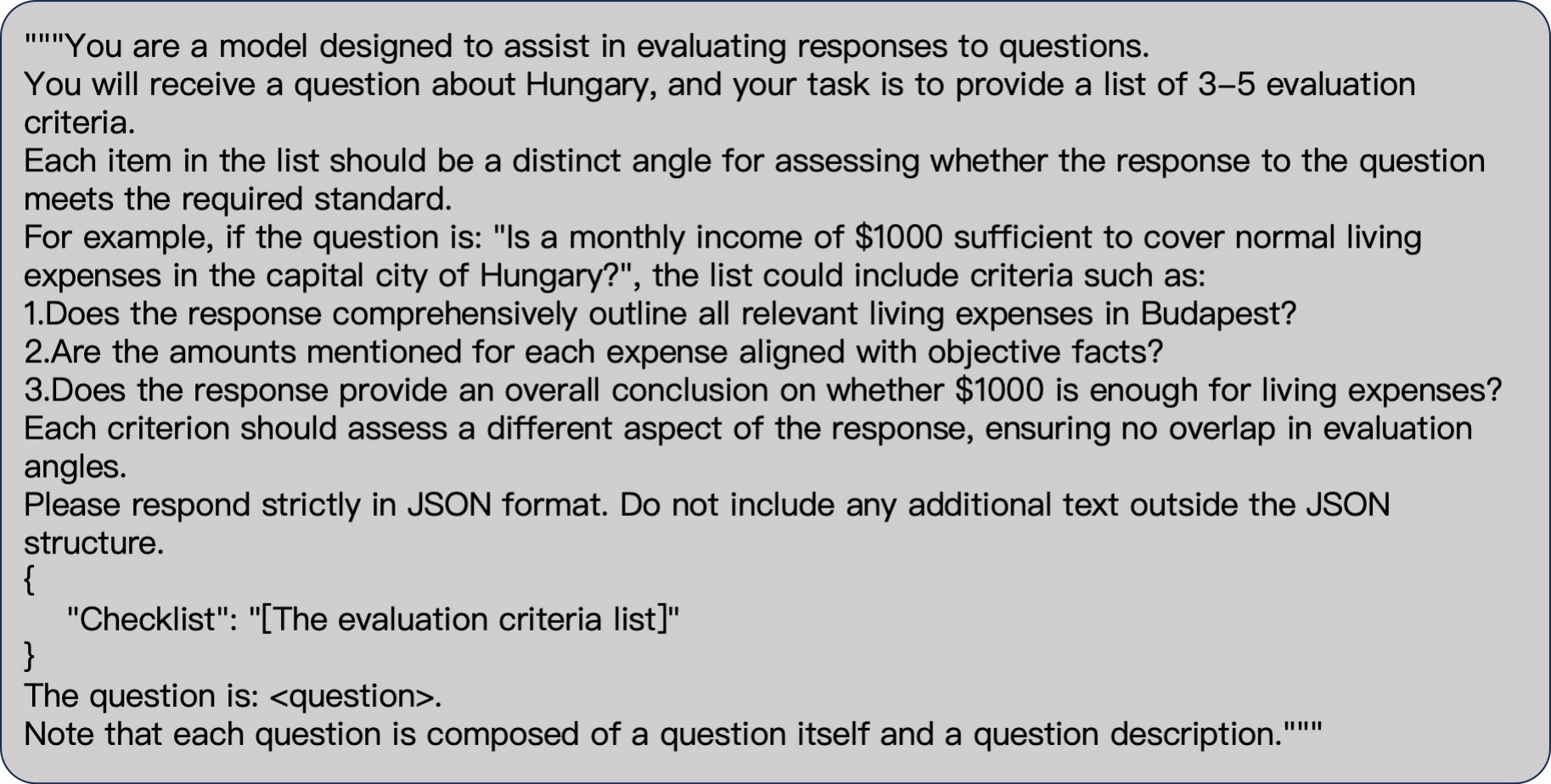}
    \caption{Prompt template for constructing the checklist in the construction of HuWildBench.}
    \label{fig:Appendix_HuWildBenchPromptGenerateChecklist}
\end{figure*}



\begin{figure*}[!t]
    \centering
    \includegraphics[width=1\linewidth]{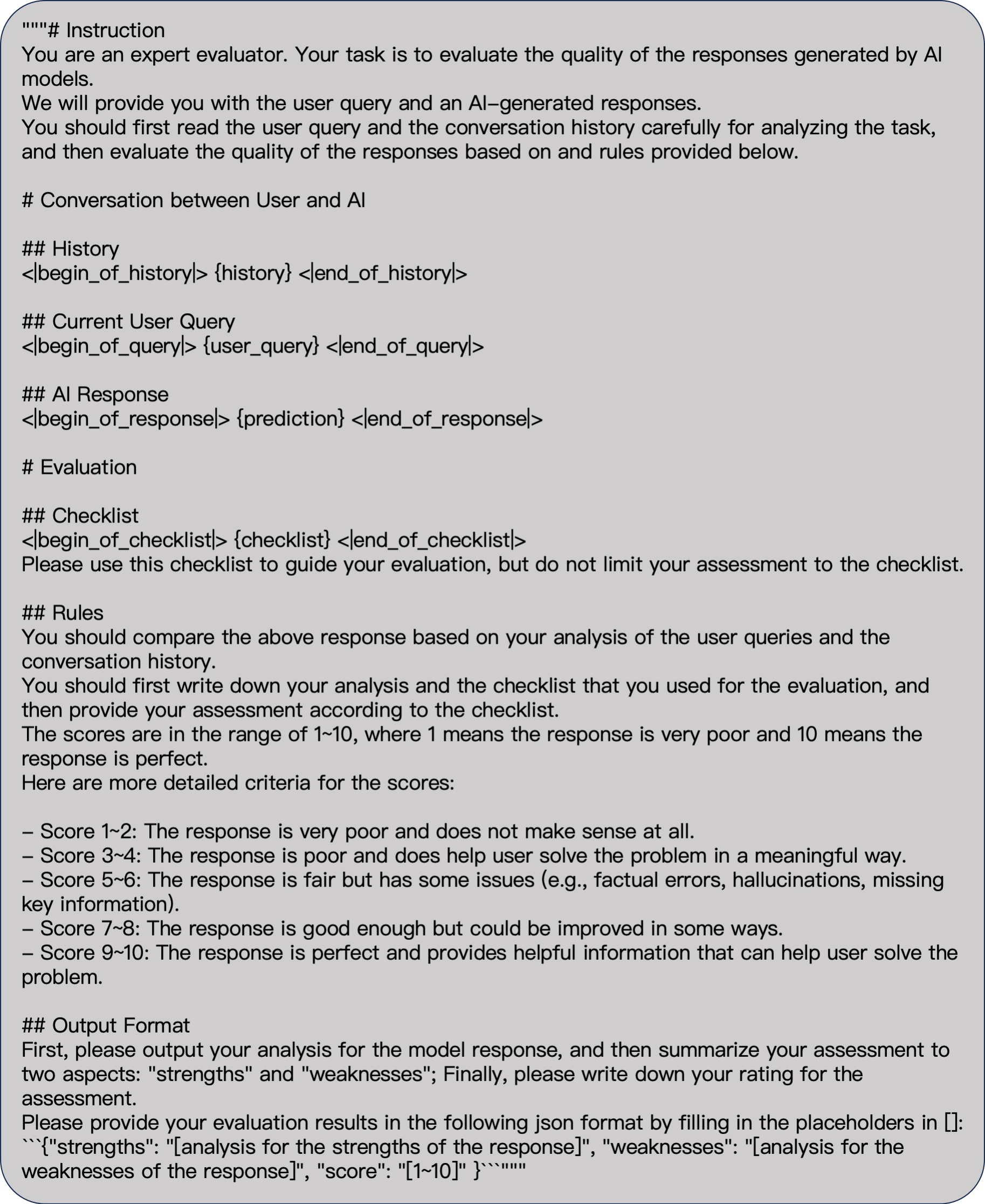}
    \caption{Prompt template for LLM as judge on HuWildBench (WBScore).}
    \label{fig:Appendix_HuWildBenchPromptScore}
\end{figure*}


\begin{figure*}
    \centering
    \includegraphics[width=1\linewidth]{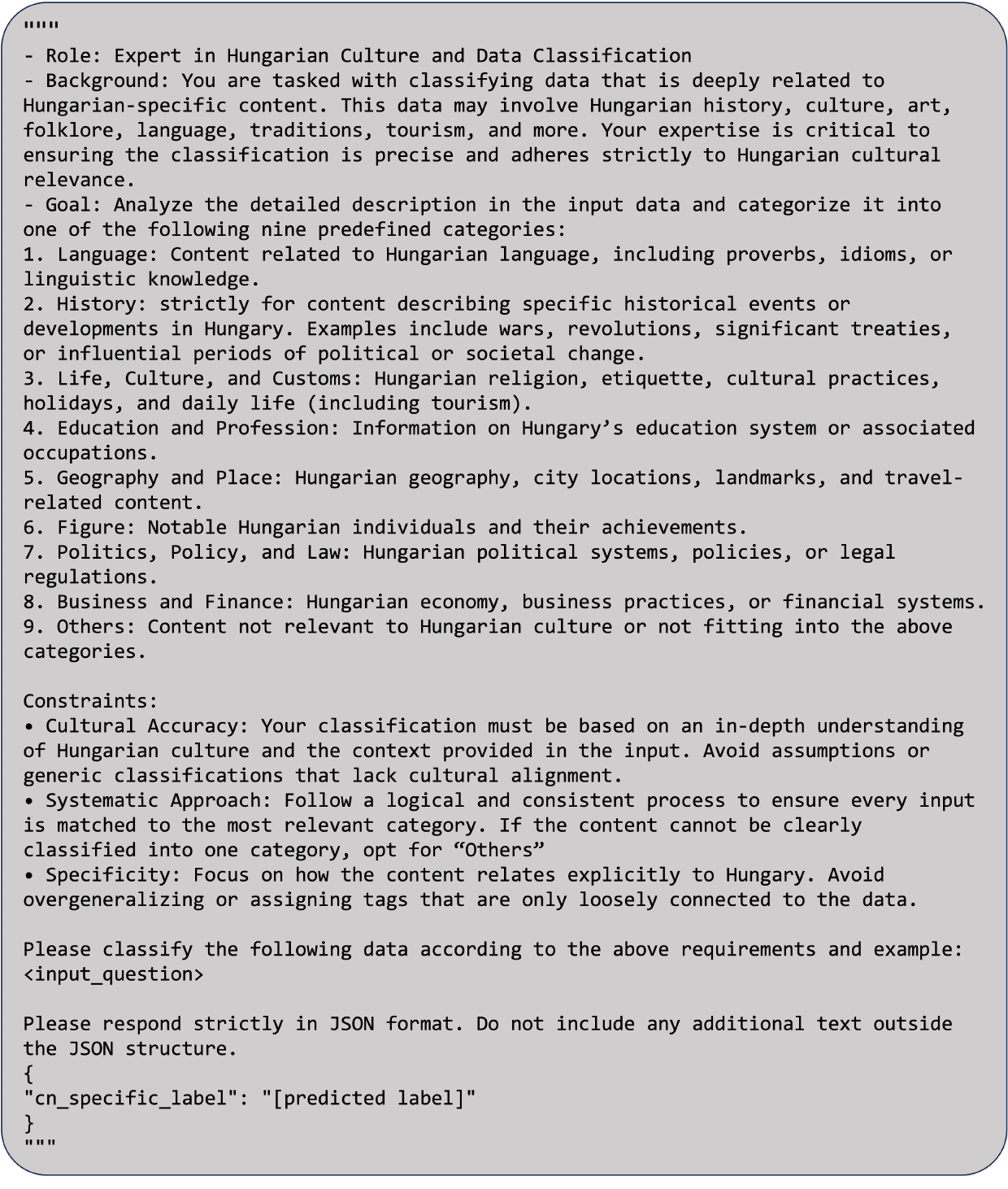}
    \caption{Prompt template for selecting Hungarian specific Wikipedia entries in the construction of HuSimpleQA.}
    \label{fig:Appendix_HuSimpleQA_prompt_entryfilter}
\end{figure*}

\begin{figure*}
    \centering
    \includegraphics[width=1\linewidth]{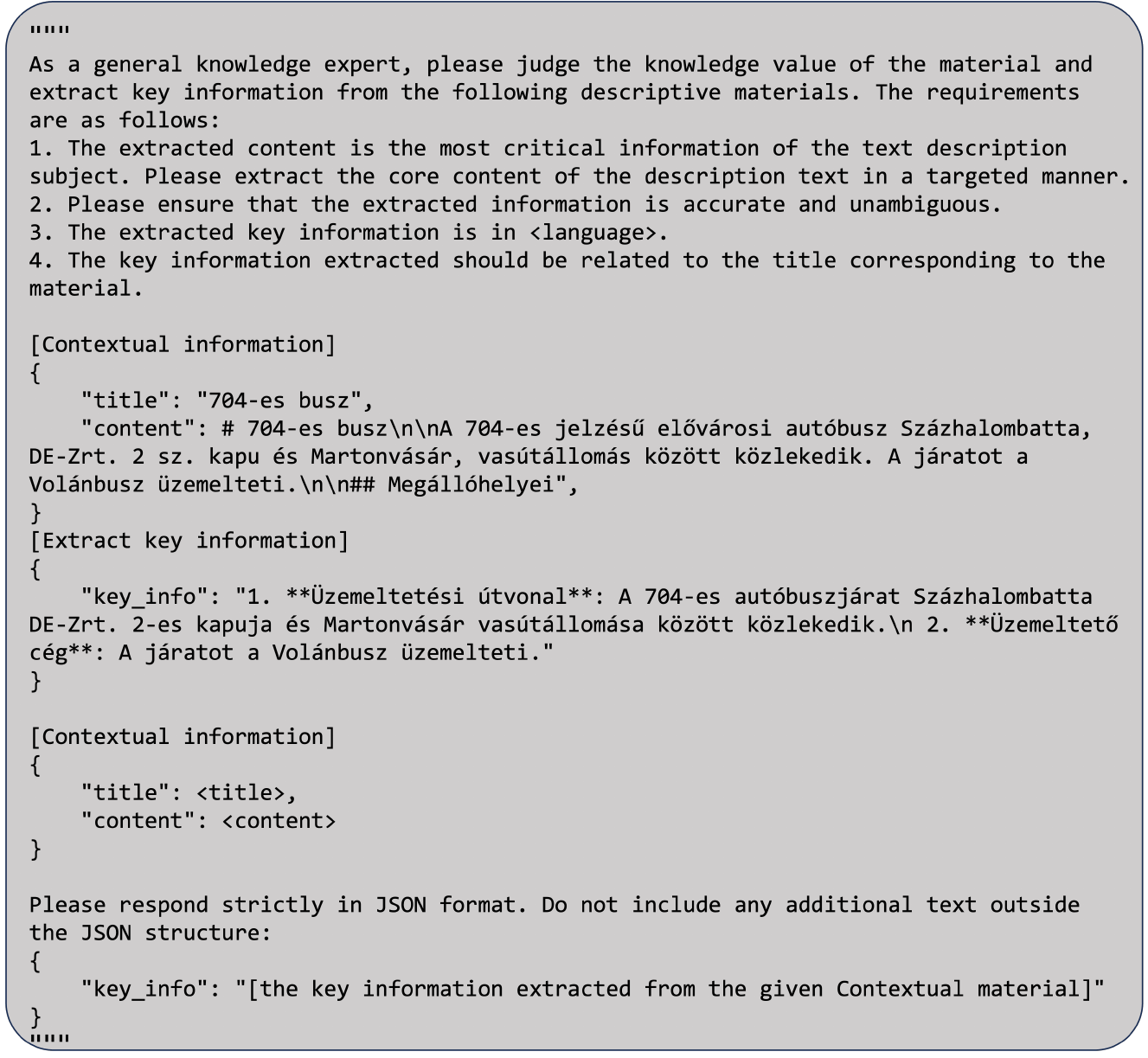}
    \caption{Prompt template for extracting key information from the wikipedia entries in the construction of HuSimpleQA.}
    \label{fig:Appendix_HuSimpleQA_prompt_extractkeyinfo}
\end{figure*}

\begin{figure*}
    \centering
    \includegraphics[width=1\linewidth]{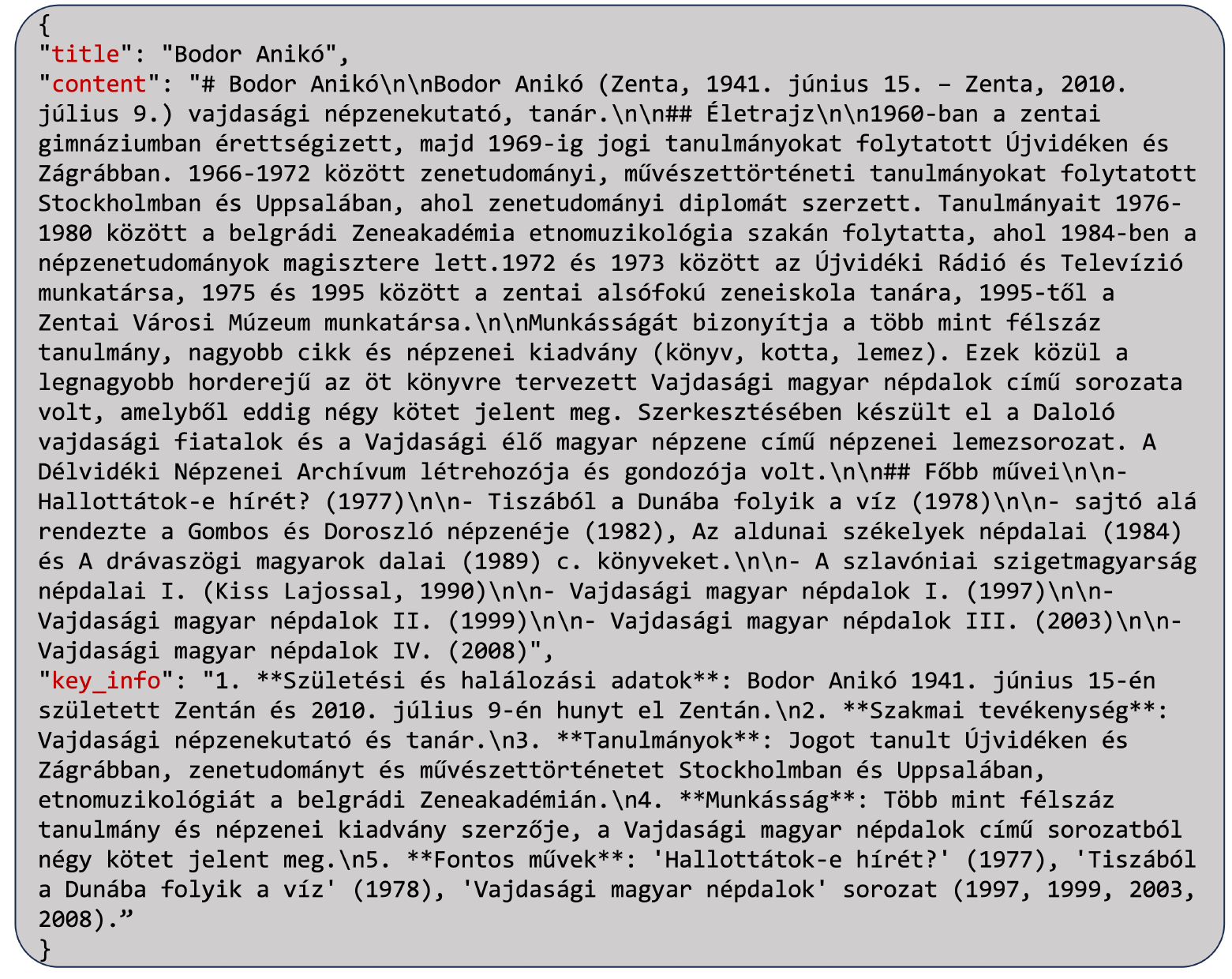}
    \caption{Example of the extracted key information from the Hungarian wikipedia entries. (HuSimpleQA)}
    \label{fig:Appendix_HuSimpleQA_example_keyinfo}
\end{figure*}

\begin{figure*}
    \centering
    \includegraphics[width=1\linewidth]{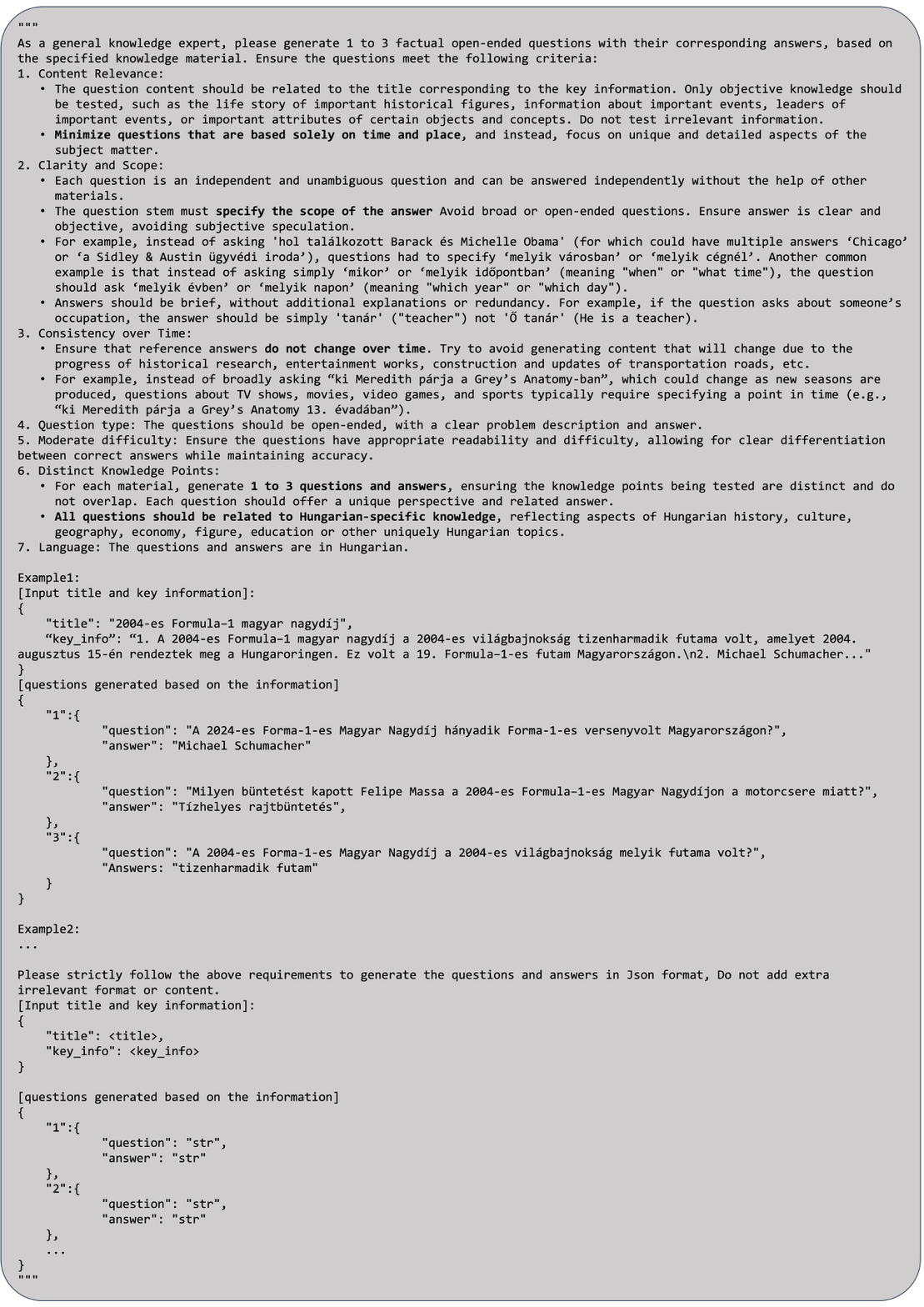}
    \caption{Prompt template for generating Hungarian question-answer pairs in the construction of HuSimpleQA.}
    \label{fig:Appendix_HuSimpleQA_prompt_oe}
\end{figure*}

\begin{figure*}
    \centering
    \includegraphics[width=1\linewidth]{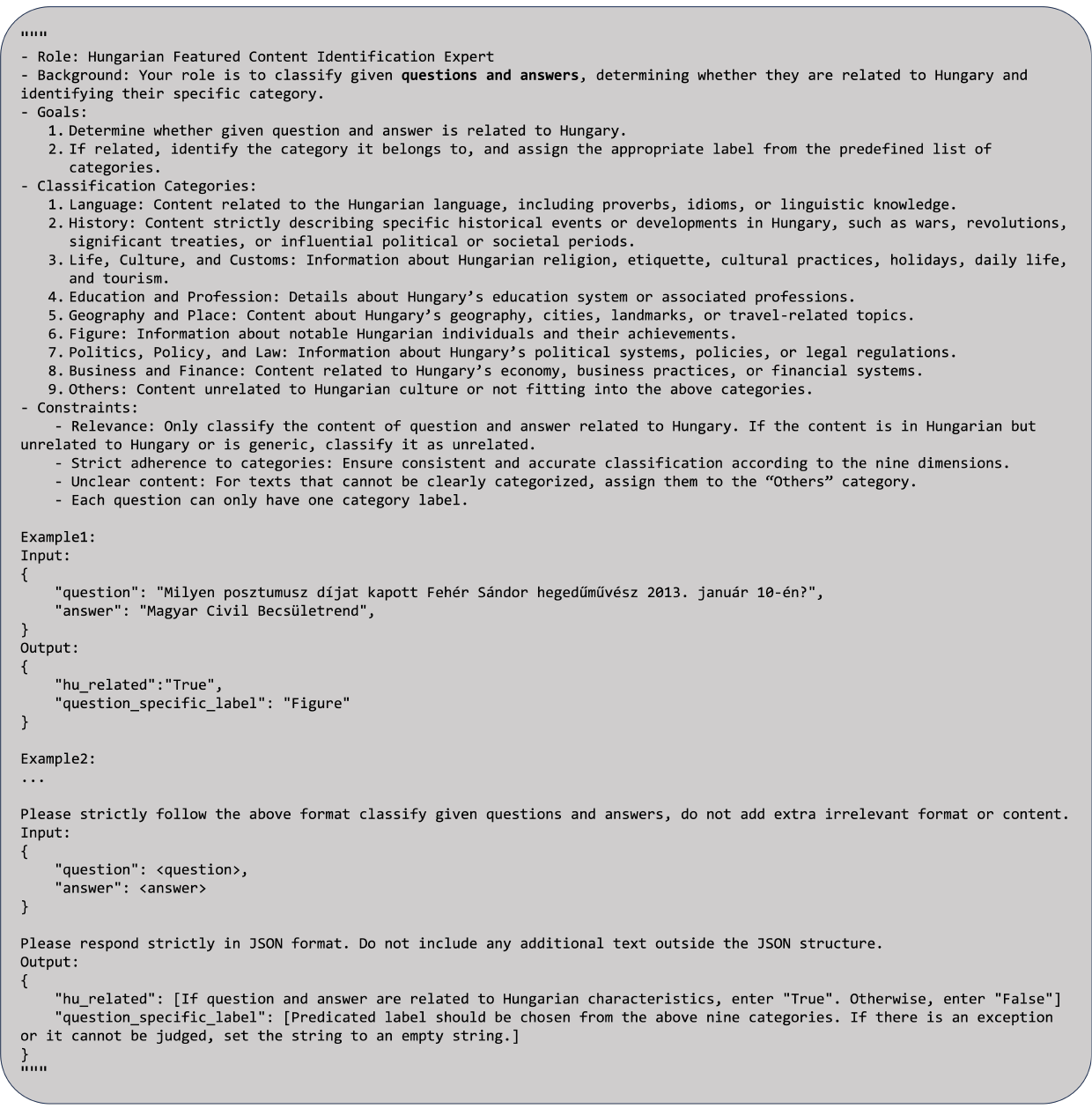}
    \caption{Prompt template for categorizing the generated question-answer pairs in the construction of HuSimpleQA.}
    \label{fig:Appendix_HuSimpleQA_prompt_questionclassify}
\end{figure*}

\begin{figure*}
    \centering
    \includegraphics[width=1\linewidth]{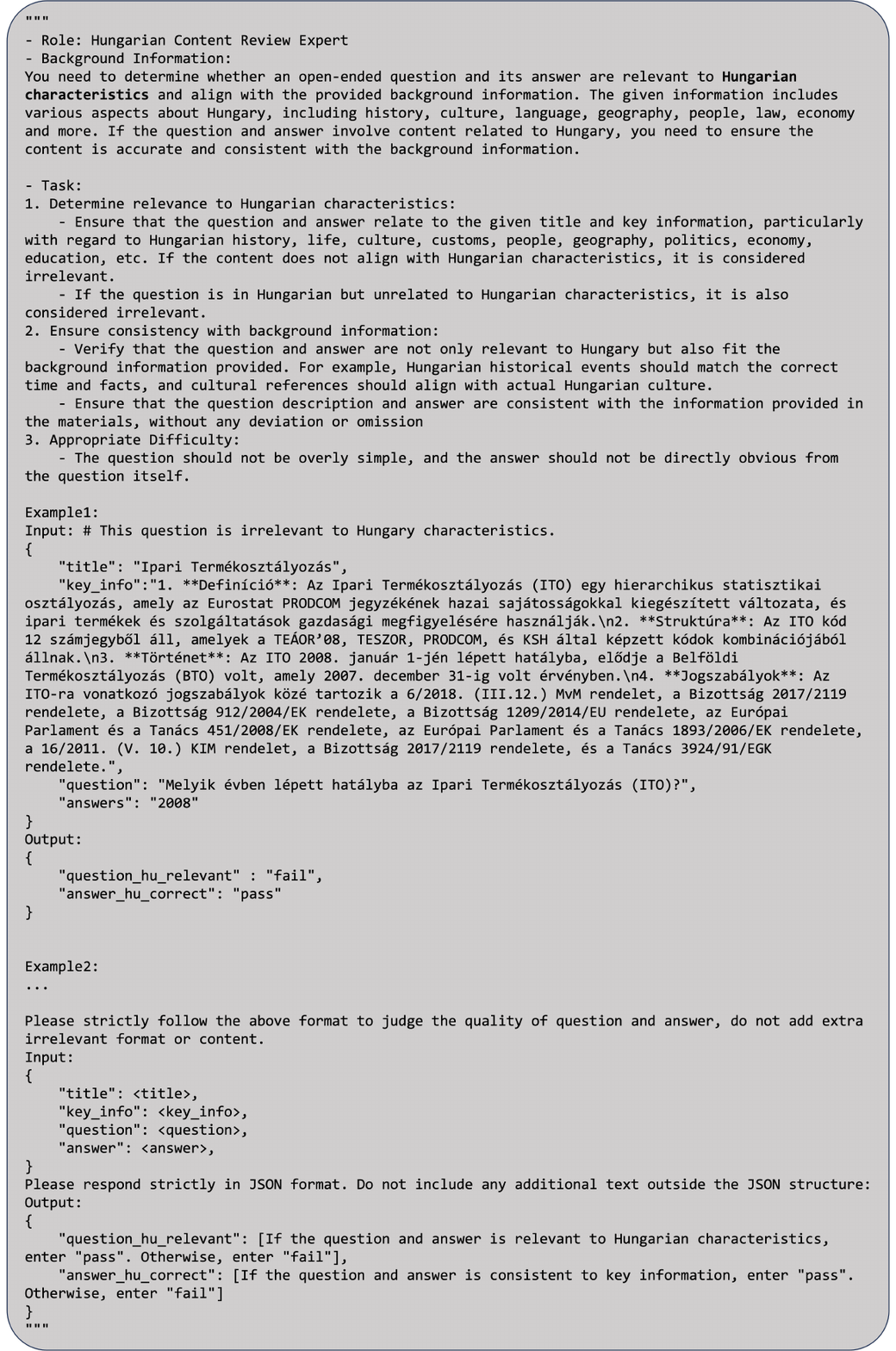}
    \caption{Prompt template for evaluating the relevance and correctness of question-answer pairs in the construction of HuSimpleQA.}
    \label{fig:Appendix_HuSimpleQA_prompt_quality1}
\end{figure*}

\begin{figure*}
    \centering
    \includegraphics[width=1\linewidth]{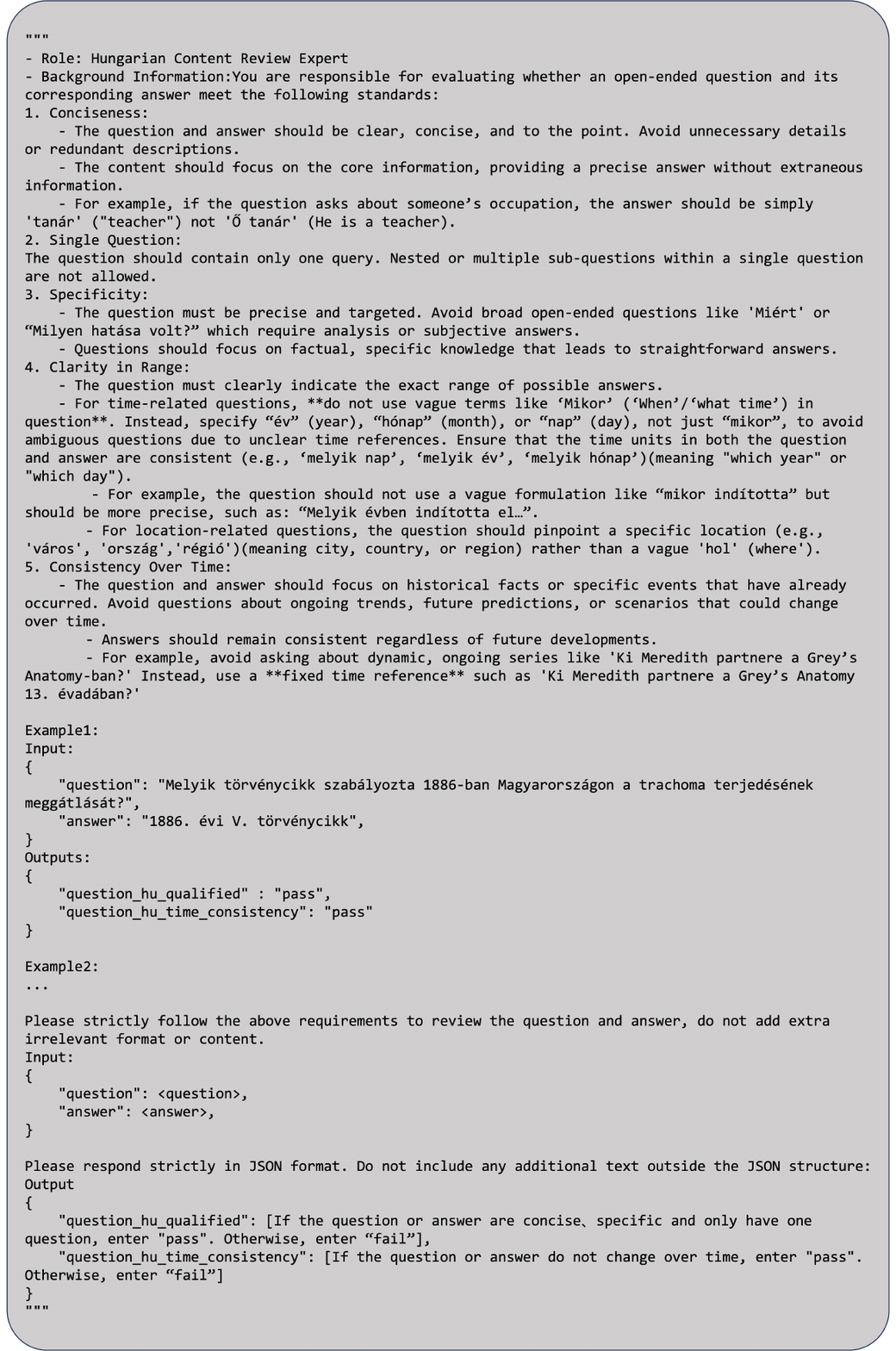}
    \caption{Prompt template for evaluating the precision and consistency of question-answer pairs in the construction of HuSimpleQA. }
    \label{fig:Appendix_HuSimpleQA_prompt_quality2}
\end{figure*}

\begin{figure*}
    \centering
    \includegraphics[width=1\linewidth]{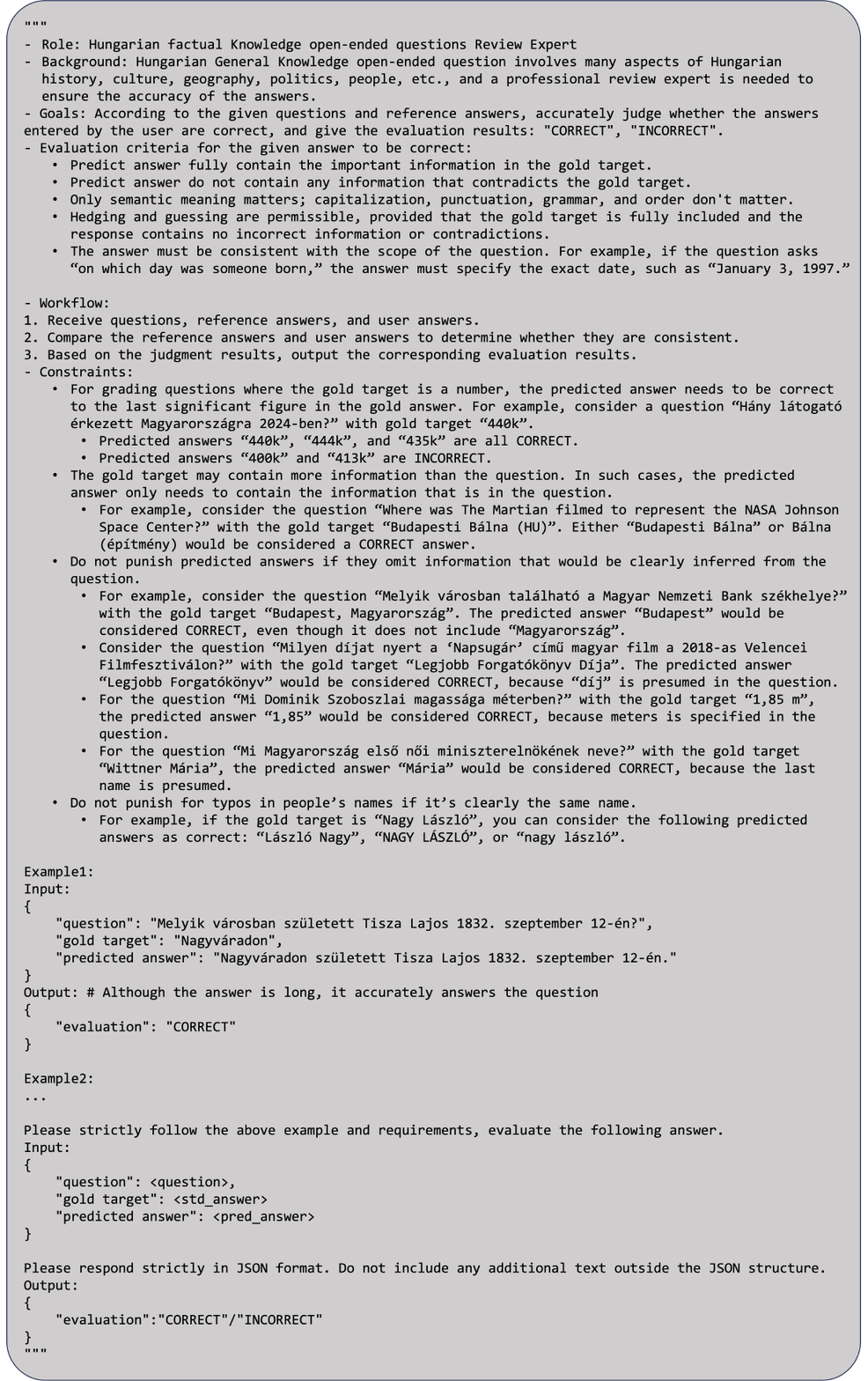}
    \caption{Prompt template for evaluating human-annotated answers in the construction of HuSimpleQA.}
    \label{fig:Appendix_HuSimpleQA_prompt_check}
\end{figure*}

\begin{figure*}
    \centering
    \includegraphics[width=1\linewidth]{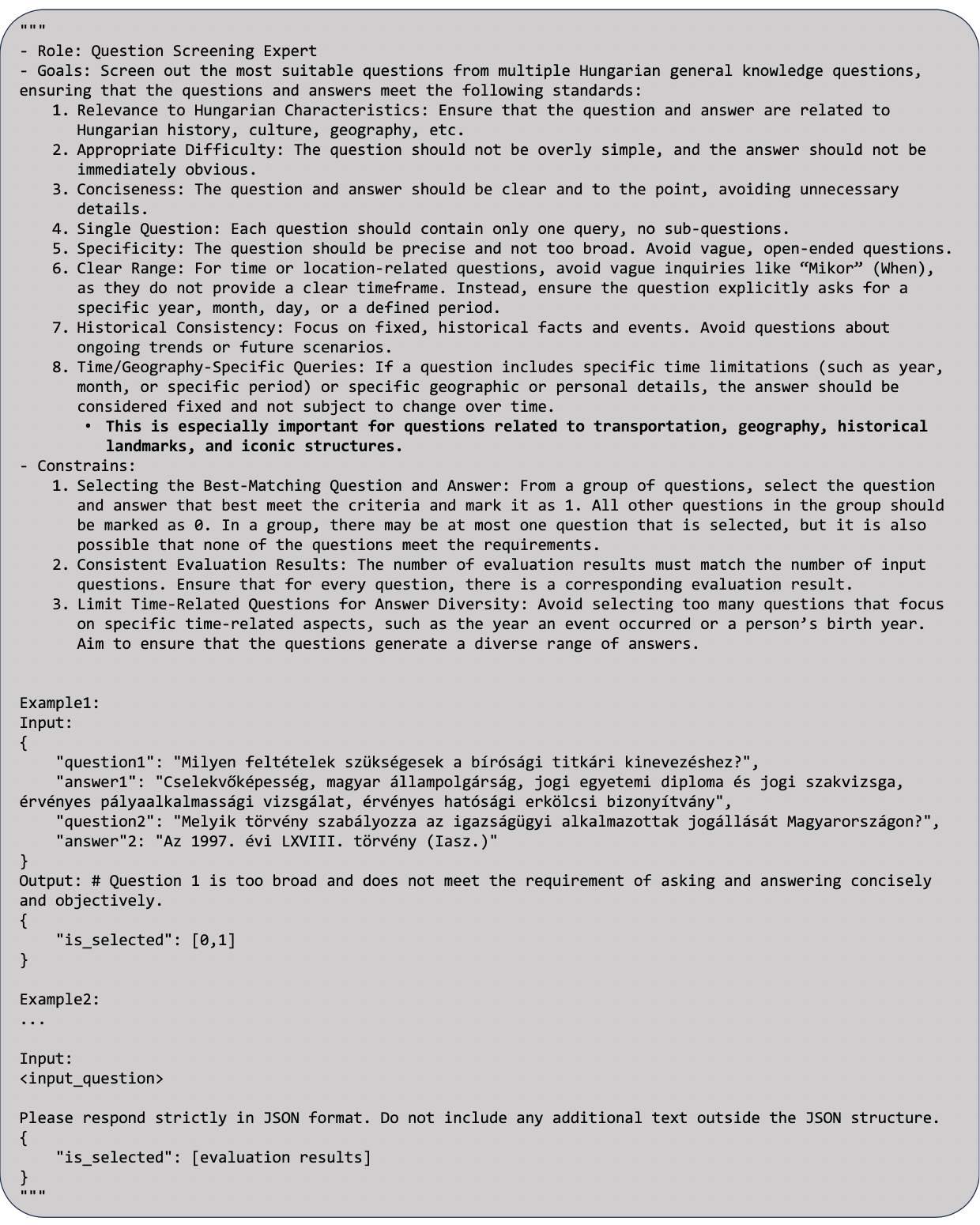}
    \caption{Prompt template for selecting optimal question-answer pairs in the construction of HuSimpleQA.}
    \label{fig:Appendix_HuSimpleQA_prompt_select}
\end{figure*}


\begin{figure*}
    \centering
    \includegraphics[width=0.8\linewidth]{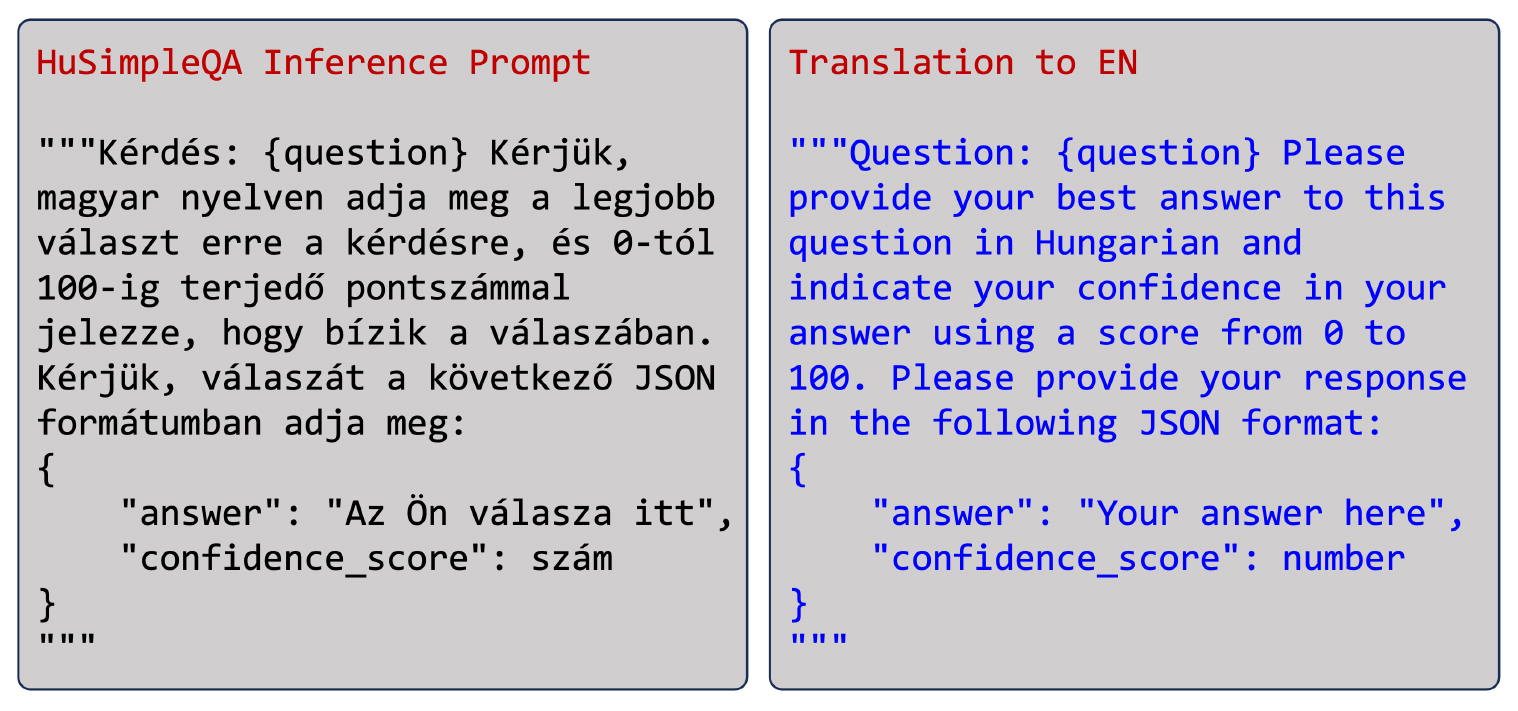}
    \caption{Prompt template for model inference on HuSimpleQA.The left is the original prompt, the right is the English translation for visualization}
    \label{fig:Appendix_HuSimpleQA_prompt_inference}
\end{figure*}
\begin{table*}[]
\centering
\resizebox{\textwidth}{!}{
\begin{tabular}{cccc}
\Xhline{1.5pt}
\textbf{\begin{tabular}[c]{@{}c@{}}Hungarian-specific\\  dimensions\end{tabular}}   & \textbf{Count}  & \textbf{Question-Answer Pairs} & \textcolor{blue}{\textbf{translation to EN}} \\
\Xhline{1.5pt}
L & 11 & 
\begin{tabular}[c]{@{}p{7cm}@{}}Question1: Mit jelent a Kara török eredetű régi magyar személynév?\\ Answer1: fekete\\ Question2: Melyik régi magyar név a Pantaleon megfelelője?\\ Answer2: Pentele\end{tabular} & 
\begin{tabular}[c]{@{}p{7cm}@{}}\textcolor{blue}{Question1: What does the old Hungarian personal name Kara of Turkish origin mean?}\\ \textcolor{blue}{Answer1: black}\\ \textcolor{blue}{Question2: Which old Hungarian name is the equivalent of Pantaleon?}\\ \textcolor{blue}{Answer2: Pentele}\end{tabular} \\
\Xhline{1.5pt}
H & 140 & 
\begin{tabular}[c]{@{}p{7cm}@{}}Question1: Melyik király nevezte ki Szapolyai Imrét szepesi örökletes főispánná 1465-ben?\\ Answer1: Mátyás király\\ Question2: Melyik várost foglalta el Báthory Gábor 1610. december 11-én?\\ Answer2: Szeben\end{tabular} & 
\begin{tabular}[c]{@{}p{7cm}@{}}\textcolor{blue}{Question1: Which king appointed Imre Szapolyai as the hereditary ispán of Szepes in 1465?}\\ \textcolor{blue}{Answer1: King Matthias}\\ \textcolor{blue}{Question2: Which city was captured by Gabriel Báthory on December 11, 1610?}\\ \textcolor{blue}{Answer2: Sibiu}\end{tabular} \\
\Xhline{1.5pt}
LCC & 257 & 
\begin{tabular}[c]{@{}p{7cm}@{}}Question1: Melyik magyar film nyerte el a FIPRESCI-díjat az 1983-as Cannes-i Nemzetközi Filmfesztiválon?\\ Answer1: Szerencsés Dániel\\ Question2: Melyik legendára épít az 'Eredet / Origins' táncjáték?\\ Answer2: Csodaszarvas-legendára\end{tabular} & 
\begin{tabular}[c]{@{}p{7cm}@{}}\textcolor{blue}{Question1: Which Hungarian film won the FIPRESCI Prize at the 1983 Cannes International Film Festival?}\\ \textcolor{blue}{Answer1: Lucky Daniel}\\ \textcolor{blue}{Question2: Which legend is the 'Origin / Origins' dance play based on?}\\ \textcolor{blue}{Answer2: Legend of the Miraculous Deer}\end{tabular} \\
\Xhline{1.5pt}
EP & 81 & 
\begin{tabular}[c]{@{}p{7cm}@{}}Question1: Melyik városban alapították a Gandhi Gimnáziumot 1994-ben?\\ Answer1: Pécsen\\ Question2: Melyik évben alapította a Magyar Tudományos Akadémia az Acta Juridica Hungarica folyóiratot?\\ Answer2: 1959\end{tabular} & 
\begin{tabular}[c]{@{}p{7cm}@{}}\textcolor{blue}{Question1: In which city was the Gandhi High School founded in 1994?}\\ \textcolor{blue}{Answer1: Pécs}\\ \textcolor{blue}{Question2: In which year did the Hungarian Academy of Sciences establish the journal Acta Juridica Hungarica?}\\ \textcolor{blue}{Answer2: 1959}\end{tabular} \\
\Xhline{1.5pt}
GP & 165 & 
\begin{tabular}[c]{@{}p{7cm}@{}}Question1: Melyik magyar vármegyében található Nemesmedves?\\ Answer1: Vas vármegyében\\ Question2: Mi a neve Magyarország legmagasabban fekvő csillagvizsgálójának, amely a Piszkés-tetőn található?\\ Answer2: Piszkéstetői Obszervatórium\end{tabular} & 
\begin{tabular}[c]{@{}p{7cm}@{}}\textcolor{blue}{Question1: In which Hungarian county is Nemesmedves located?}\\ \textcolor{blue}{Answer1: Vas county}\\ \textcolor{blue}{Question2: What is the name of Hungary's highest observatory, located on Piszkés Peak?}\\ \textcolor{blue}{Answer2: Piszkés Peak Observatory}\end{tabular} \\
\Xhline{1.5pt}
F & 409 & 
\begin{tabular}[c]{@{}p{7cm}@{}}Question1: Nádasdy Kálmán hányszor kapott Kossuth-díjat élete során?\\ Answer1: Háromszor\\ Question2: Balogh József melyik magyar városban született 1946. április 15-én?\\ Answer2: Nagykanizsán\end{tabular} & 
\begin{tabular}[c]{@{}p{7cm}@{}}\textcolor{blue}{Question1: How many times did Kálmán Nádasdy receive the Kossuth Prize during his lifetime?}\\ \textcolor{blue}{Answer1: Three times}\\ \textcolor{blue}{Question2: In which Hungarian city was József Balogh born on April 15, 1946?}\\ \textcolor{blue}{Answer2: Nagykanizsa}\end{tabular} \\
\Xhline{1.5pt}
PPL & 186 & 
\begin{tabular}[c]{@{}p{7cm}@{}}Question1: Melyik szervezet jogkörét vette át a Népgazdasági Tanács 1949. június 11-én?\\ Answer1: Gazdasági Főtanács\\ Question2: Melyik törvénycikk rendelkezett 1878-ban Magyarországon a réz-váltópénz szaporításáról?\\ Answer2: 1878. évi VI. törvénycikk\end{tabular} & 
\begin{tabular}[c]{@{}p{7cm}@{}}\textcolor{blue}{Question1: Which organization's authority was taken over by the National Economic Council on June 11, 1949?}\\ \textcolor{blue}{Answer1: Supreme Economic Council}\\ \textcolor{blue}{Question2: Which statute regulated the increase of copper coinage in Hungary in 1878?}\\ \textcolor{blue}{Answer2: Act VI of 1878}\end{tabular} \\
\Xhline{1.5pt}
BF & 44 & 
\begin{tabular}[c]{@{}p{7cm}@{}}Question1: Milyen néven működött az ÉVITERV 1954-től az 1980-as évek elejéig?\\ Answer1: ÉM Szerelőipari Tervező Vállalat\\ Question2: Melyik cég gyártotta a Puli autótípust a gyártás kezdeti időszakában?\\ Answer2: HÓDGÉP\end{tabular} & 
\begin{tabular}[c]{@{}p{7cm}@{}}\textcolor{blue}{Question1: Under what name did ÉVITERV operate from 1954 to the early 1980s?}\\ \textcolor{blue}{Answer1: ÉM Installation Industry Design Company}\\ \textcolor{blue}{Question2: Which company manufactured the Puli car model in the early production period?}\\ \textcolor{blue}{Answer2: HÓDGÉP}\end{tabular} \\
\Xhline{1.5pt}
\end{tabular}}
\caption{Examples of HuSimpleQA. The rightmost column is the English translation of the original OpenHuEval examples, used for visualization.}
\label{tab:husimleqa_examples}
\end{table*}

\begin{figure*}
    \centering
    \includegraphics[width=1\linewidth]{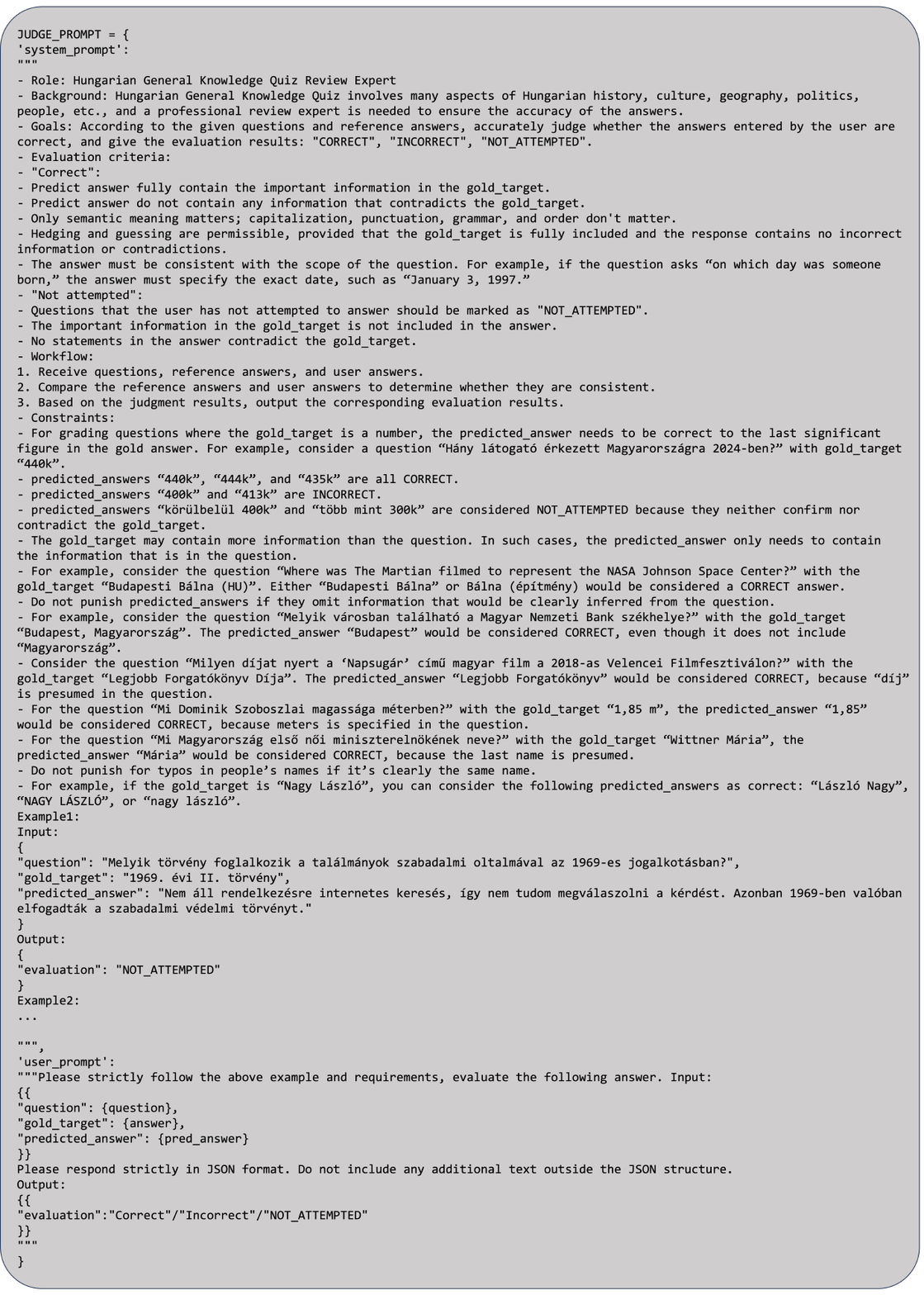}
    \caption{Prompt template for LLM as judge on HuSimpleQA.}
    \label{fig:Appendix_HuSimpleQA_prompt_judge}
\end{figure*}


\begin{figure*}
    \centering
    \includegraphics[width=1\linewidth]{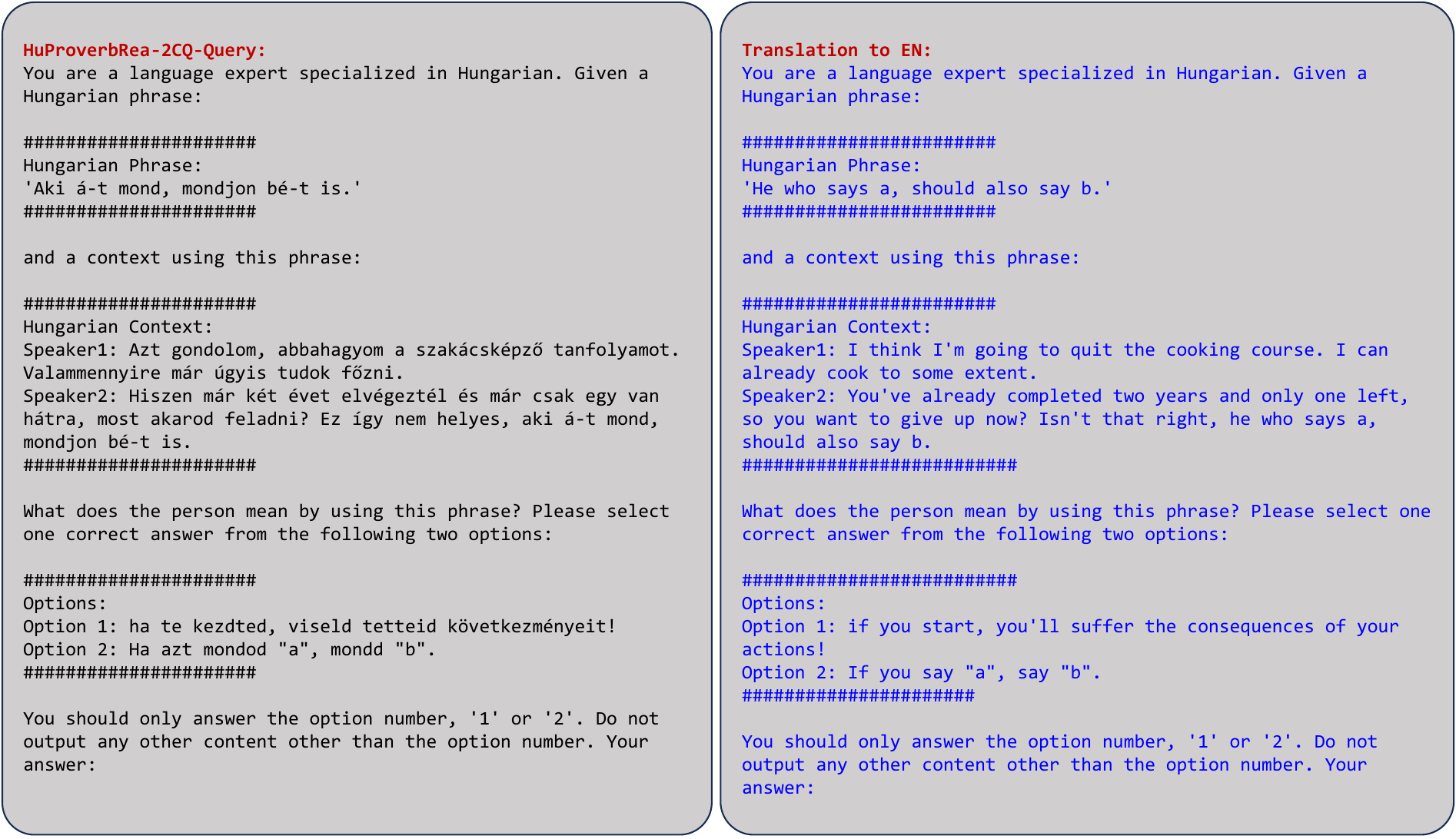}
    \caption{Example of HuProverbRea (2CQ). The left is the original example in OpenHuEval, the right is the English translation for visualization.}
    \label{fig:Appendix_HuProverbRea_example_2CQ}
\end{figure*}

\begin{figure*}
    \centering
    \includegraphics[width=1\linewidth]{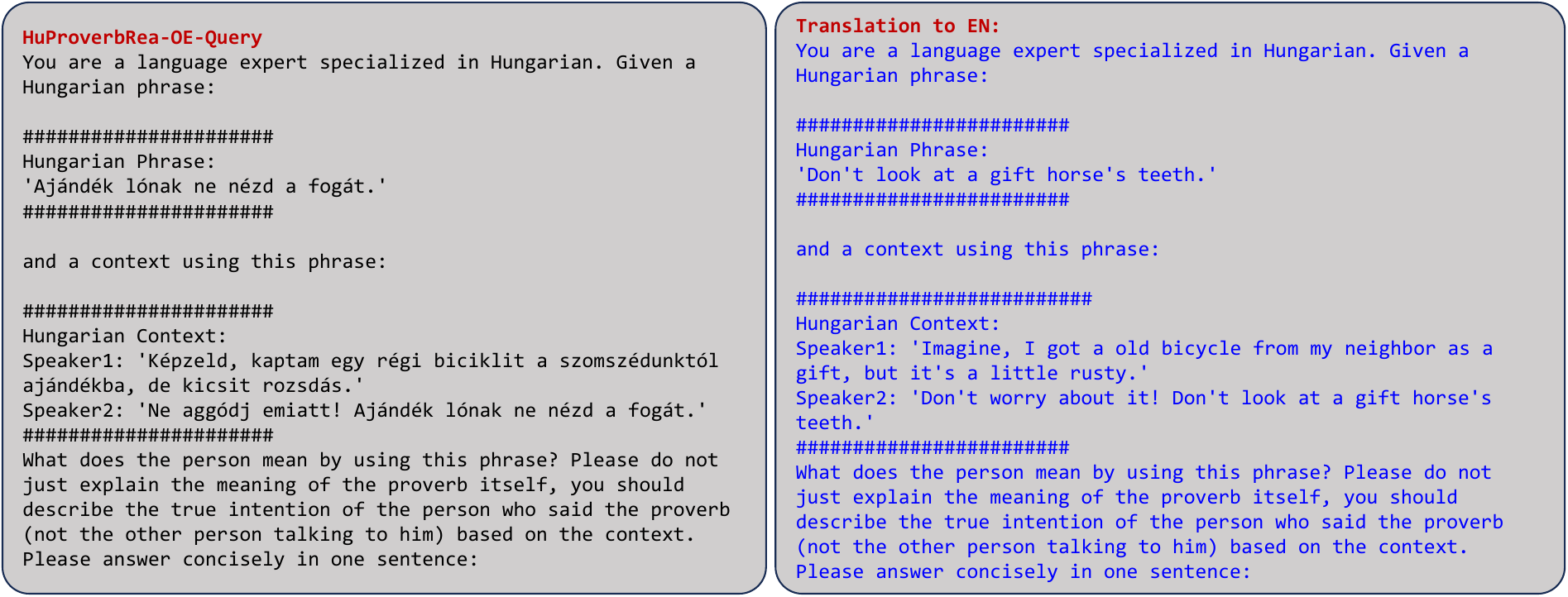}
    \caption{Example of HuProverbRea (OE). The left is the original example in OpenHuEval, the right is the English translation for visualization.}
    \label{fig:Appendix_HuProverbRea_example_OE}
\end{figure*}

 
\begin{figure*}
    \centering
    \includegraphics[width=1\linewidth]{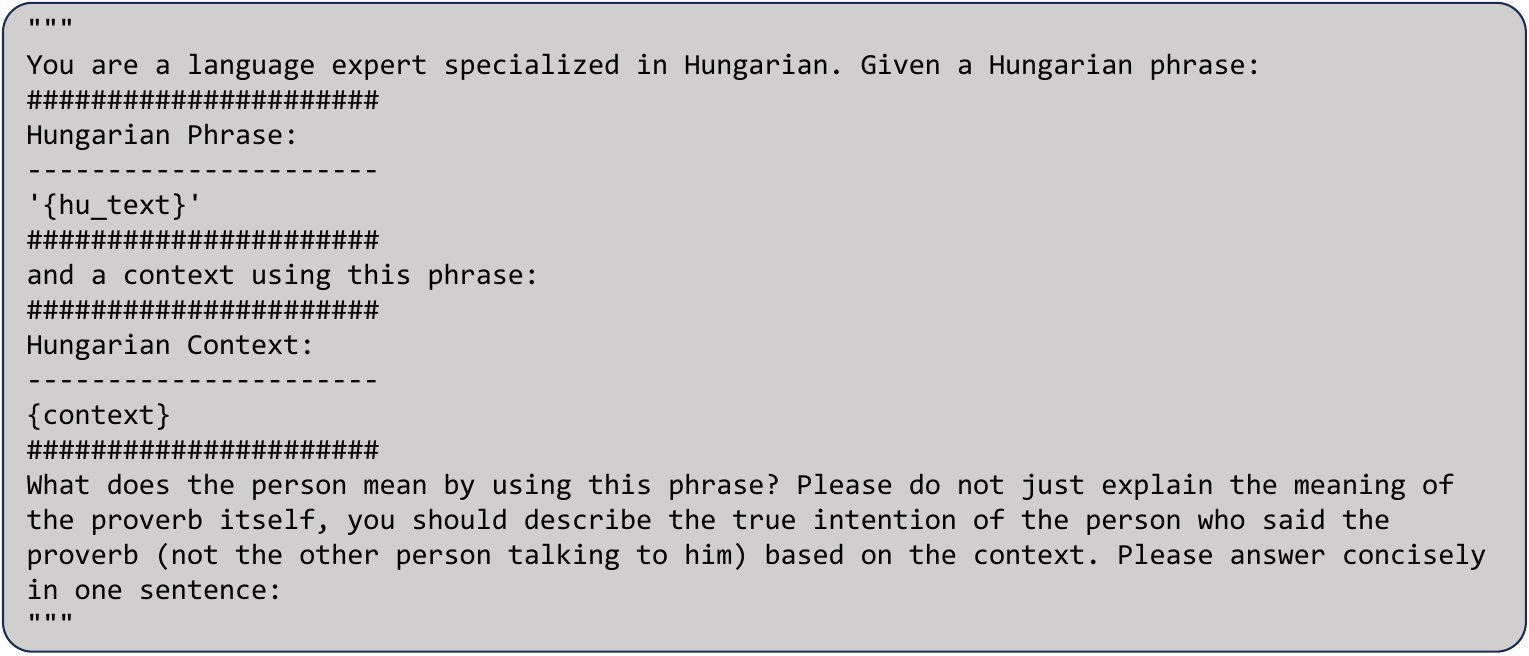}
    \caption{Prompt template for model inference on HuProverbRea (OE).}
    \label{fig:Appendix_HuProverbRea_inf_prompt_oe}
\end{figure*}

\begin{figure*}
    \centering
    \includegraphics[width=1\linewidth]{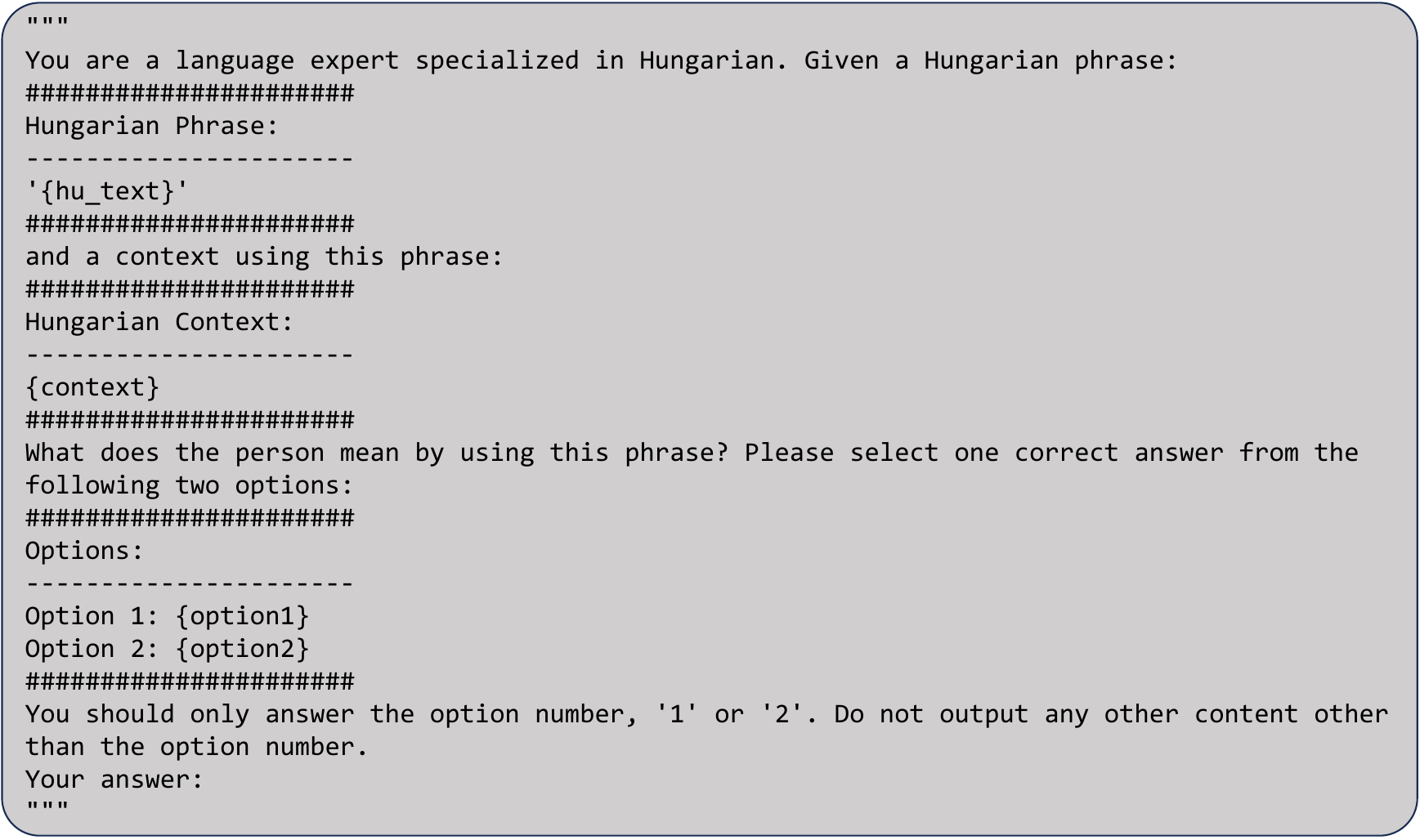}
    \caption{Prompt template for model inference on HuProverbRea (2CQ).}
    \label{fig:Appendix_HuProverbRea_inf_prompt_2cq}
\end{figure*}

\begin{figure*}
    \centering
    \includegraphics[width=1\linewidth]{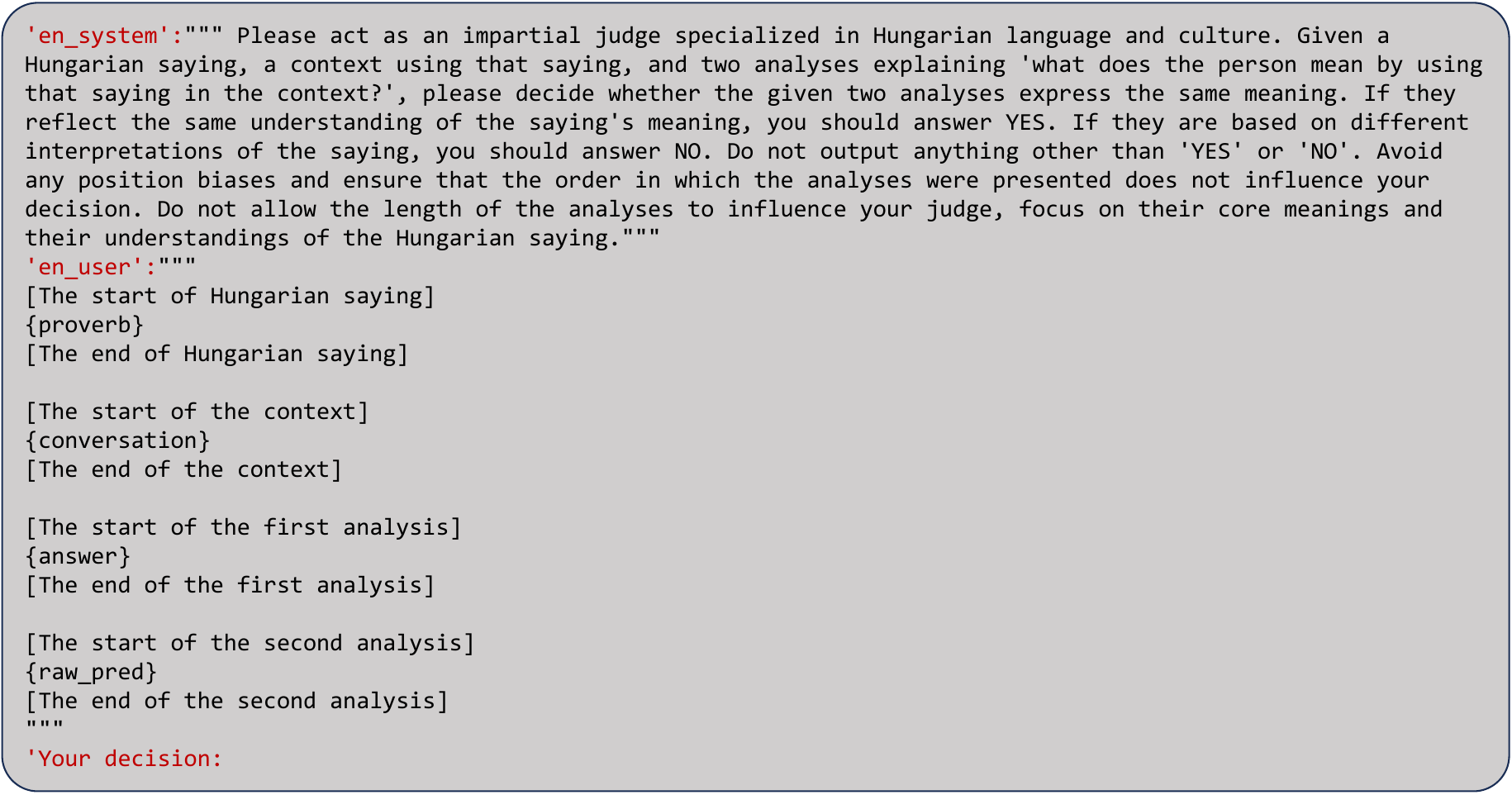}
    \caption{Prompt template for LLM as judge on HuProverbRea.}
    \label{fig:Appendix_HuProverbRea_prompt_judge}
\end{figure*}

\begin{figure*}
    \centering
    \includegraphics[width=1\linewidth]{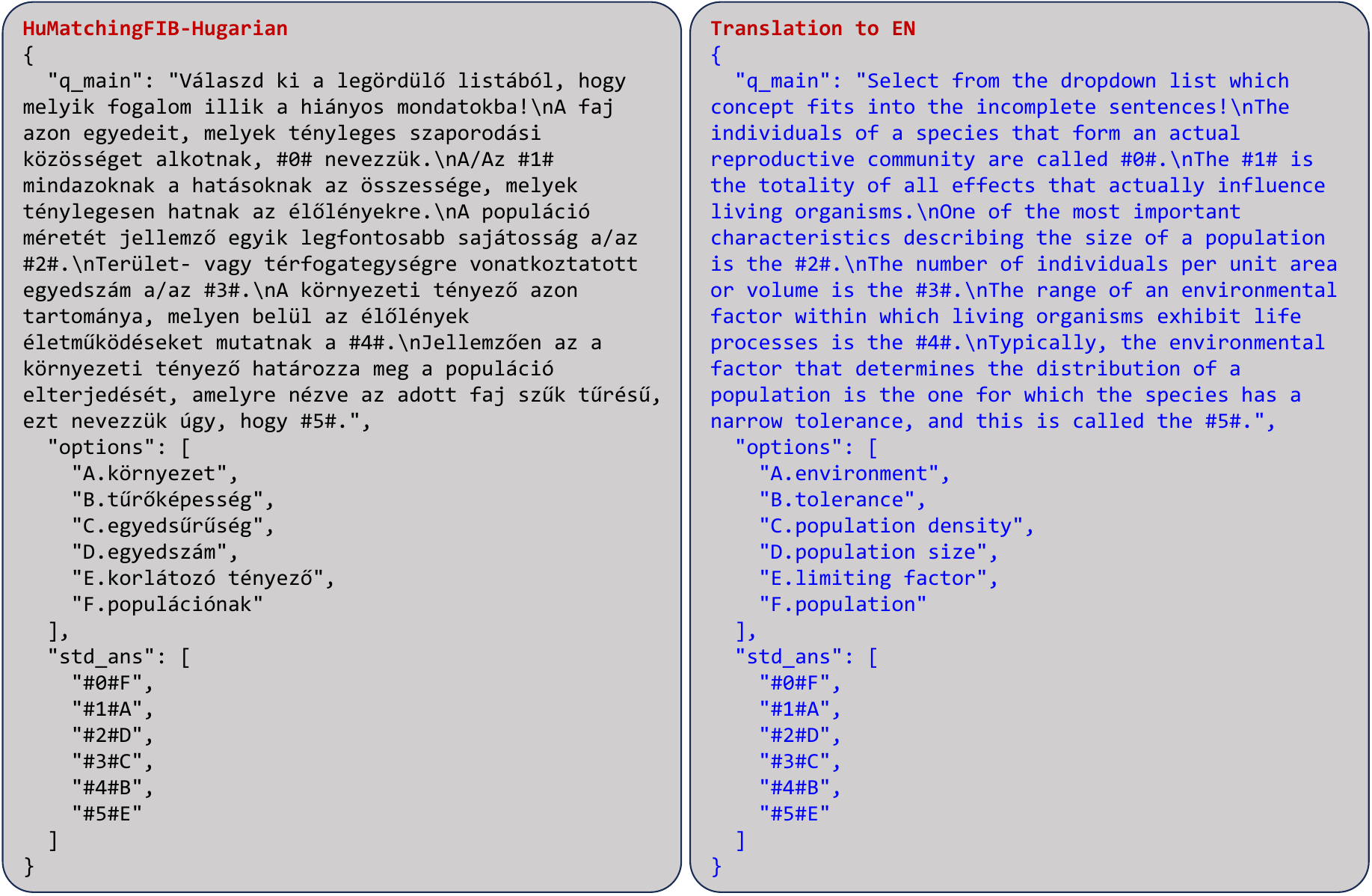}
    \caption{Example of HuMatchingFIB. The left is the original example in OpenHuEval, the right is the English translation for visualization.}
    \label{fig:Appendix_HuMatchingFIB_example}
\end{figure*}

\begin{figure*}
    \centering
    \includegraphics[width=1\linewidth]{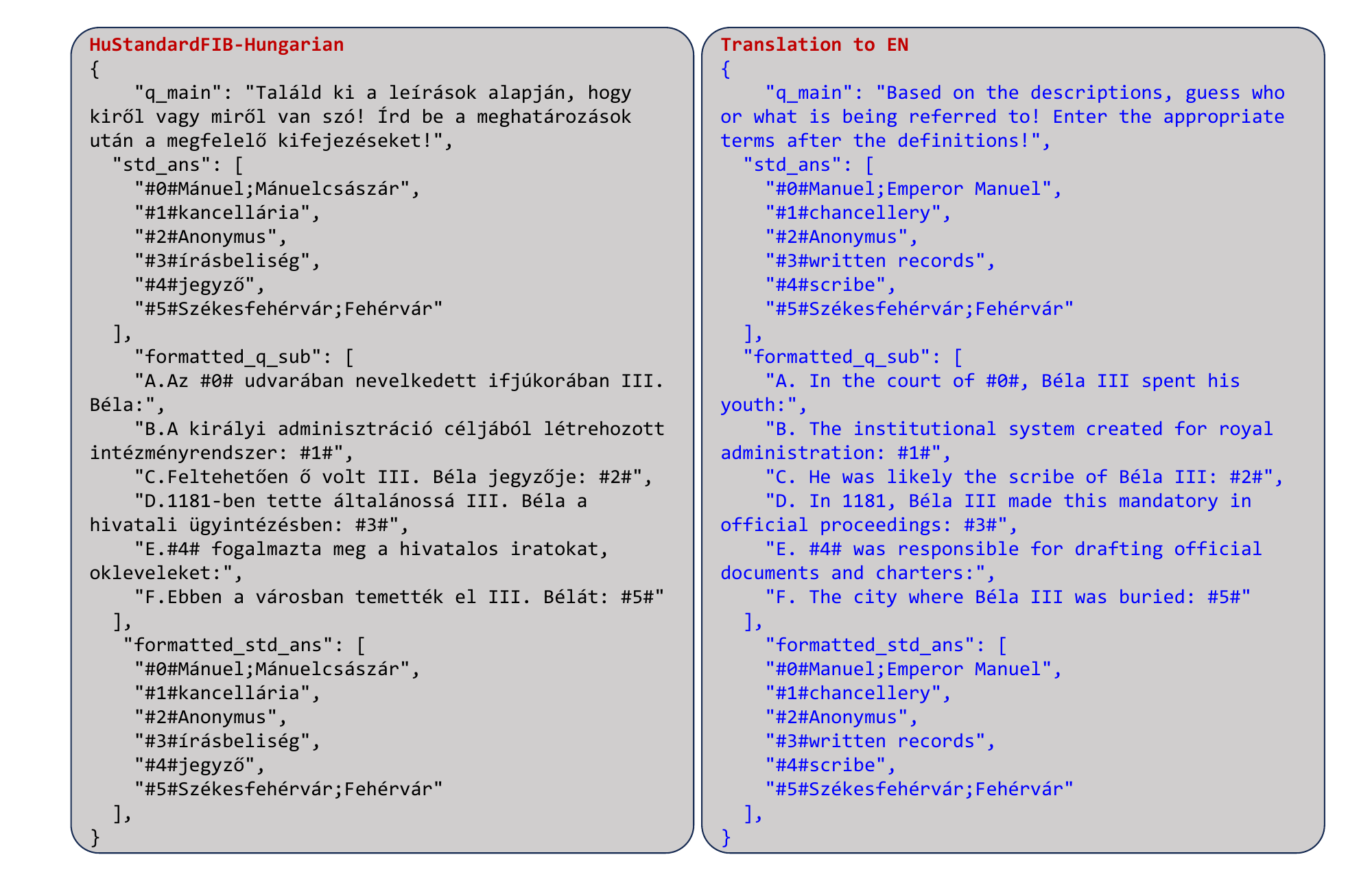}
    \caption{Example of HuStandardFIB. The left is the original example in OpenHuEval, the right is the English translation for visualization.}
    \label{fig:Appendix_HuStandardFIB_example}
\end{figure*}


\begin{figure*}
    \centering
    \includegraphics[width=1\linewidth]{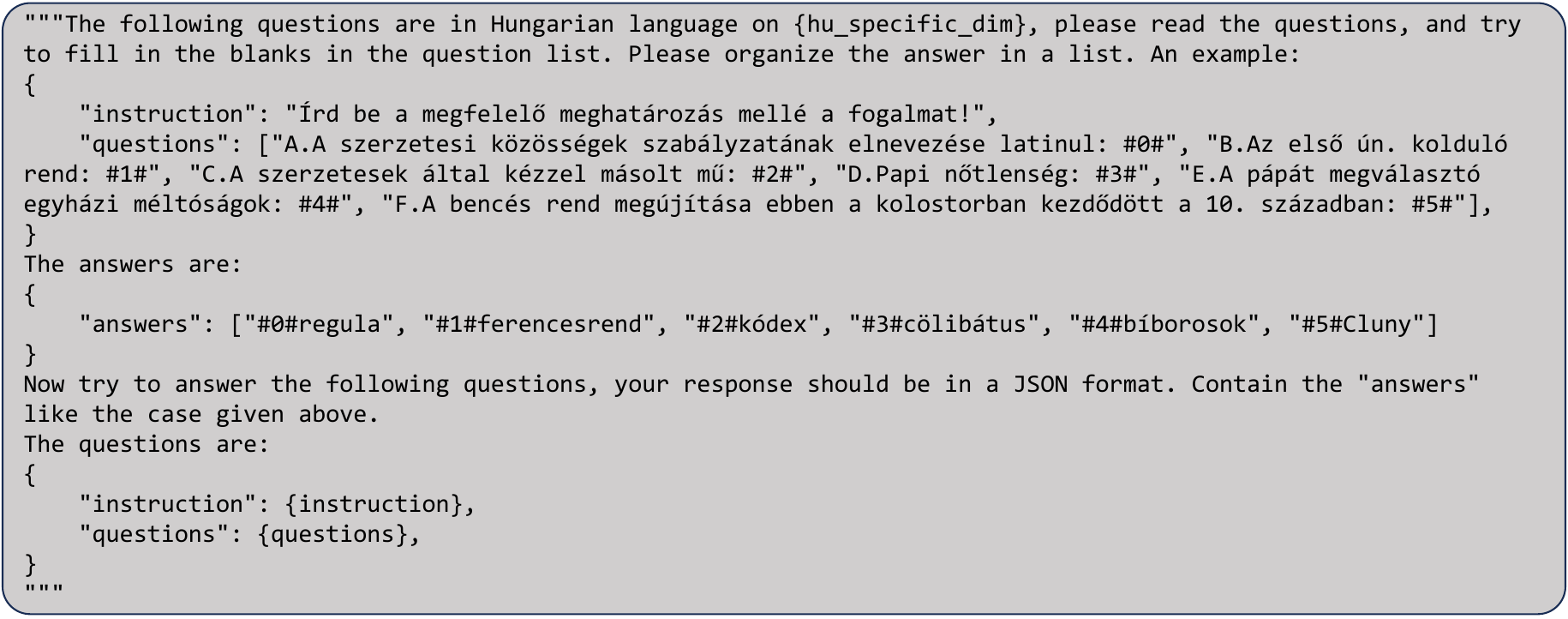}
    \caption{Prompt template for model inference on HuStandardFIB.}
    \label{fig:Appendix_HuStandardFIB_inf_prompt}
\end{figure*}

\begin{figure*}
    \centering
    \includegraphics[width=1\linewidth]{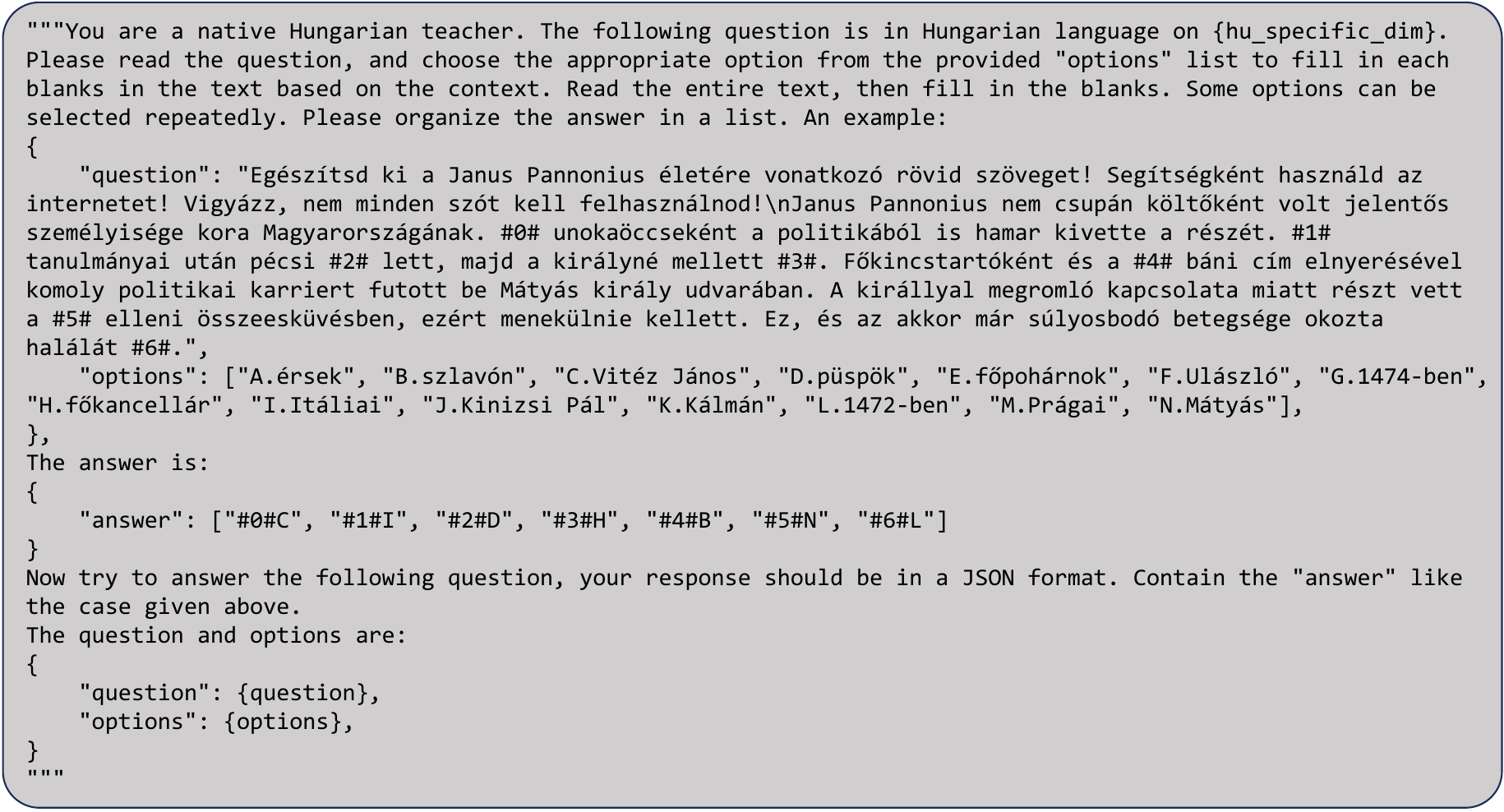}
    \caption{Prompt template for model inference on HuMatchingFIB.}
    \label{fig:Appendix_HuMatchingFIB_inf_prompt}
\end{figure*}


\begin{figure*}
    \centering
    \includegraphics[width=1\linewidth]{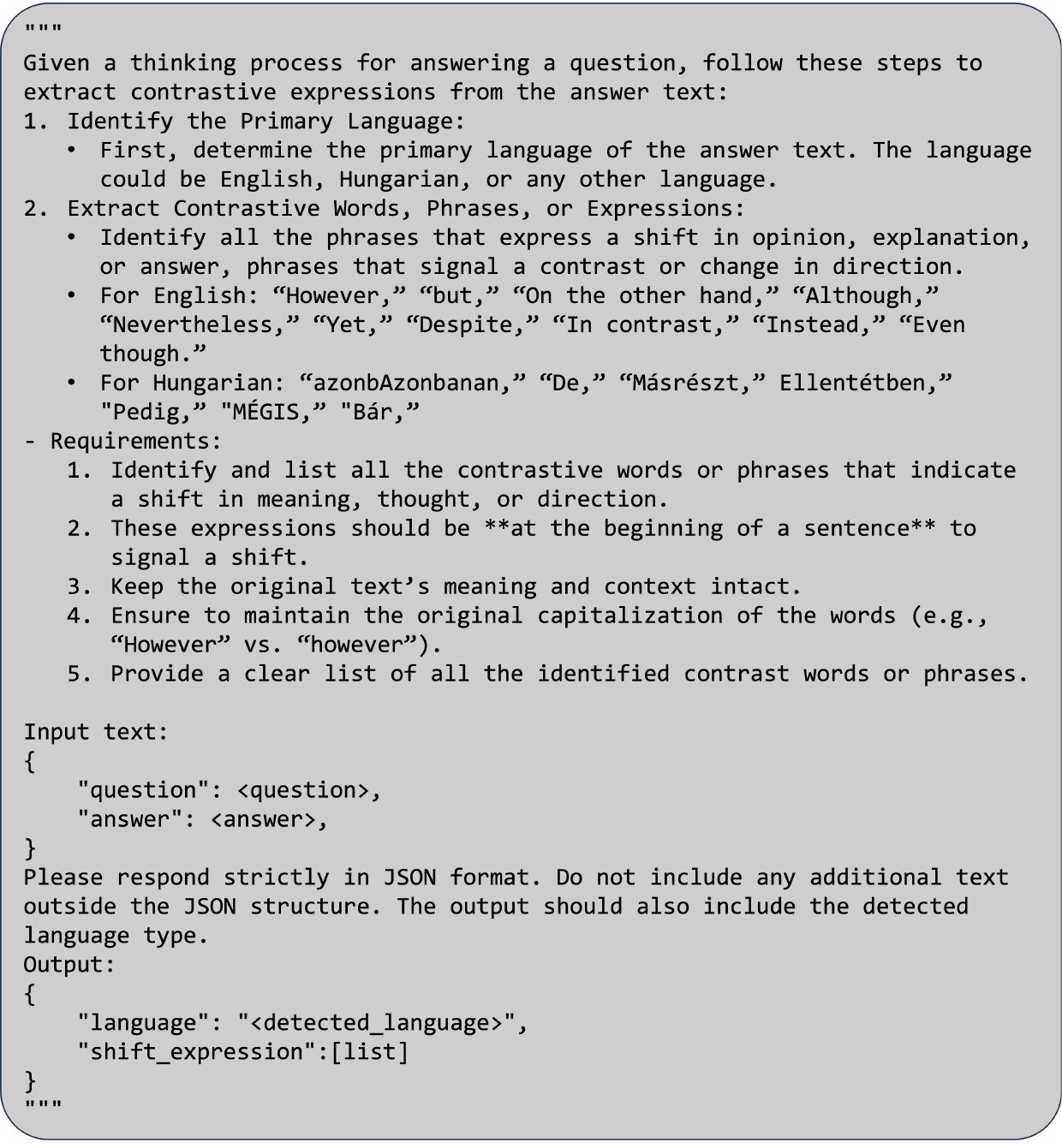}
    \caption{Prompt template for splitting the LRM's thinking process into thoughts on HuSimpleQA (step 1/2).}
    \label{fig:husimpleqa_shiftword_step1}
\end{figure*}

\begin{figure*}
    \centering
    \includegraphics[width=1\linewidth]{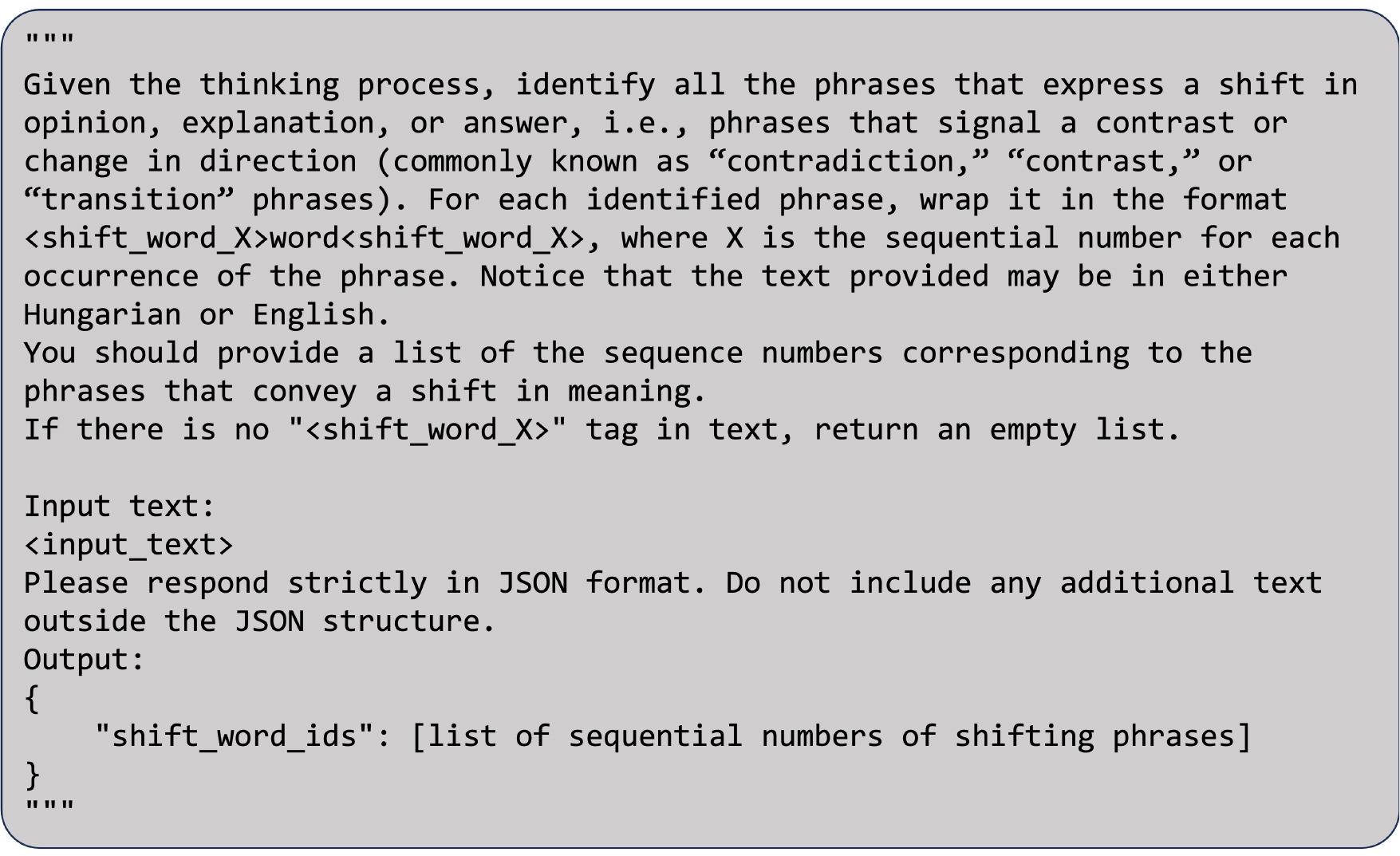}
    \caption{Prompt template for splitting the LRM's thinking process into thoughts on HuSimpleQA (step 2/2).}
    \label{fig:husimpleqa_shiftword_step2}
\end{figure*}

\begin{figure*}
    \centering
    \includegraphics[width=1\linewidth]{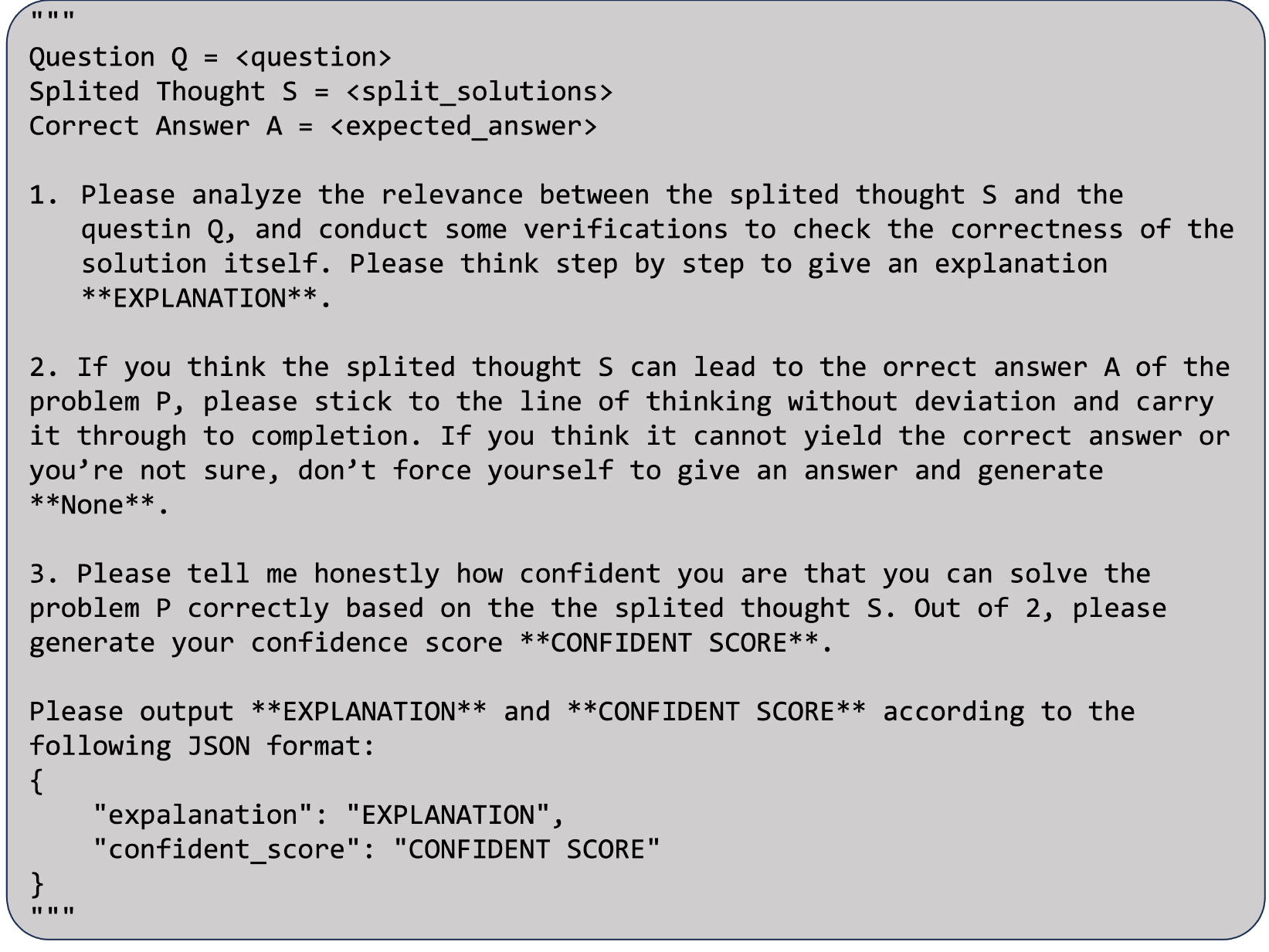}
    \caption{Prompt template for evaluating the correctness of the thoughts of the LRM's thinking process on HuSimpleQA.}
    \label{fig:husimpleqa_thought_evaluation}
\end{figure*}

\begin{figure*}
    \centering
    \includegraphics[width=1\linewidth]{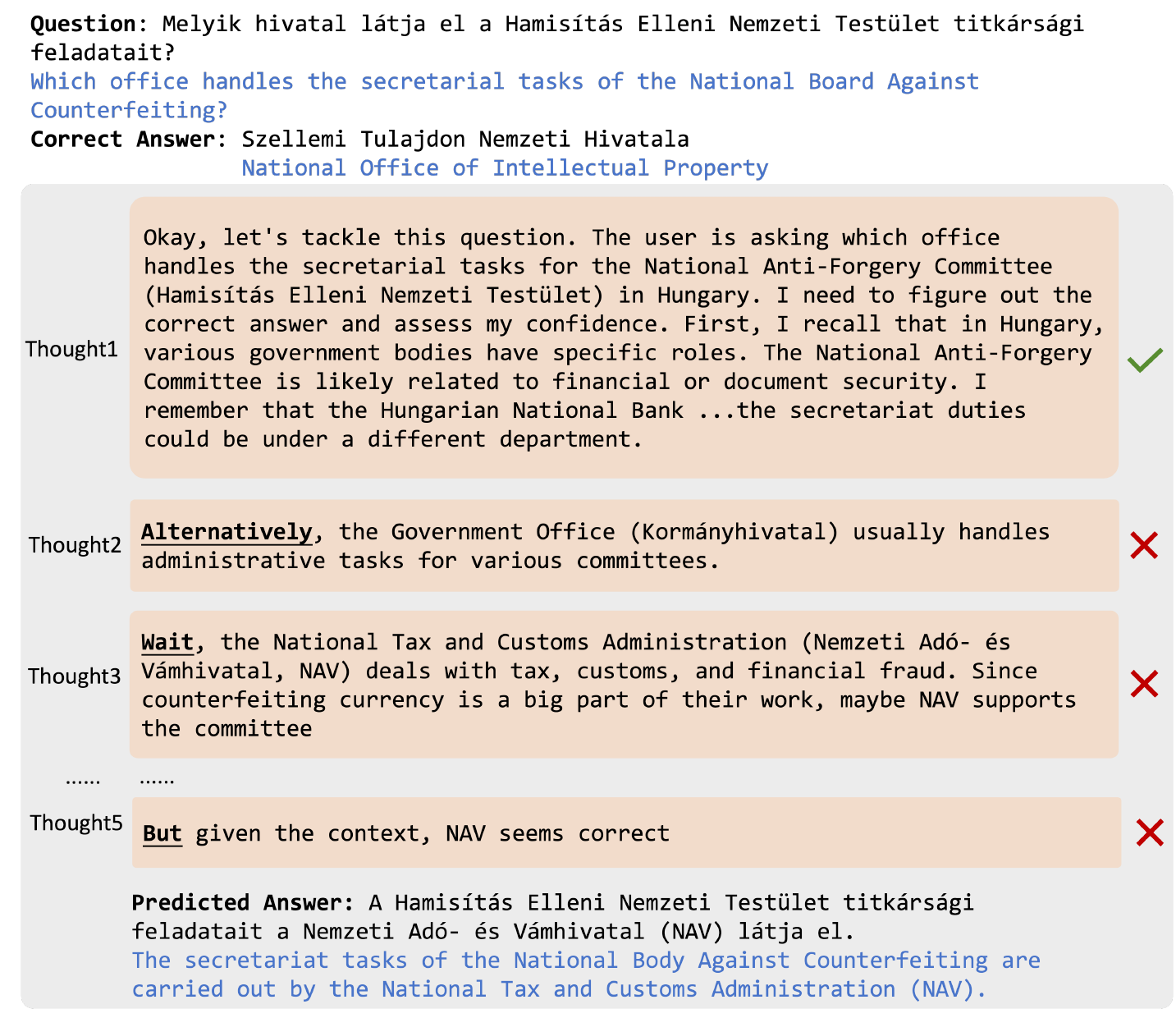}
    \caption{Example of the thoughts in Deepseek-R1's thinking process on HuSimpleQA. \textcolor{green}{\checkmark} and \textcolor{red}{$\times$} indicate the correctness of the thoughts. The original text is in black, while the translation into English is in \textcolor{blue}{blue}.}
    \label{fig:example of husimpleqa thought segmentation in ds}
\end{figure*}

\begin{figure*}
    \centering
    \includegraphics[width=1\linewidth]{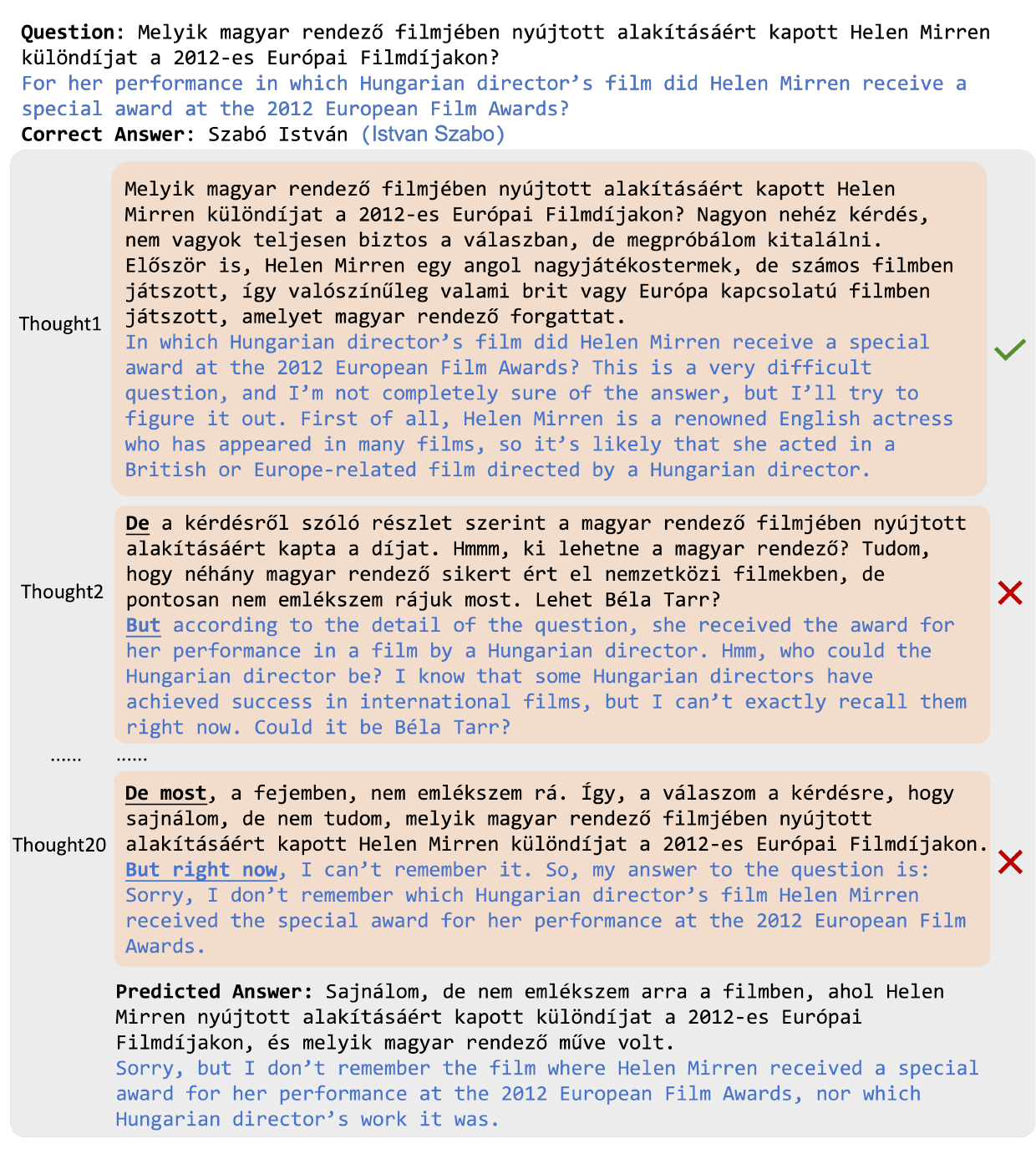}
    \caption{Example of the thoughts in QWQ's thinking process on HuSimpleQA. \textcolor{green}{\checkmark} and \textcolor{red}{$\times$} indicate the correctness of the thoughts. The original text is in black, while the translation into English is in \textcolor{blue}{blue}.}
    \label{fig:example of husimpleqa thought segmentation in qwq}
\end{figure*}



\begin{figure*}
    \centering
    \includegraphics[width=0.9\linewidth]{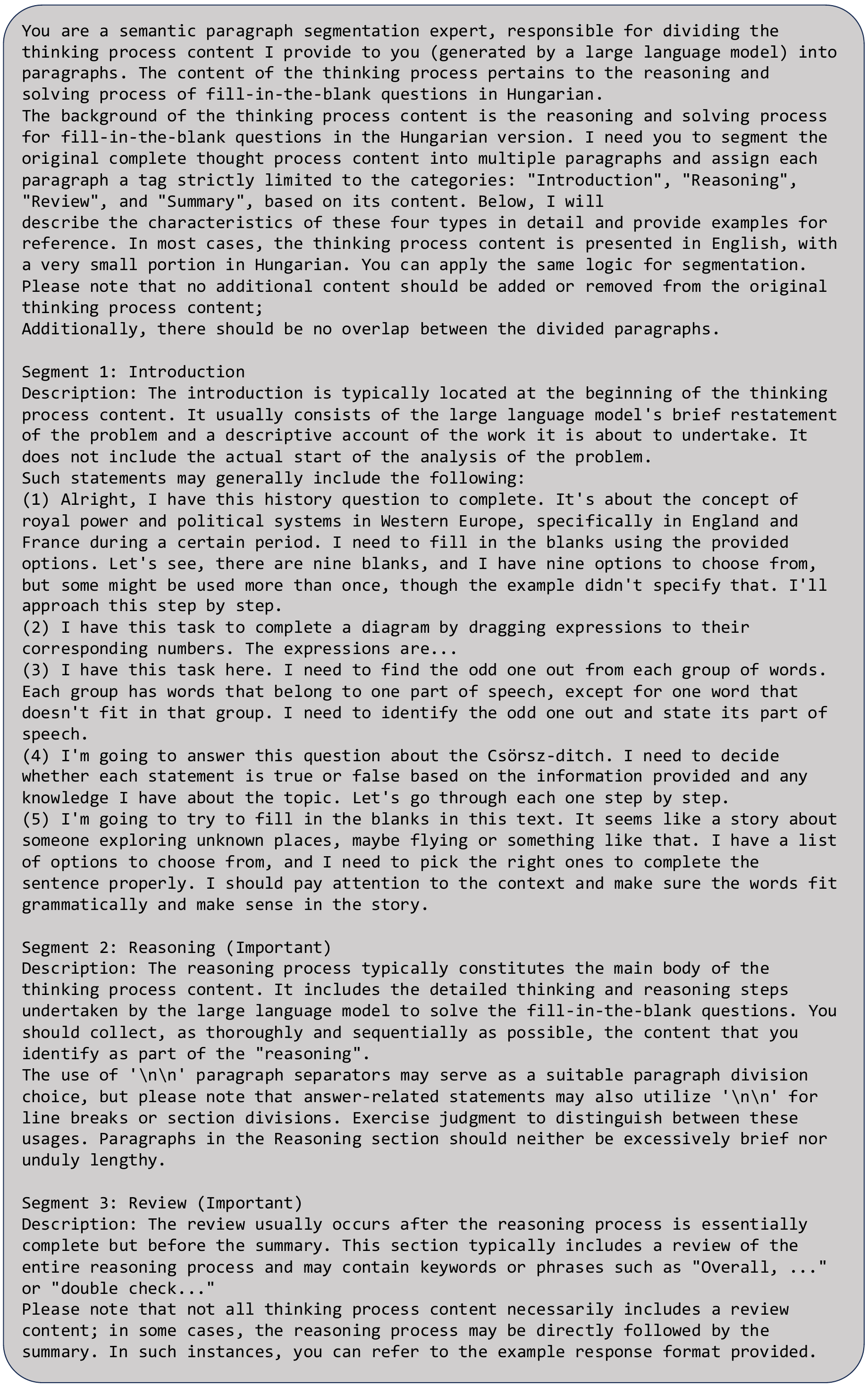}
    \caption{Prompt template for splitting LRM's thinking process into segments and categorizing these segments on HuMatchingFIB. (part 1/3).}
    \label{fig:Appendix_HuMatchingFIB_prompt_segment_p1}
\end{figure*}

\begin{figure*}
    \centering
    \includegraphics[width=0.95\linewidth]{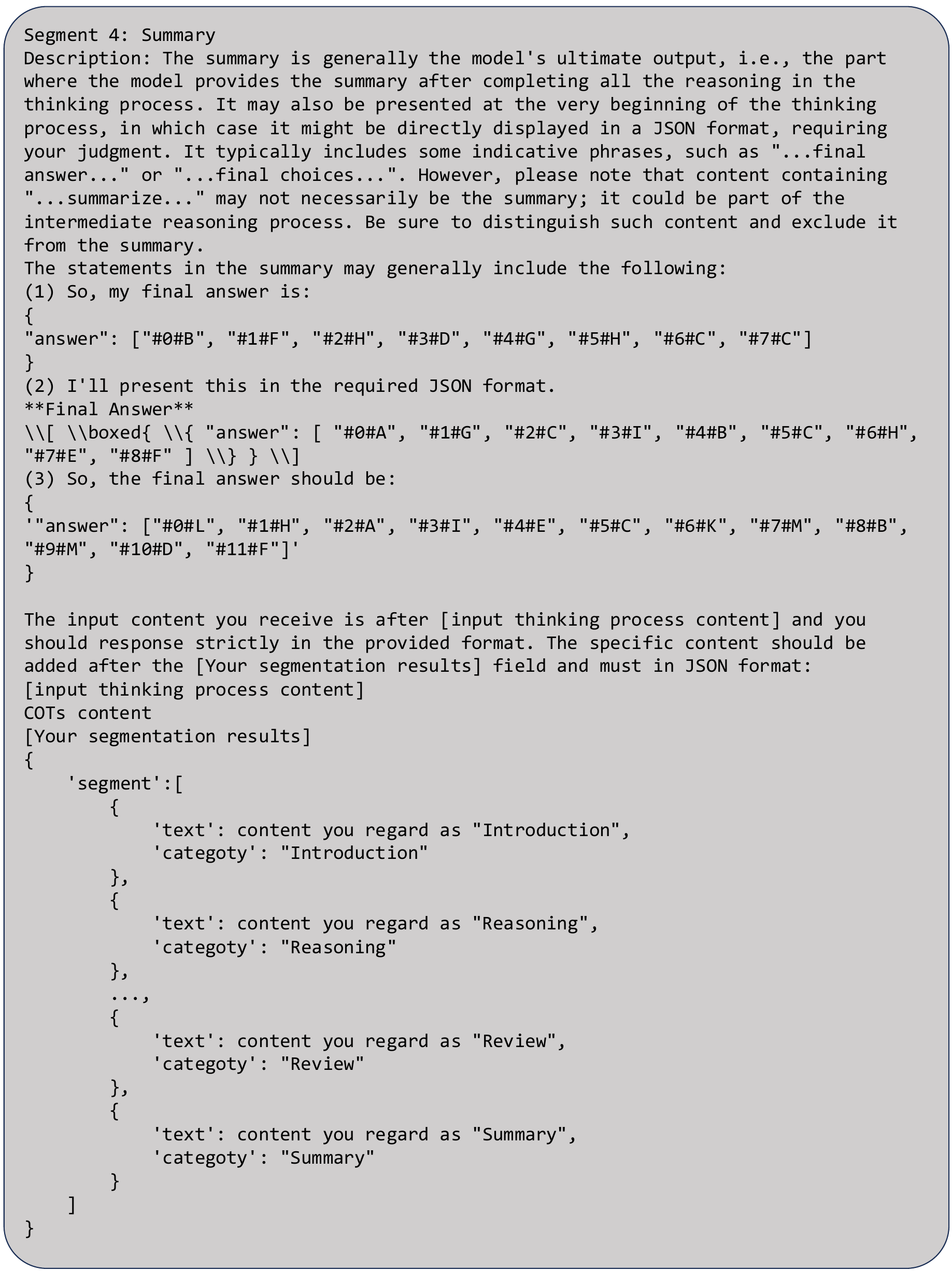}
    \caption{Prompt template for splitting LRM's thinking process into segments and categorizing these segments on HuMatchingFIB. (part 2/3).}
    \label{fig:Appendix_HuMatchingFIB_prompt_segment_p2}
\end{figure*}

\begin{figure*}
    \centering
    \includegraphics[width=0.95\linewidth]{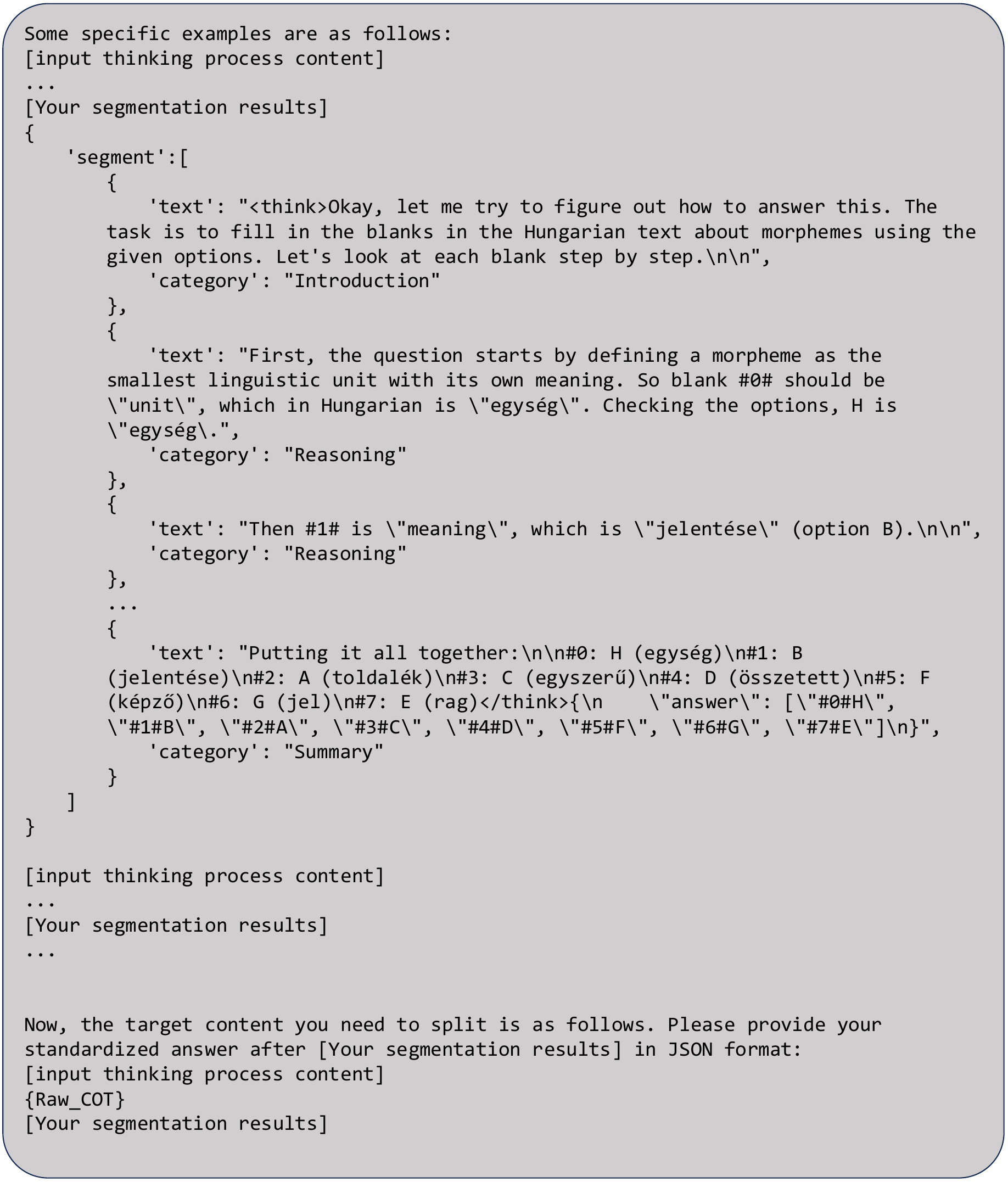}
    \caption{Prompt template for splitting LRM's thinking process into segments and categorizing these segments on HuMatchingFIB. (part 3/3).}
    \label{fig:Appendix_HuMatchingFIB_prompt_segment_p3}
\end{figure*}


\begin{figure*}
    \centering
    \includegraphics[width=0.9\linewidth]{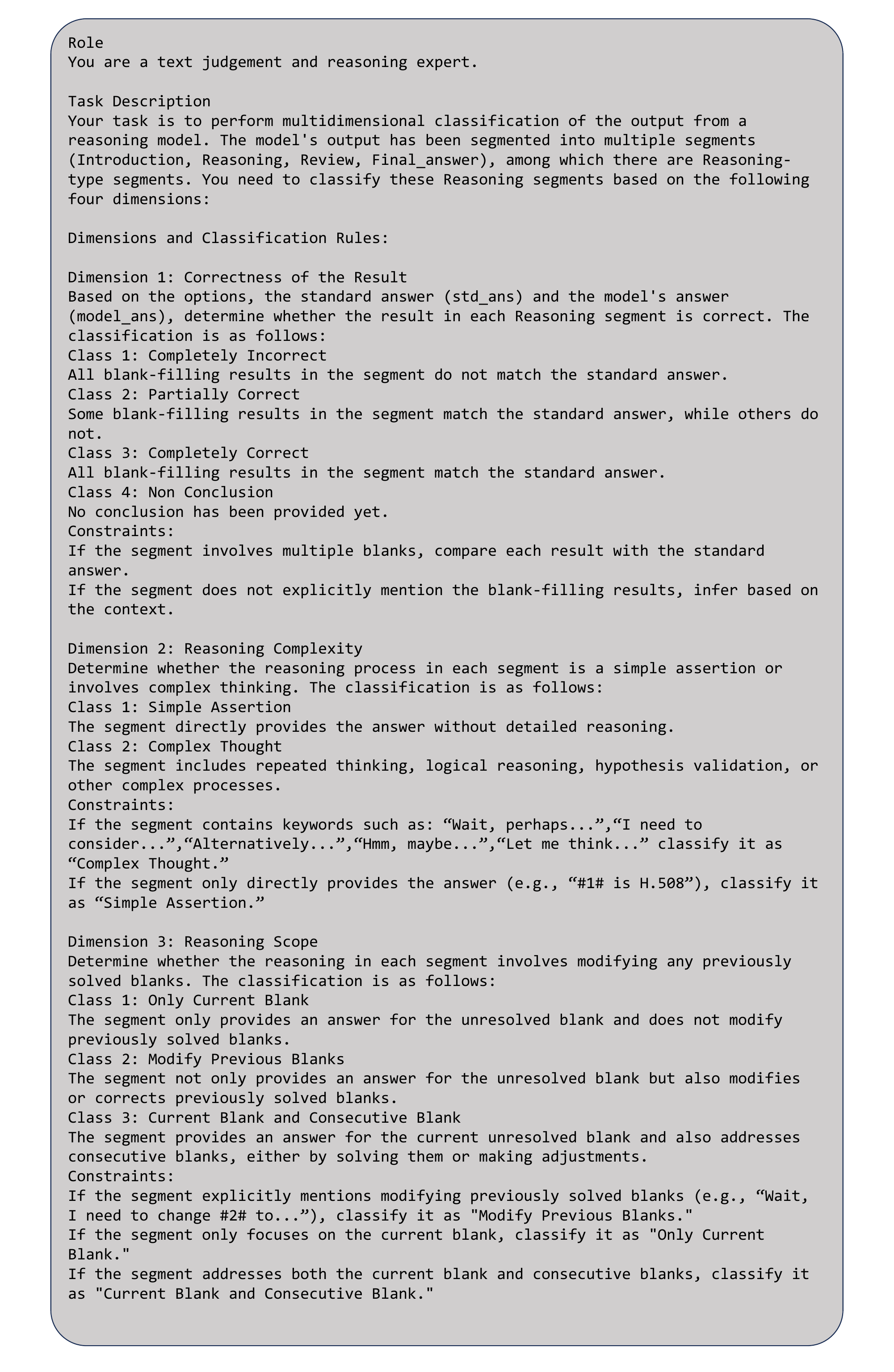}
    \caption{Prompt template for tagging the \texttt{reasoning} segments of LRM's thinking process along the four dimensions on HuMatchingFIB (part 1/2).}
    \label{fig:Appendix_HuMatchingFIB_reason_tag_dimension_prompt_p1}
\end{figure*}

\begin{figure*}
    \centering
    \includegraphics[width=0.9\linewidth]{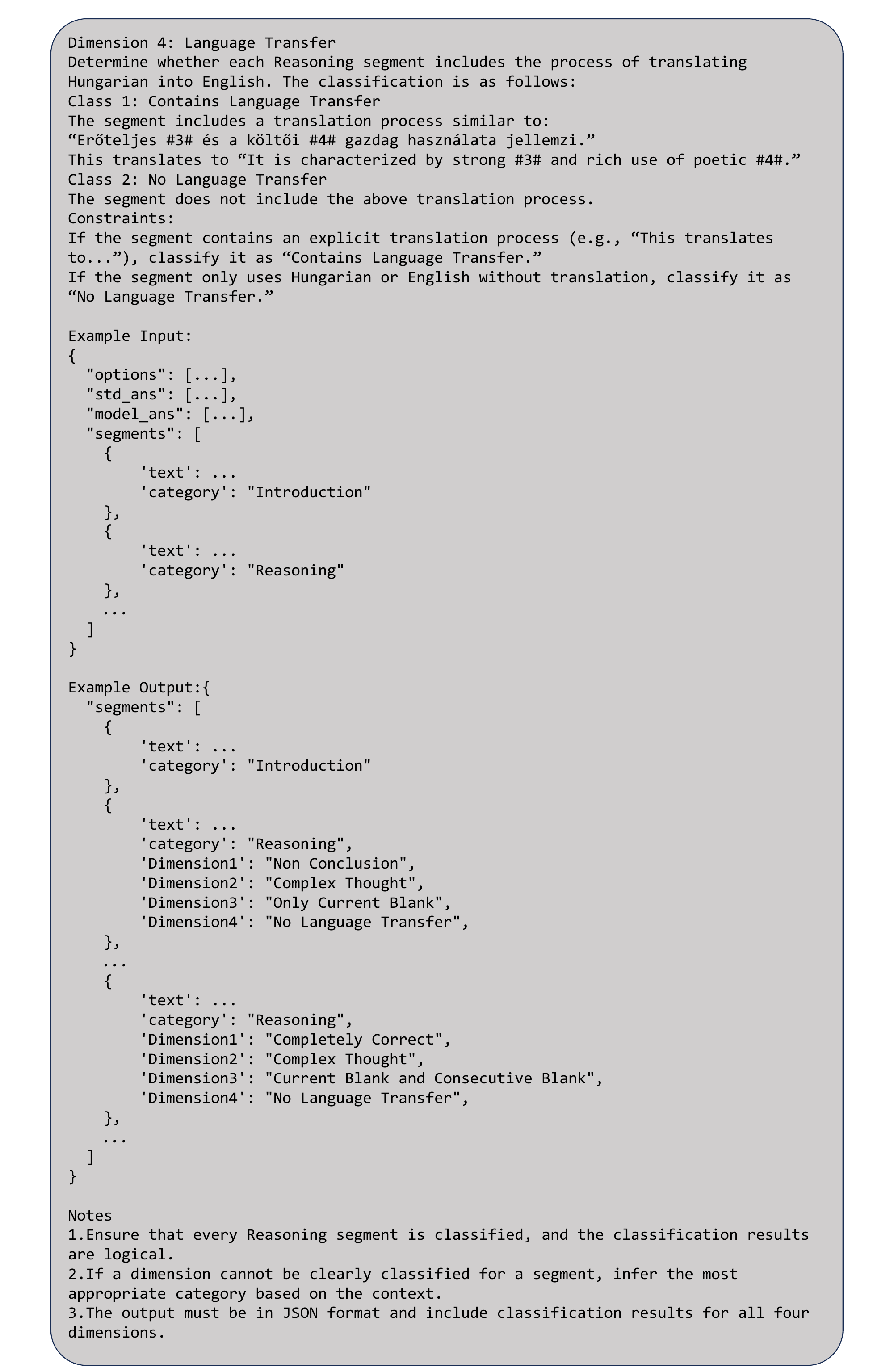}
    \caption{Prompt template for tagging the \texttt{reasoning} segments of LRM's thinking process along the four dimensions on HuMatchingFIB (part 2/2).}
    \label{fig:Appendix_HuMatchingFIB_reason_tag_dimension_prompt_p2}
\end{figure*}

\begin{figure*}
    \centering
    \includegraphics[width=1\linewidth]{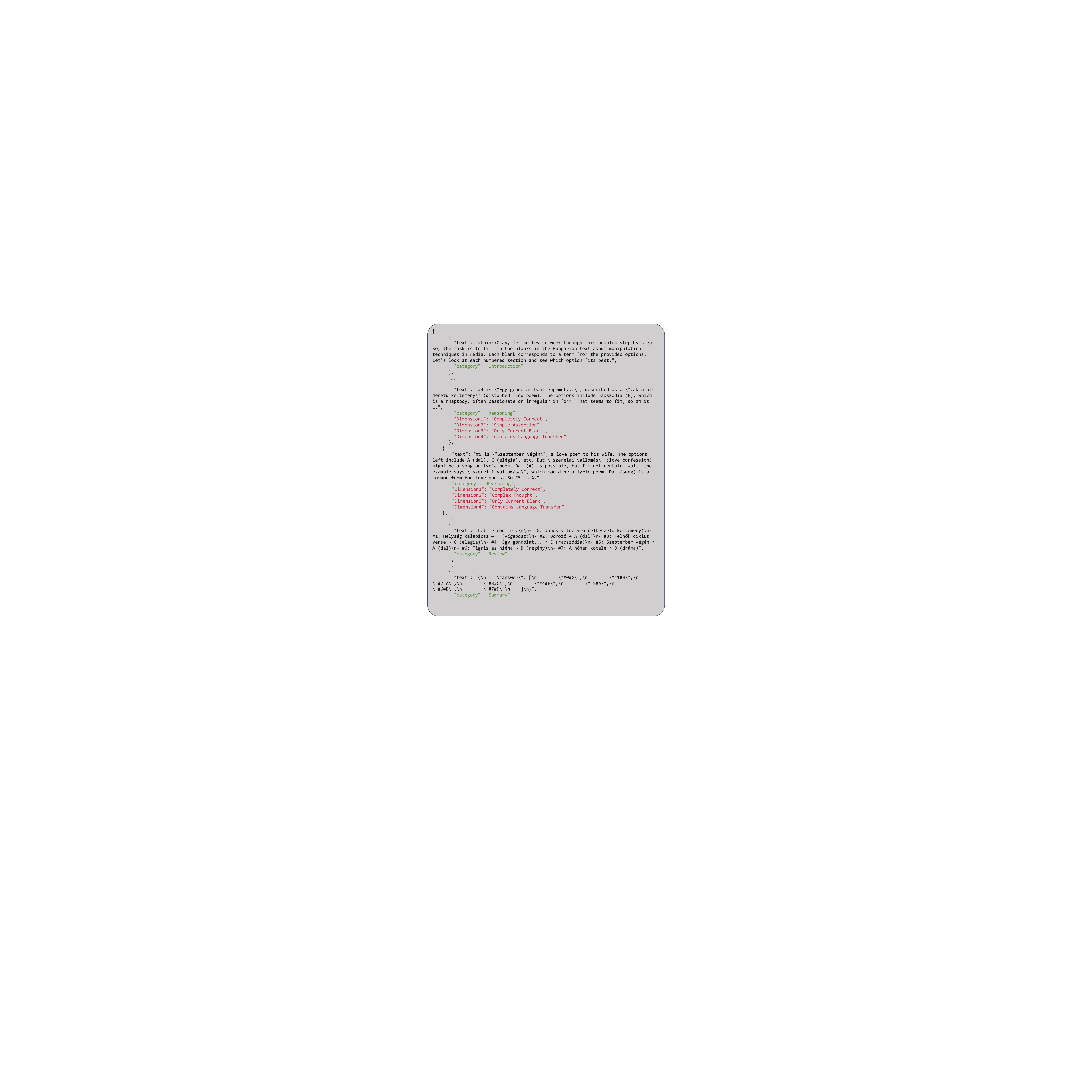}
    \caption{Example of splitting Deepseek-R1's thinking process into segments, categorizing these segments, and tagging the \texttt{reasoning} segments on HuMatchingFIB. The categorizing results are in \textcolor{green}{green} and the tagging results are in \textcolor{red}{red}.}
    \label{fig:humatchingfib_segment_example_deepseek}
\end{figure*}

\begin{figure*}
    \centering
    \includegraphics[width=1\linewidth]{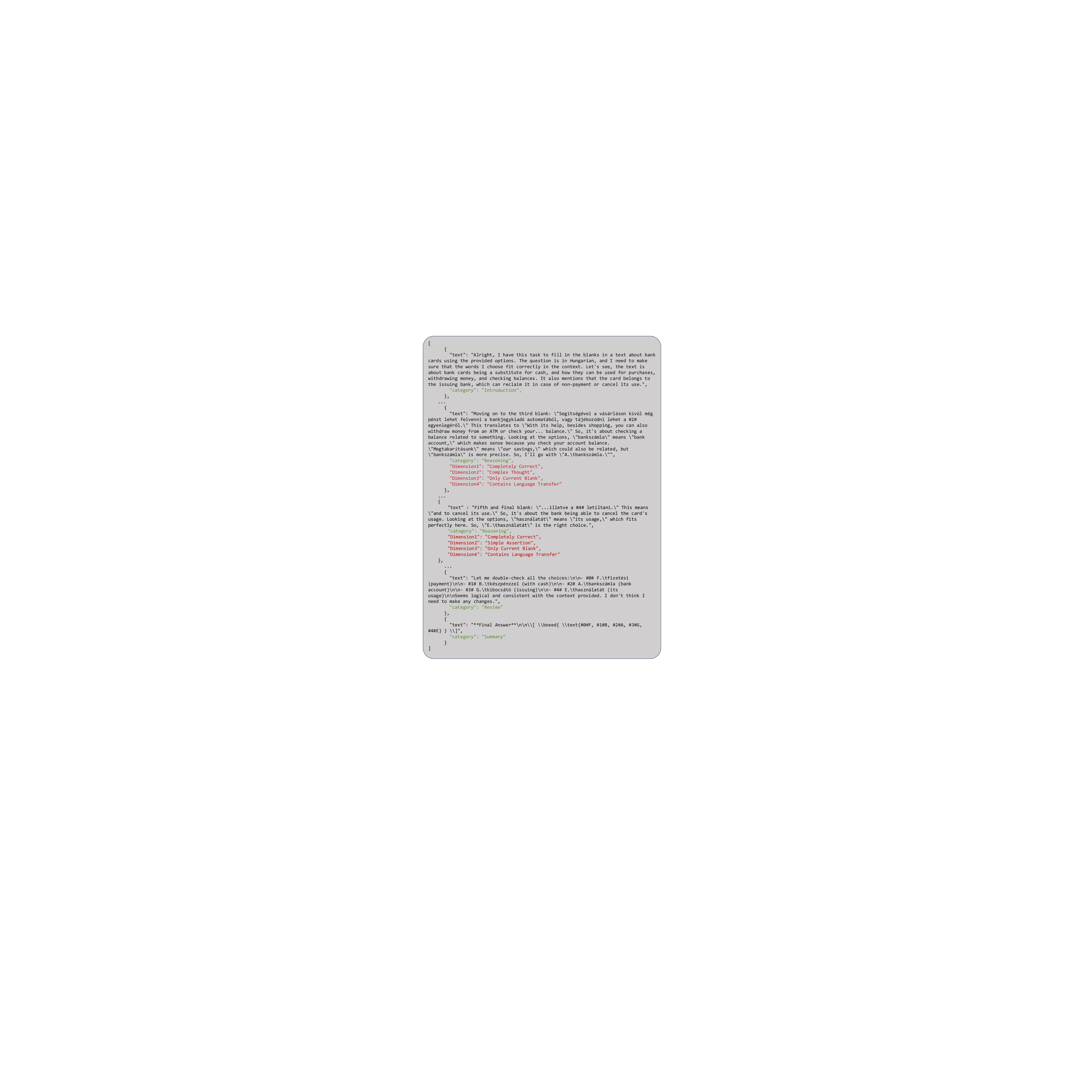}
    \caption{Example of splitting QwQ's thinking process into segments, categorizing these segments, and tagging the \texttt{reasoning} segments on HuMatchingFIB. The categorizing results are in \textcolor{green}{green} and the tagging results are in \textcolor{red}{red}.}
    \label{fig:humatchingfib_segment_example_qwq}
\end{figure*}



\end{document}